\documentclass[article,12pt]{article} 

\usepackage{amscd,amsmath,amssymb,amsfonts,latexsym,mathrsfs,amsthm}
\usepackage{graphicx} 
\usepackage[numbers]{natbib}
\usepackage{graphics,epsfig}
\usepackage{sectsty}
\usepackage{natbib}
\usepackage[english]{babel}
\usepackage{amsmath}
\usepackage{amsfonts}
\usepackage{amssymb,amsthm}
\usepackage{bm}
\usepackage{mathrsfs}
\usepackage[usenames]{color}
\usepackage{xcolor}
\usepackage{subfigure}
\usepackage{url}
\usepackage{changepage}

\usepackage{multicol}
\usepackage{multirow}
\usepackage{pbox}

\UseRawInputEncoding

\makeatother

\usepackage{marginnote}

\renewcommand{\vec}[1]{\mathbf #1}
\newcommand{\rvec}[1]{\bm #1}

\def\vmu{\rvec{\mu}}
\def\mSigma{\vec{\Sigma}}
\def\mS{\vec{S}}

\def\vrest{\widehat{\rvec{\rho}}}

\def\vcoef{\vrest}
\def\vtheta{\boldsymbol\theta}
\def\vphi{{\bm \varphi}}
\def\ph{{\varphi}}

\def\f{\vec{f}}
\def\e{\rvec{e}}
\def\fvest{\widehat{\f}}
\def\bx{{\mathbf x}}
\def\x{{\mathbf x}}
\def\X{{\mathbf X}}

\def\y{{\mathbf y}}

\def\f{{\mathbf f}}
\def\bx{{\mathbf x}}
\def\x{{\mathbf x}}
\def\X{{\mathbf X}}

\def\y{{\mathbf y}}

\newcommand{\by}{{\mathbf y}}

\newcommand{\fhat}{\widehat{f}}

\newcommand{\R}{\mathbb{R}}
\newcommand{\Exp}{\mathbb{E}}
\newcommand{\D}{\mathcal{D}}
\newcommand{\N}{\mathcal{N}}

\newtheorem{Remark}{\bf Remark}

\title{A Joint introduction to Gaussian Processes and Relevance Vector Machines with Connections to Kalman filtering and other Kernel Smoothers} 

\author{Luca Martino$^\star$ and Jesse Read$^\top$ \\
{\small$^\star$ Dept.\ of Statistical Signal Processing, Universidad Rey Juan Carlos,  Madrid (Spain)}\\
{\small $^\top$ LIX, Ecole Polytechnique, Institut Polytechnique de Paris, France.} }

\date{}
 
\begin{document}

\maketitle

\thispagestyle{empty}

\begin{abstract} 
The expressive power of Bayesian kernel-based methods has led them to become an important tool across many different facets of artificial intelligence, and useful to a plethora of modern application domains, providing both power and interpretability via uncertainty analysis. 
This article introduces and discusses two methods which straddle the areas of probabilistic Bayesian schemes and kernel methods for regression: Gaussian Processes and Relevance Vector Machines. 
Our focus is on developing a common framework with which to view these methods, via intermediate methods a probabilistic version of the well-known kernel ridge regression, and drawing connections among them, via dual formulations, and discussion of their application in the context of major tasks: regression, smoothing, interpolation, and filtering. 
Overall, we provide understanding of the mathematical concepts behind these models, and we summarize and discuss in depth different interpretations and highlight the relationship to other methods, such as linear kernel smoothers, Kalman filtering and Fourier approximations. Throughout, we provide numerous figures to promote understanding, and we make numerous recommendations to practitioners. Benefits and drawbacks of the different techniques are highlighted.  
To our knowledge, this is the most in-depth study of its kind to date focused on these two methods, and will be relevant to theoretical understanding and practitioners throughout the domains of data-science, signal processing, machine learning, and artificial intelligence in general. 
\newline 
{\bf Keywords:}
	Gaussian processes, Relevance Vector Machines, Bayesian Learning, Bayesian Ridge, Kernel Smoothing, Kalman Filtering.
\end{abstract}

%


\section{Introduction}

This work details and discusses techniques and methods lying on the intersection of two areas: probabilistic Bayesian schemes and kernel methods; in a regression framework. Such techniques have become increasingly popular in statistics, signal processing, and machine learning \cite{bishop2006pattern,rasmussen2003gaussian_b,mackay1998introduction,scholkopf2002learning,Sarkka_2013}. The expressive power of these methods increases with the number of data points observed, and they can be effective for dealing with structured (non-tabular) sources such as sequential data. Indeed, despite the soaring popularity of deep neural network architectures in recent decades, the methods we approach in this work are still relevant to a plethora of modern application domains; and an understanding of the mathematical concepts behind them is still of fundamental importance across science and mathematics under the general umbrella of modern artificial intelligence. 

More specifically, we look at Gaussian Processes (GPs) \cite{rasmussen2003gaussian,mackay1998introduction} and Relevance Vector Machines (RVMs) for regression \cite{Tipping01,QuineroCandela04phd} -- both Bayesian nonparametric approaches which have yielded convincing results in recent years and attracted a correspondingly significant interest \cite{gewali2019gaussian,alvarez2009latent,gomez2016efficient,svendsen2017joint,svendsen2020active,MartinoRVM}.
The main scope of this manuscript encompasses a unified introduction to GPs, RVMs. We link these two methods via Kernel Ridge Regression -- which we refer to as quasi GP in the probabilistic sense (often known elsewhere as Bayesian Ridge Regression); covering important material presented in the literature, which deserves to be properly highlighted \cite{Silverman85,Szeliski1987RegularizationUF,Tipping01,RasmussenQuinonero05,Poggio20}.  We allow a correct comparison between these  techniques and perform a detailed analysis of uncertainty estimation. 
Moreover, we leverage the opportunity to connect these methods to a number of other important methods in neighboring areas, including splines, kernel smoothers, k-Nearest Neighbor (kNN) schemes and a Fourier interpolation \cite{bishop2006pattern,murphy12}, 
and particularly exploring the connection between GPs and Kalman filtering which has not been completely elaborated in the literature (although recent work remarked upon this link \cite{Hartikainen10,pmlr-v22-sarkka12,Sarkka_2013} we propose a gentle explanation with examples in discrete time, that is not specifically discussed in those works). 
We considering different possible scenarios in the framework of regression models, including prediction, filtering, smoothing and interpolation, providing the specific solutions in each one of these cases.   

We will see that the main benefit of the RVM approach is the flexibility in the choice of the basis functions, whereas the main advantage of the GP approach is the good behavior of the predictive variance. And we will discuss the implication thereof. We will also discuss the interpretability of the chosen bases/kernel functions, the uncertainty analysis with each techniques and the generation of random functions from (direct or induced) priors and/or posteriors over the underlying function. Furthermore, several related concepts  well-known in signal processing (e.g.,  the Fourier upsampling in Section \ref{FourierRecFilter}, and the linear digital filters in Section \ref{DigitalFilterSect}) are described and connected to the rest of techniques. In this sense, this work builds bridges among different concepts in statistics, machine learning and signal processing. 

\medskip

The paper is structured as follows: 
\begin{itemize}
	\item In Section \ref{Sect1}, we introduce the notation and provides a joint introduction of the RVM and GP  methods (considering joint formulas and properties). 
	\item The derivation of the RVM solution is given in Section \ref{RVMsect}. 
	\item The probabilistic version of KRR is described in Section \ref{ProbKRR}. 
	\item The GP derivation is provided in Section \ref{GPsect}. 
	\item Section \ref{DualRepres} describes the dual representation of RVM as a GP.  
	\item An initial summary with important considerations and remarks is provided in Section \ref{MostImportSect}; then 
	\item Section \ref{UncertGP_SECT} provides a discussion regarding  the uncertainty analysis with GPs.
	\item Section \ref{linearSmoothertodo} shows that RVM and GP can be seen as linear kernel smoothers and describes other well-known examples in the literature. 
	\item In Section \ref{GPKalmanSect}, we describes the connections between Kalman filtering (and smoothing) with the GP solution. 
	\item A final discussion and concluding summary is provided in Section \ref{ConclSect}.
\end{itemize}

\section{A Joint Framework for RVMs and GPs}\label{Sect1}

In this section, we introduce the main notation and the problem statement. Moreover, we provide a joint introduction of GPs and RVMs in the form of a common framework, elaborating all the equations shared by both models.
Namely, we introduce the analytic form of the regression function $\widehat{f}(\x)$, the observation model and the likelihood function  (all shared by both methods), as well as the design matrix and the interpolation case (where both schemes provide the same solution).  The main notation of the work is summarized in Table \ref{table_notation0}.  

\begin{table}[!ht]
	\caption{ Main notation of the work.}
	\vspace{0.2cm}
\centering
\renewcommand{\arraystretch}{1.2}
	\begin{tabular}{ll}
		\hline
		 $\x \in \R^{d_X}$ & a $d_X$-dimensional input observation  \\
		 $y\in \mathbb{R}$ & a scalar output \\ 
		 $\rvec{y}$ & $\rvec{y}=[y_1,\ldots,y_N]^\top$, vector of outputs/observations \\
		$e$, $\e$  & Gaussian noise perturbation $e \sim \N(\mu_e,\sigma^2_e)$, $\e \sim \N(\vmu_e,\mSigma_e)$ \\
		 $\D$ &  Dataset: $\D = \{\x_i,y_i\}_{i=1}^N$ or $\D = \{\X,\y\}$, of $N$ points  \\ 
		 $f(\x)$ & underlying/hidden function (unknown) $f$ evaluated at $\x$ \\
		 $\fhat(\x)$ & regression function $\fhat$, est./approx.\ of $f$, evaluated at $\x$ \\
		 $\f$ or $\f(\x)$ & vector $[f_1(\x),\ldots,f_N(\x)]^\top$, \\
		 $\fvest$ or $\fvest(\x)$ & vector $[\fhat_1(\x),\ldots,\fhat_N(\x)]^\top$, \\
		 $\vtheta$ & vector of (hyper-)parameters of the model $\vtheta = [\theta_1,\ldots,\theta_{d_\theta}]^{\top}$ \\ 
		 $\psi_i(\x,\x_i)$ & nonlinear basis, localized around $\x_i$ \\
		 ${\bm\psi}(\x)$ & ${\bm\psi}(\x)=[\psi_1({\bf x},{\bf x}_1),...,\psi_N({\bf x},{\bf x}_N)]^{\top}$, the $N\times 1$ design vector. \\ 
		 ${\bm \Psi}$ & the $N\times N$ design matrix. \\ 
		 $\vcoef$ & vector of estimated coefficients, or $\vcoef(\y)$ if determined by $\y$ \\
		 $\vphi(\x)$ & vector of smoothing kernels, $\vphi = [\varphi_1(\x|\x_1),\ldots,\varphi_N(\x|\x_N)]$ \\
		\hline
	\end{tabular}
	\label{table_notation0}
\end{table}

\noindent
\newline
{\bf Observation model.} We have a dataset consisting of $N$ data points,  $\D = \{\x_i,y_i\}_{i=1}^N$, where each $i$-th \emph{input} $\x_i=[x_{i,1},x_{i,2},\ldots,x_{i,d_X}]^{\top} \in \mathcal{X}\subseteq \R^{d_X}$ is associated with \emph{scalar output} $y_i \in \R$.\footnote{In this work, we consider single-output regression problems. For multi-output approaches, see \cite{Alvarez_MultiOutput,READ2020471}. } 
To simplify notation we consider $\Exp[y_i]=0$ for all $i=1,\ldots,N$, without loss of generality. This assumption can be easily relaxed with the addition of a bias (possibly varying with $\x$) to the probabilistic models. Namely, the goal is to approximate an unknown underlying function 
	$f(\x): \R^{d_X} \rightarrow \R$,
which we assume has generated our training points, in the form 
\begin{equation}
\label{EqNum1}
y_i=f({\bf x}_i)+e_i,
\end{equation}
where $e_i$ is a Gaussian perturbation with zero mean and variance $\sigma_e^2$, i.e., $e_i \sim \mathcal{N}(e|0,\sigma_e^2)$.  That is to say, ideally, we want to learn some model $\fhat$ (let us call it the \textit{regression function}) such that $
\fhat \approx f$. 
We emphasize that this sample pair $\x,y$ do \emph{not necessarily} belong to the training set, which means that in fact $y$ may not be observed at all. In prediction, we want to \emph{generalize} to new test points. Note that, $\y$, $f(\x)$ are random variables, whereas the inputs $\x$ play the role of (non-random) parameters. More specifically, for each $\x\in \mathcal{X}$, then $f(\x)$ represents a different random variable. 
\newline
\newline
{\bf Regression function.} In this work, we describe theory and properties of different methods where the regression function can be expressed as a linear combination of $N$ non-linear functions, i.e.,  
\begin{equation} 
	\label{aquiF}
	\widehat{f}({\bf x})=\sum_{i=1}^N \widehat{\rho}_i \psi_i({\bf x},{\bf x}_i),
 \end{equation}
where the  non-linear functions $\psi_i({\bf x},{\bf z}): \mathcal{X}\times \mathcal{X}\rightarrow \mathbb{R}$ have been selected in advance by the user,  according to the problem domain, i.e.,  encoding some prior knowledge about the underlying function $f(\x)$.  
The coefficients $\widehat{\rho}_n \in \mathbb{R}$, $n=1,\ldots,N$, are analytically obtained according to the chosen probabilistic derivation, e.g., either RVM or GP as covered in this article. 
Note that non-linearity $\psi_i$ is indexed by $i$, 
since, in RVM, these functions can 
differ among inputs $i$. In a GP model, we will consider the simpler notation $ \psi_i({\bf x},{\bf x}_i)= \psi({\bf x},{\bf x}_i)$, precisely as we need to impose that the analytical form of $\psi$ does not vary with $i$.  



\begin{figure}[!h]
	\centering
	\centerline{	
		\includegraphics[width=10cm]{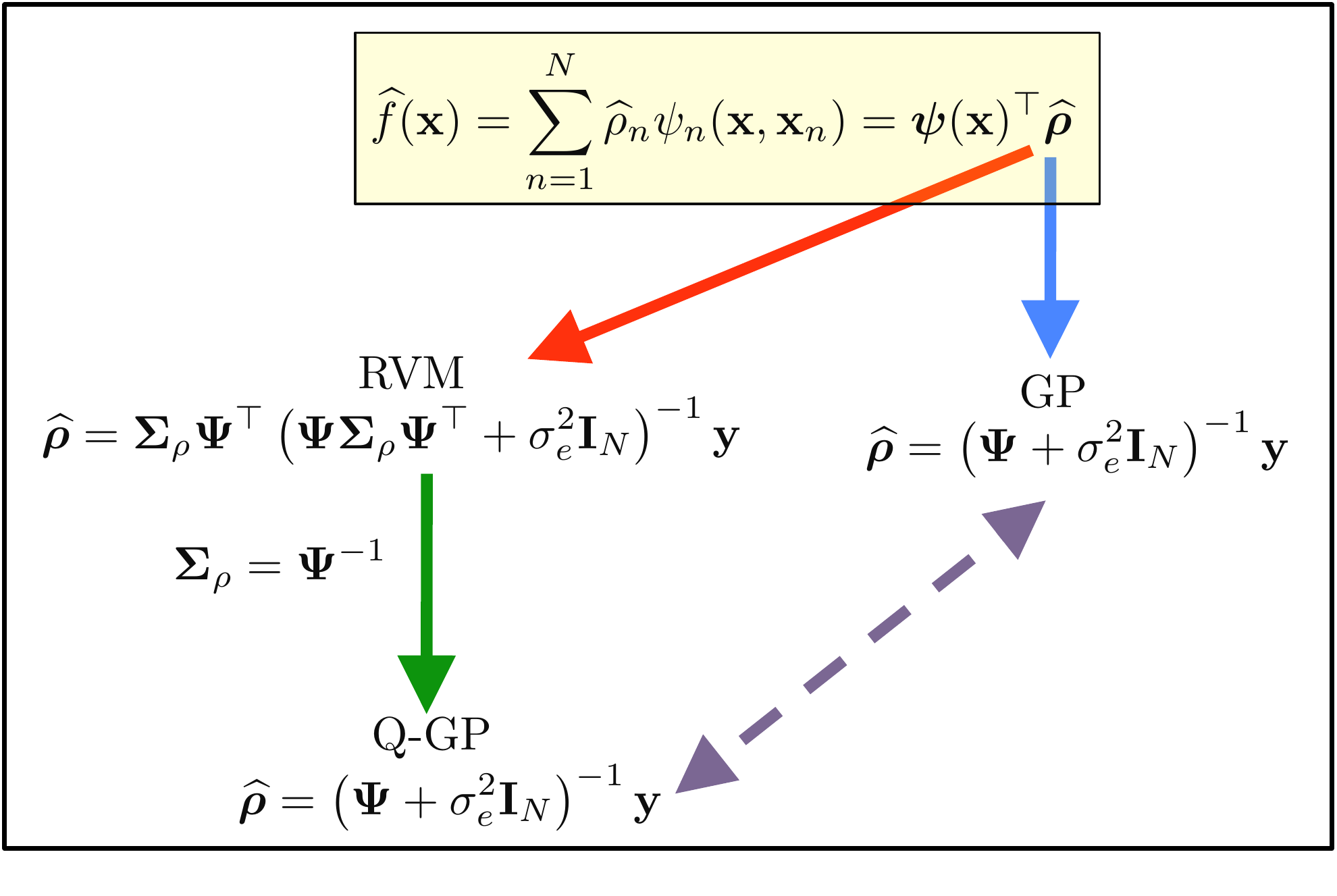}
	}
	 \caption{Graphical representation of the relationships among the solutions $\widehat{f}({\bf x})$ of the main methods described in this work, RVM, GP and Quasi-GP (Q-GP). Note that  the estimated function $\widehat{f}({\bf x})$ coincides in GP and Q-GP. Note that, in GP and Q-GP, the matrix ${\bm \Psi}$ must be symmetric, ${\bm \Psi}={\bm \Psi}^{\top}$, and invertible.
	 }
	\label{figmain_starting2}
\end{figure}

\begin{figure}[!h]
	\centering
\centerline{	
	\subfigure[]{\includegraphics[width=7cm]{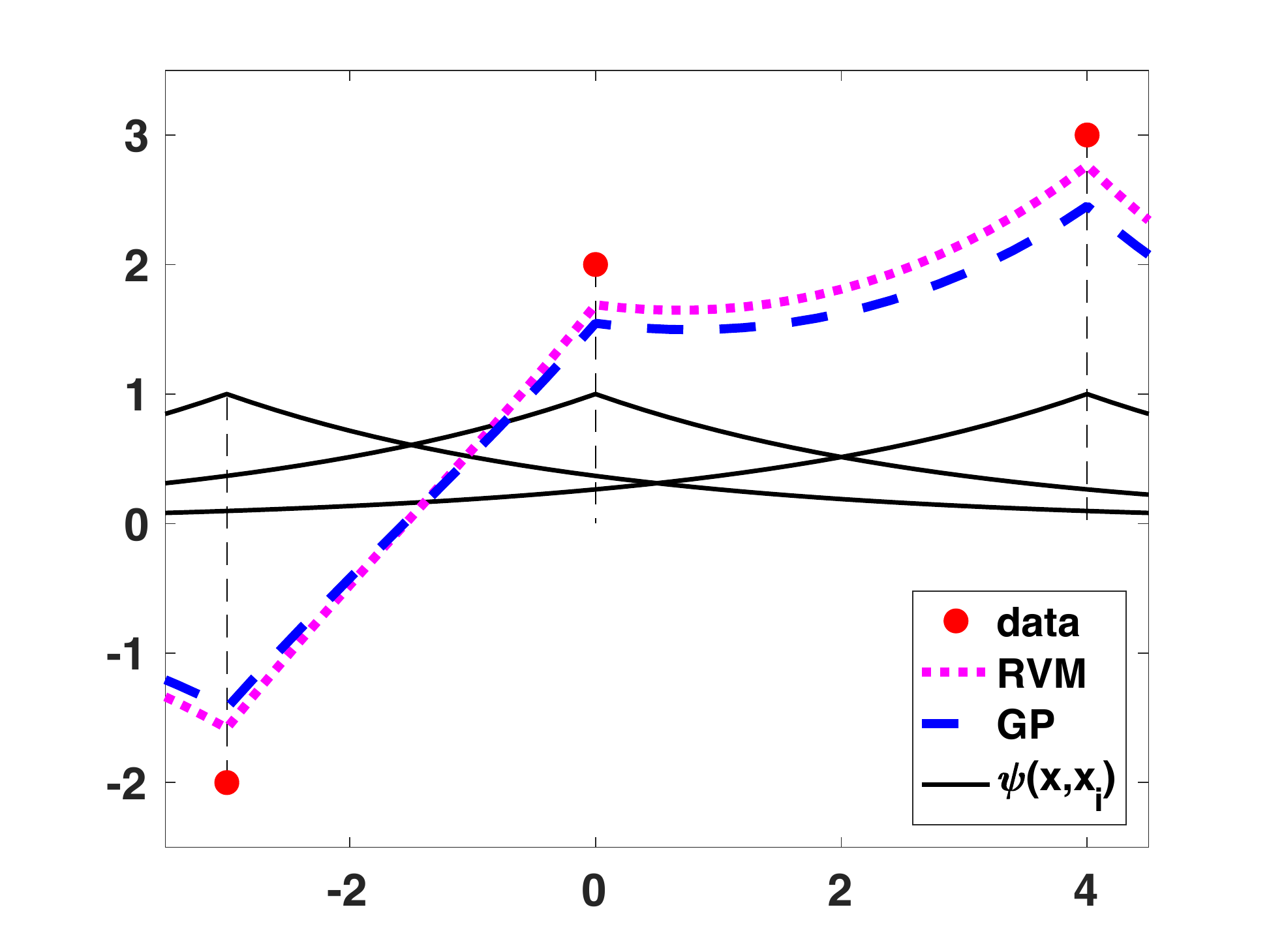}}
	\subfigure[]{\includegraphics[width=7cm]{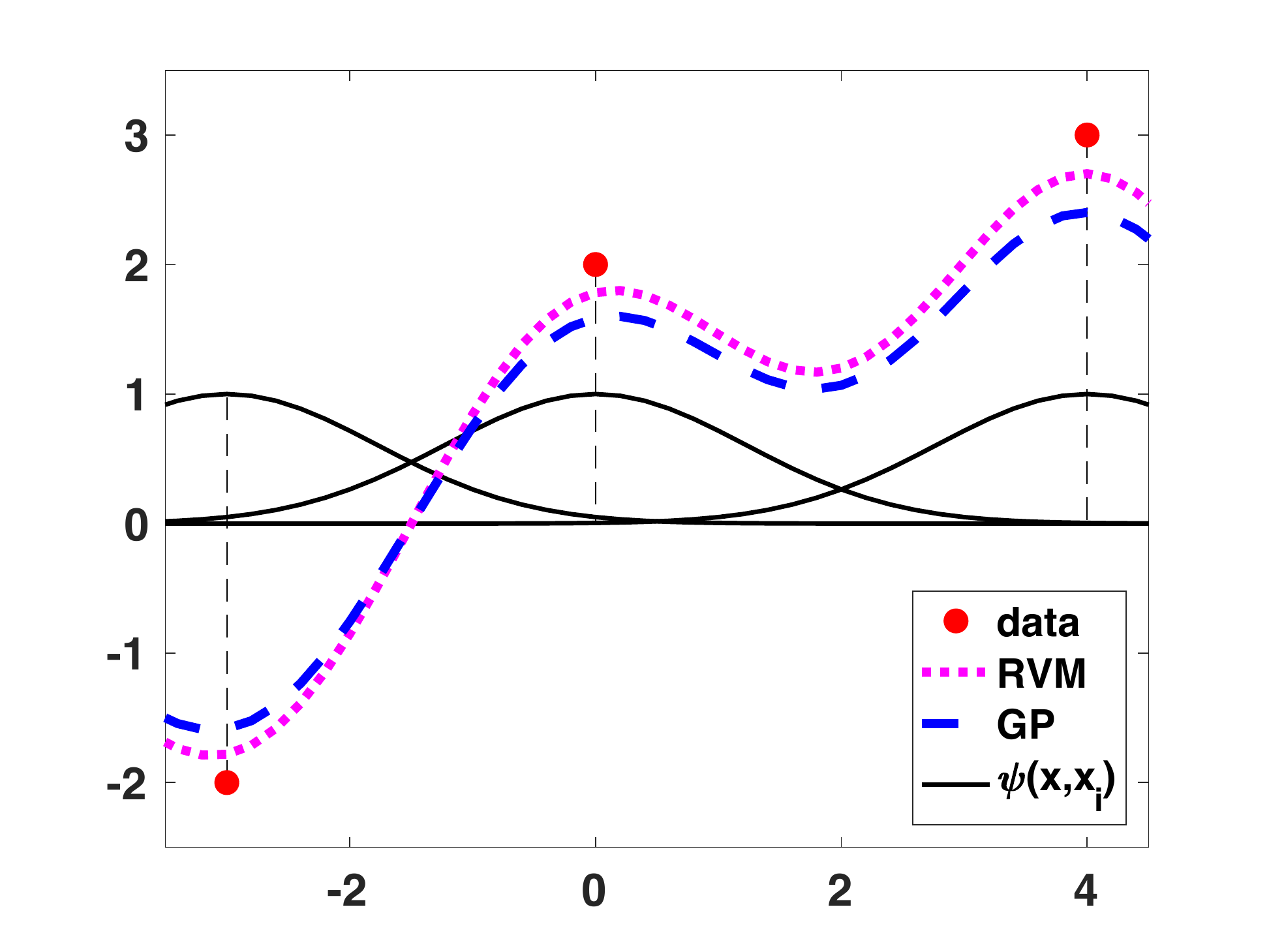}}
}
	\caption{ Examples of RVM and GP solutions $\widehat{f}(\x)$ with {\bf (a)} Laplacian and  {\bf (b)} Gaussian bases (depicted with solid lines). The $N=3$ data points as shown with red dots. The GP solutions are depicted with  dashed lines, whereas the RVM solutions are shown with dotted lines.}
	\label{figmain_starting}
\end{figure}

{\Remark The number of components in Eq.~\eqref{aquiF} is exactly the number of data, i.e., $N$. Therefore, the flexibility of the model increases as the number of data $N$ grows. The regression methods, represented by Eq.~\eqref{aquiF}, are {\it non-parametric} models. 
}
 {\Remark The  RVM and GP solutions differ for the choice of the coefficients $\widehat{\rho}_n$.  This different choice is due to  the probabilistic approach employed by each method. 
}
\newline
\newline
Defining also the $N\times 1$ {\it design} vector ${\bm \psi}({\bf x})=[\psi_1({\bf x},{\bf x}_1),\ldots,\psi_N({\bf x},{\bf x}_N)]^{\top}$, the approximating function (of all the methods derived in this work) can be also written in a vectorial form,
\begin{equation}
\label{aquiF2}
\widehat{f}({\bf x})={\bm \psi}({\bf x})^{\top} \widehat{{\bm \rho}}.
 \end{equation}
 Note that the the coefficients will be determined considering the vector of outputs ${\bf y}=[y_1,\ldots,y_N]^{\top}$, hence a more complete notation will be $\widehat{{\bm \rho}}=\widehat{{\bm \rho}}({\bf y})$. 
Let us also define the $N\times N$ {\it design matrix} ${\bm \Psi}=[{\bm \psi}({\bf x}_1),\ldots., {\bm \psi}({\bf x}_N)]^{\top}$, i.e., 
\begin{eqnarray}
\label{psiMAT}
{\bm \Psi} =
\begin{bmatrix}
\psi_1({\bf x}_1,{\bf x}_1) & \psi_1({\bf x}_1,{\bf x}_2) &  \hdots  &  \psi_1({\bf x}_1,{\bf x}_N) \\
\psi_2({\bf x}_2,{\bf x}_1)& \psi_2({\bf x}_2,{\bf x}_2) & \hdots & \psi_2({\bf x}_2,{\bf x}_N) \\
\vdots \\
\psi_N({\bf x}_N,{\bf x}_1) & \psi_N({\bf x}_N,{\bf x}_2) &\hdots & \psi_N({\bf x}_N,{\bf x}_N) 
\end{bmatrix}.
\end{eqnarray}
In the GP case, we will require that ${\bm \Psi}$ be symmetric, but for RVMs it could be a non-symmetric matrix.
We will show below that the vectors of coefficients for RVMs and GPs are given by the formulas
\begin{eqnarray}
\label{aquiCoeffnow}
\mbox{RVM:} \quad \widehat{{\bm \rho}}&=&\boldsymbol{\Sigma}_\rho{\bm \Psi}^{\top} \left({\bm \Psi}\boldsymbol{\Sigma}_\rho{\bm \Psi}^\top+\sigma_e^2 {\bf I}_N\right)^{-1}\by,  \nonumber \\
\mbox{GP:} \quad  \widehat{{\bm \rho}}&=&\left({\bm \Psi}+\sigma_e^2 {\bf I}_N\right)^{-1}\by, 
\end{eqnarray}
where $\boldsymbol{\Sigma}_\rho$ is a $N\times N$ matrix decided by the user. By substituting expressions \eqref{aquiCoeffnow} into Eq.~\eqref{aquiF2}, we obtain the following regression functions: 
\begin{eqnarray}
\label{formulafinalaqui}
\mbox{RVM:} \quad \widehat{f}({\bf x})&=&{\bm \psi}({\bf x})^{\top}\boldsymbol{\Sigma}_\rho{\bm \Psi}^{\top} \left({\bm \Psi}\boldsymbol{\Sigma}_\rho{\bm \Psi}^\top+\sigma_e^2 {\bf I}_N\right)^{-1}\by,  \nonumber \\
\mbox{GP:} \quad  \widehat{f}({\bf x})&=&{\bm \psi}({\bf x})^{\top} \left({\bm \Psi}+\sigma_e^2 {\bf I}_N\right)^{-1}\by. 
\end{eqnarray}
	The relationships among the solutions $\widehat{f}({\bf x})$ of the main methods described in this work (RVM, GP and Q-GP) is graphically summarized  in Figure \ref{figmain_starting2}. Furthermore, Figure \ref{figmain_starting} provides some examples of the solution $\widehat{f}({\bf x})$ with $N=3$ data points.

 {\Remark RVMs and GPs  provide a complete description of the posterior-predictive distribution over the underlying function $f(\x)$. In both case, this posterior density is Gaussian.  The expected value of the posterior-predictive distribution is $\widehat{f}({\bf x})$ in Eq.~\eqref{aquiF}.  
}

 \noindent
 \newline
{\bf Localized nonlinearities.}
 \newline
     In this work,  we denote the nonlinearities {\it localized ``around'' the inputs} with the notation $\psi_n(\x,\x_n)$, with $n=1,\ldots,N$ (since they are localized around $\x_n$, we need exactly $N$ functions $\psi_n$). In some scenarios, we have the same functions translated in different regions of the space, i.e., $\psi_n(\x,\x_n)=\psi(\x,\x_n)$.
As an example of localized function, consider for instance the constant basis
 \begin{gather}
 \label{LOCaqui1}
 \psi({\bf x},{\bf x}_n)=\left\{
   \begin{split}
&1 \quad ||{\bf x}-{\bf x}_n||_p \leq \epsilon,\\
&0 \quad ||{\bf x}-{\bf x}_n||_p > \epsilon,
\end{split}
\right.
\end{gather}
where $\epsilon >0$ and $||{\bf z} ||_p=\left(\sum_{i=1}^{d_X} |z_i|^p\right)^{1/p}$ represents the $L_p$ vector norm. Other example of localized function is the following radial exponential function,
\begin{equation}
\label{LOCaqui2}
 \psi({\bf x},{\bf x}_n)=\exp\left(-\frac{||{\bf x}-{\bf x}_n||_p}{\lambda}\right).
\end{equation}
For simplicity, we have removed the subindex $n$ in $\psi$, however in RVM we can employ different types of nonlinear functions, for instance, combining constant basis  at some inputs and radial exponential functions at other inputs.  
The bases in Eqs.~\eqref{LOCaqui1}-\eqref{LOCaqui2} are also {\it isotropic} (or {\it homogeneous}) since  they depend only on the $L_p$ distance $r=||{\bf x}-{\bf x}_n||_p$, that is a scalar value \cite[Chapter 4]{rasmussen2003gaussian}. They are  also {\it stationary} kernels/bases.  A kernel function is stationary if satisfies the condition $\psi({\bf x},{\bf x}_n)=\psi({\bf x}-{\bf x}_n)$, i.e., it depends only on the difference vector ${\bf d}={\bf x}-{\bf x}_n$, but not on the values of the inputs, $\x$ and ${\bf x}_n$, themselves. Generally, a stationary kernel is an {\it anisotropic} kernel, since it depends on both the direction and the length of the difference vector ${\bf d}$. Clearly, an isotropic kernel is always a stationary kernel. 
 
{\Remark The spline models are special cases of GPs,  where the support of the bases is bounded (with a support smaller of the domain $\mathcal{X}$). In this scenario, the matrix ${\bm \Psi}$ is sparse and, in some scenarios, is a band matrix \cite[Chapter 6]{rasmussen2003gaussian}, \cite{Wahba90}.}


 \subsection{Posterior-predictive distribution}
 
 We consider two different probabilistic approaches which provide different regression models. 
   In the standard Bayesian derivation, the nonlinearities $\psi_n({\bf x},{\bf x}_n)$ play the role of basis functions. 
   In the Gaussian process (GP) approach, the nonlinearities $\psi_n({\bf x},{\bf x}_n)$ play the role of kernel functions specifying the correlation among different  pairs of inputs.
 A prior density over the underlying function $p(f(\x))$ is assumed (explicitly or implicitly) in both cases.
  Thus, in both cases, we obtain a complete description of a Gaussian posterior distribution of the hidden function in a generic test input ${\bf x}$, i.e.,
 \begin{align}
p(f({\bf x})|{\bf y})&= \frac{1}{p({\bf y})}p({\bf y}|f({\bf x})) p(f({\bf x})),
\end{align} 
where  $p({\bf y}|f({\bf x}))$ is the {\it likelihood function} (which is induced by Eq.~\eqref{EqNum1}), $p(f({\bf x}))$ represents the {\it prior density} over $f(\x)$ (which is given by the specific probabilistic approach), and  $p({\bf y})$ is the so-called {\it marginal likelihood}, useful for model selection (e.g., hyperparameter tuning). It is given by the expression 
$p({\bf y})=\int_{\mathcal{X}} p({\bf y}|f({\bf x})) p(f({\bf x})) df(\x)$. Below, we will derive the marginal likelihood for the different techniques (see  also Section \ref{Aqui_ML}).
\newline
\newline
 In the RVM and GP schemes, the posterior density is in both cases Gaussian, i.e., 
 \begin{align}
p(f({\bf x})|{\bf y})
&=\mathcal{N}(f({\bf x})|\widehat{f}({\bf x}),\sigma_{f|y}^2({\bf x})).
\end{align} 
  where the mean is the function $\widehat{f}({\bf x})$, i.e., $\mu_{f|y}({\bf x})=\widehat{f}({\bf x})$ in Eqs.~\eqref{aquiF} and \eqref{formulafinalaqui}. The final expressions of the coefficient vector $\widehat{{\bm \rho}}$ and of the variance $\sigma^2({\bf x})$ depend on the probabilistic derivation employed.\footnote{Note that a complete notation should be $p(f({\bf x})|{\bf y},\x_{1:N}, \mathcal{M})$, i.e., we consider all the training input points $\x_{1:N}=\{{\bf x}_n\}_{n=1}^N$ given and fixed, and with $\mathcal{M}$ we denote the bases $\psi_n$ and all the parameters of the model. In the rest of the work, for simplicity, we keep the simpler notation $p(f({\bf x})|{\bf y})=p(f({\bf x})|{\bf y},\x_{1:N}, \mathcal{M})$.} We will derive the variance for both techniques, obtaining 
  \begin{eqnarray}
\label{variancebothaqui}
\mbox{RVM:} \quad \sigma_{f|y}^2(\x)&=&{\bm \psi}({\bf x})^{\top}  \left(\frac{1}{\sigma_e^2}{\bm \Psi}^\top{\bm \Psi}+\boldsymbol{\Sigma}_\rho^{-1}\right)^{-1}{\bm \psi}({\bf x}),  \nonumber \\
\mbox{GP:} \quad  \sigma_{f|y}^2(\x)&=&\psi(\x,\x)-{\bm \psi}({\bf x})^{\top}({\bm \Psi} +\sigma_e^2 {\bf I}_N)^{-1}{\bm \psi}({\bf x}).
\end{eqnarray}

{\Remark The RVM and GP models consider the same likelihood function, since they assume the same observation model in Eq.~\eqref{EqNum1}. }
\newline
\newline 
The likelihood function is described below.  

 \subsubsection{Likelihood function}\label{sectLilke} 
 Given the observation model in Eq.~\eqref{EqNum1}, the induced likelihood function is given by 
 \begin{equation}
p(y_i|f({\bf x}_i)) =\mathcal{N}(y_i|f({\bf x}_i),\sigma_e^2).
\end{equation}
Furthermore, defining ${\bf f}=[f({\bf x}_1),\ldots,f({\bf x}_N)]^{\top}$ and considering conditional independence for the observations $y_i$, we also have
 \begin{equation}
 \label{likef1}
p({\bf y}|{\bf f}) =\prod_{i=1}^N p(y_i| f({\bf x}_i))=\mathcal{N}({\bf y}|{\bf f},\sigma_e^2 {\bf I}_N),
\end{equation}
where ${\bf I}_N$ is an $N\times N$ identity matrix.  Depending on the employed probabilistic approach (see below), one can assume that $f({\bf x})$ has exactly the form in Eq.~\eqref{aquiF} or Eq.~\eqref{aquiF2}, i.e., $f({\bf x})={\bm \psi}({\bf x}_i)^{\top} {\bm \rho}$, so that the observation model can be written as
 \begin{equation}
 \label{aquiEqPsiVect}
 y_i= {\bm \psi}({\bf x}_i)^{\top} {\bm \rho}+e_i.
\end{equation}
Considering that  the nonlinearities are known and chosen by the user, the likelihood of a single observations with respect to the weights is
 \begin{equation}
p(y_i| {\bm \rho}) =\mathcal{N}(y_i|{\bm \psi}({\bf x}_i)^{\top} {\bm \rho},\sigma_e^2).
\end{equation}
The complete likelihood function with respect to the coefficients is 
 \begin{equation}
 \label{CompLikeEq}
p({\bf y}| {\bm \rho})=\prod_{i=1}^N p(y_i| {\bm \rho})=\mathcal{N}({\bf y}|{\bm \Psi} {\bm \rho},\sigma_e^2 {\bf I}_N),
\end{equation}
and can be obtained from Eq.~\eqref{likef1} setting $\f={\bm \Psi} {\bm \rho}$.

 \subsection{Smoothing and prediction}

\begin{figure}[!h]
	\centering
\centerline{	
	\subfigure[]{\includegraphics[width=8cm]{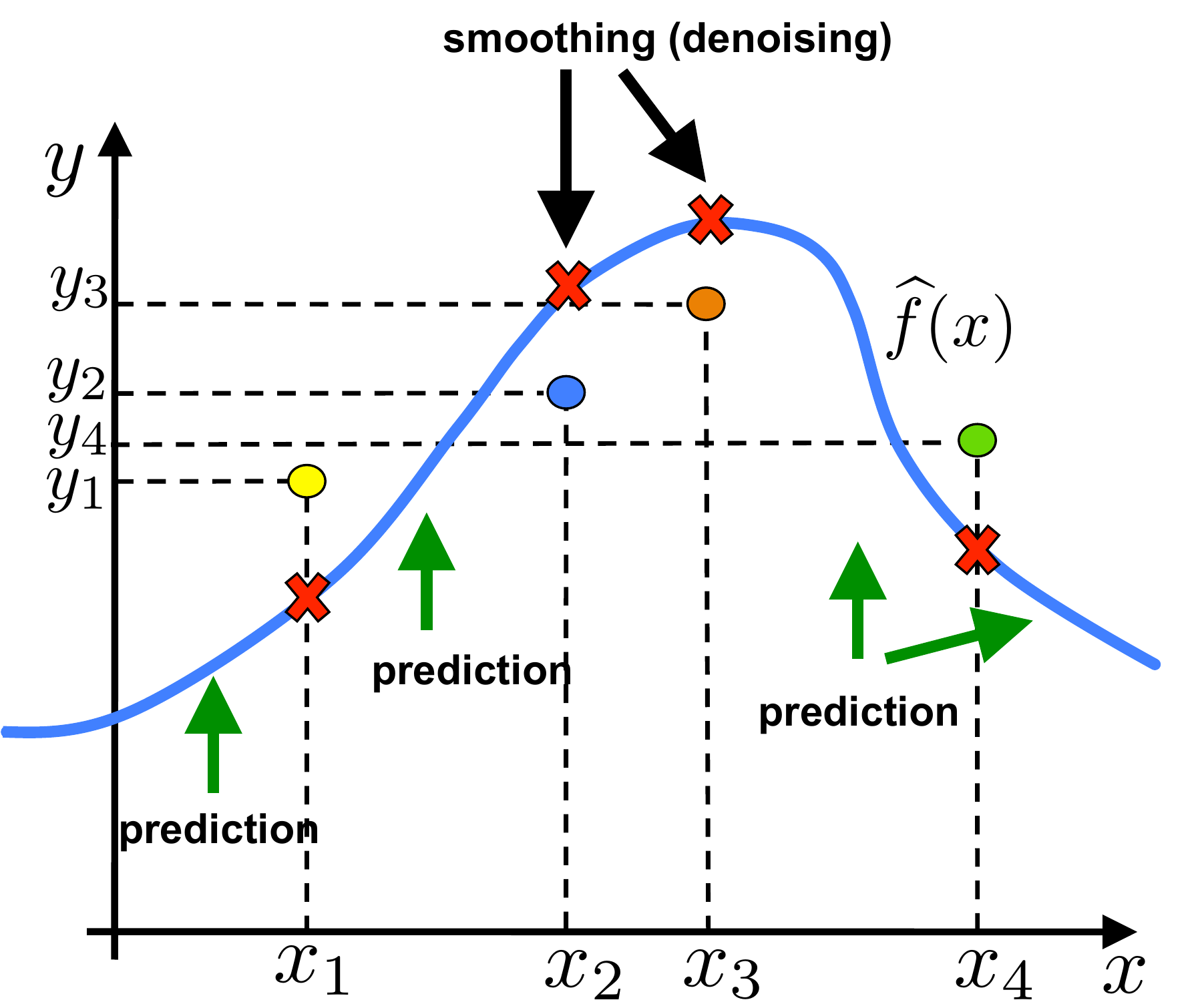}}
	\subfigure[]{\includegraphics[width=8cm]{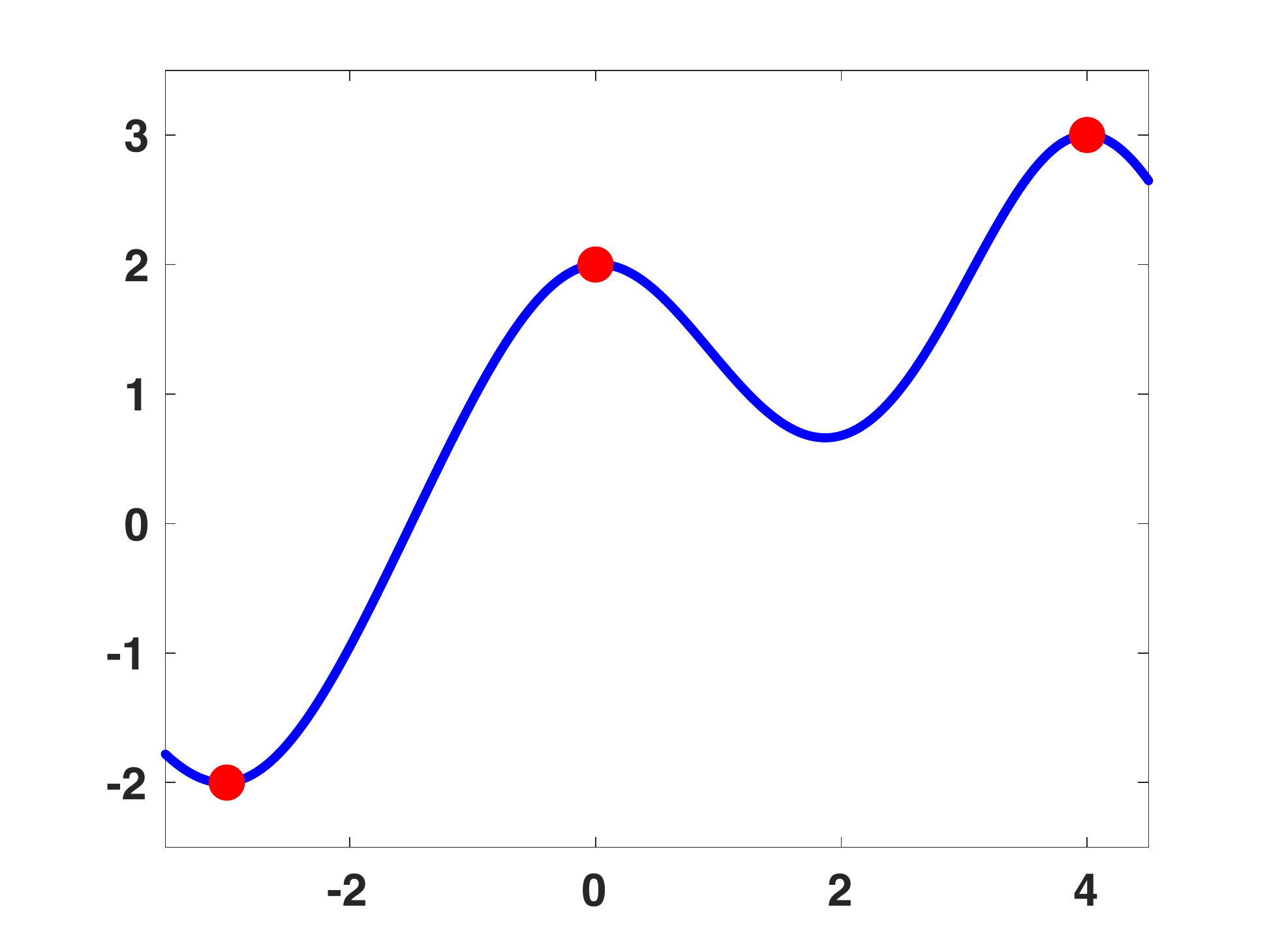}}
}
	\caption{{\bf (a)} Graphical representation of the differences between smoothing and prediction problems. The red crosses represent the solutions of the smoothing problem. The solid blue line represents the prediction at different inputs $\x^*$ which may not belong to the training set. Prediction and smoothing jointly form a complete regression problem. {\bf (b)} Examples of RVM and GP solutions $\widehat{f}(\x)$ for interpolation (with $N=3$ data points). }
	\label{figSmoothPred}
\end{figure}

 
\noindent
{\bf Smoothing.} We refer to smoothing problem when one is interested only to obtain the  estimation values $\widehat{f}({\bf x}_1),\ldots,\widehat{f}({\bf x}_N)$, i.e., to know the estimations only at the training inputs, $\x_{1:N}=\{{\bf x}_n\}_{n=1}^N$. In this scenario, Eq.~\eqref{EqNum1} can be expressed in the following vectorial form
 \begin{align}
  \label{Ynowtodos}
{\bf y}&={\bf f}+{\bf e}={\bm \Psi} {\bm \rho}+{\bf e},
\end{align}
where ${\bf f}=[f({\bf x}_1),\ldots,f({\bf x}_N)]^{\top}$ and ${\bf e}=[e_1,\ldots,e_N]^{\top} \sim \mathcal{N}({\bf 0},\sigma_e^2 {\bf I}_N)$ with ${\bf I}_N$ is an $N\times N$ unit matrix. The goal of the smoothing problem is to obtain a vector ${\bf \widehat{f}}$ that approximates ${\bf f}$.  This is also known as {\it denoising}.
\newline
\newline
{\bf Prediction.} We refer to a prediction problem if we consider the approximation of the underlying function $f({\bf x})$ at some ${\bf x}$ which is not contained in $\{{\bf x}_n\}_{n=1}^N$. Namely, the goal in prediction  is to infer the value $f(\x^*)$ at some test point $\x^*$ that is not contained in the training set, i.e., $\x^* \notin \{\x_1,\ldots,\x_N\}$. This is also referred as {\it extrapolation}. Some regression methods can differ in prediction but provide the same results in smoothing, for instance. We show some examples below. 
Figure \ref{figSmoothPred}-(a) provides a graphical representation of the differences between smoothing and prediction problems. 

{\Remark The regression problem can be considered the union of the two important sub-problems,  smoothing and prediction. }


\subsection{Interpolation}\label{GPinterp}

If we force the conditions $\widehat{f}(\x_n)=y_n$ as shown in Figure \ref{figSmoothPred}-(b)  (i.e., perfect fitting with the data, a.k.a., interpolation), RVMs and GPs provide the same solution in terms of mean of the posterior $\widehat{f}(\x)$ (but different predictive variances). Let us consider an interpolating function of the form
\begin{equation}
\label{aquiF_GPinter}
\widehat{f}({\bf x})=\sum_{i=1}^N \widehat{\rho}_i \psi_n({\bf x},{\bf x}_i)={\bm \psi}({\bf x})^{\top} \widehat{{\bm \rho}},
 \end{equation}
i.e., a linear combination of the nonlinearities $\psi_i({\bf x},{\bf x}_i)$. We would like that $\widehat{f}(\x_n)=y_n$ for all $n=1,\ldots,N$. Therefore, in order to obtain the proper coefficients $\widehat{\rho}_i$, we can write a $N\times N$ linear system of  $N$ conditions of passing  through the points $(\x_n,y_n)$, 
\begin{gather}\label{SystemInterp}
\left\{
\begin{split}
&  \widehat{\rho}_1 \psi_1({\bf x}_1,{\bf x}_1)+  \widehat{\rho}_2 \psi_2({\bf x}_1,{\bf x}_2)+\ldots.+   \widehat{\rho}_N \psi_N({\bf x}_1,{\bf x}_N)=y_1,\\
&  \widehat{\rho}_1 \psi_1({\bf x}_2,{\bf x}_1)+  \widehat{\rho}_2 \psi_2({\bf x}_2,{\bf x}_2)+\ldots.+   \widehat{\rho}_N \psi_N({\bf x}_2,{\bf x}_N)=y_2,\\
& \vdots \\
&  \widehat{\rho}_1 \psi_1({\bf x}_N,{\bf x}_1)+  \widehat{\rho}_2 \psi_2({\bf x}_N,{\bf x}_2)+\ldots.+   \widehat{\rho}_N \psi_N({\bf x}_N,{\bf x}_N)=y_N,
\end{split}
\right.
\end{gather}
i.e., in matrix form ${\bm \Psi} \widehat{{\bm \rho}}=\y$. 
If ${\bm \Psi}$ is invertible, then we get
\begin{equation}
\widehat{{\bm \rho}}=[\rho_1,\ldots,\rho_N]^{\top} ={\bm \Psi}^{-1} \y. 
\end{equation}
Thus, the interpolative function of both methods can be expressed as 
\begin{equation}
\label{aquiF_GPinter_2}
\widehat{f}({\bf x})={\bm \psi}({\bf x})^{\top} \widehat{{\bm \rho}}= {\bm \psi}({\bf x})^{\top}{\bm \Psi}^{-1} \y.
 \end{equation}
 Therefore, by definition we have $\widehat{f}({\bf x}_n)=y_n$ (i.e., $\widehat{f}({\bf x})$ is an interpolator).
\newline
In the following sections, we present the derivations of RVM, GP and Quasi-GP methods. Then, we discuss the connections among them.

\section{Relevance Vector Machine (RVM)}\label{RVMsect}

Following a standard Bayesian approach, we  consider a Gaussian prior density over the weights 
  ${\bm \rho}=[\rho_1,\ldots,\rho_N]^{\top}$, i.e.,
\begin{eqnarray}
p({\bm \rho}) =\mathcal{N}({\bm \rho}|{\bf 0}, {\bm \Sigma}_\rho),
\end{eqnarray}
 where  ${\bm \Sigma}_\rho$ is an $N\times N$ matrix.
 Thus, observing Eq.~\eqref{Ynowtodos}, i.e., ${\bf y}={\bm \Psi} {\bm \rho}+{\bf e}$, we can see that the vector ${\bf y}$ is the sum of two independent multivariate Gaussian variables, one with zero mean and covariance matrix ${\bm \Psi} {\bm \Sigma}_\rho{\bm \Psi}^{\top}$ and the other one with   zero mean and covariance matrix $\sigma_e^2 {\bf I}_N$. The sum of two independent Gaussian variables is itself a Gaussian variable with mean the sum of the means, and covariance matrix the sum of the covariance matrices, i.e., the marginal likelihood is 
 \begin{equation}
p({\bf y}) =\mathcal{N}({\bf y}|{\bf 0}, {\bm \Psi} {\bm \Sigma}_\rho{\bm \Psi}^{\top}+\sigma_e^2 {\bf I}_N).
\end{equation}

\subsection{Posterior and induced prior distributions of RVM}

As we have done with the likelihood functions in Section \ref{sectLilke}, in this section we describe posterior distributions of ${\bm \rho}$ and $f(\x)$. Moreover, we derive the induced prior density over $f(\x)$.
\subsubsection{Posterior of the weights ${\bm \rho}$} \label{postofWEights}
Recalling that the likelihood $p({\bf y}| {\bm \rho})=\mathcal{N}({\bf y}|{\bm \Psi} {\bm \rho},\sigma_e^2 {\bf I}_N)$ is Gaussian,
the posterior density of the weights is thus proportional to the product of two Gaussians, $p({\bm \rho}|{\bf y})  \propto p({\bf y}| {\bm \rho})p({\bm \rho})$, and therefore it is also Gaussian: 
 \begin{eqnarray}
p({\bm \rho}|{\bf y}) = \frac{1}{p({\bf y})}p({\bf y}| {\bm \rho})p({\bm \rho}) 
= \mathcal{N}({\bm \rho}|{\bm \mu}_{\rho|y}, {\bm \Sigma}_{\rho|y}),   
\end{eqnarray}
After some algebra, the mean of the posterior ${\bm \mu}_{\rho|y}=\widehat{{\bm \rho}}$  can be expressed in different ways,  
\begin{eqnarray}
 {\bm \mu}_{\rho|y}=\widehat{{\bm \rho}}&=&\frac{1}{\sigma_e^2}\left(\frac{1}{\sigma_e^2}{\bm \Psi}^\top{\bm \Psi}+\boldsymbol{\Sigma}_{\rho}^{-1}\right)^{-1}{\bm \Psi}^\top\by,   \label{Mean_Post_rho0}  \\
&=&\left({\bm \Psi}^\top{\bm \Psi}+\sigma_e^2\boldsymbol{\Sigma}_\rho^{-1}\right)^{-1}{\bm \Psi}^\top\by,  \label{Mean_Post_rho} \\
&=& \boldsymbol{\Sigma}_\rho{\bm \Psi}^{\top} \left({\bm \Psi}\boldsymbol{\Sigma}_\rho{\bm \Psi}^\top+\sigma_e^2 {\bf I}_N\right)^{-1}\by, \label{Mean_Post_rho_last}
\end{eqnarray}
 and likewise, the covariance matrix can be written variously as
\begin{eqnarray}
{\bm \Sigma}_{\rho|y}&=&\left(\frac{1}{\sigma_e^2}{\bm \Psi}^\top{\bm \Psi}+\boldsymbol{\Sigma}_\rho^{-1}\right)^{-1},\label{Var_Post_rho}  \\
&=& \sigma_e^2\left({\bm \Psi}^\top{\bm \Psi}+\sigma_e^2\boldsymbol{\Sigma}_\rho^{-1}\right)^{-1},\\
&=&{\bm \Sigma}_\rho-{\bm \Sigma}_\rho{\bm \Psi}^{\top}\left({\bm \Psi}{\bm \Sigma}_\rho{\bm \Psi}^{\top}+\sigma_e^2{\bf I}_N\right)^{-1}{\bm \Psi}{\bm \Sigma}_\rho.
\end{eqnarray}
 See \cite{rasmussen2003gaussian,bishop2006pattern} and \ref{RVMvarianceAPP} for additional details regarding the last equality. 
Note also that we may substitute ${\bf S}^{-1}=\sigma_e^2\boldsymbol{\Sigma}_\rho^{-1}$  
into \eqref{Mean_Post_rho}. 
where $\mS$ can be interpreted as an inverse of a ``signal-to-noise ratio'' (SNR), where $\sigma_e^2$ is the noise power and the covariance of the prior $\boldsymbol{\Sigma}_\rho$ plays the role of ``power of the signal''.

\subsubsection{Posterior of the function: predictive distribution} 
 The posterior of $f(\x)$ in a generic $\x \in \mathcal{X}$, is also Gaussian, 
 $$
 p(f(\x) | {\bf y}) =\mathcal{N}\left(f(\x) \mid \mu_{f|y}(\x), \sigma_{f|y}^2(\x)\right),
 $$ 
 with  
\begin{gather} \label{RVMsolMean}
\begin{split}
\mu_{f|y}(\x)= \widehat{f}(\x)&={\bm \psi}({\bf x})^{\top}  \widehat{{\bm \rho}}\\ 
&={\bm \psi}({\bf x})^{\top}\left({\bm \Psi}^\top{\bm \Psi}+\sigma_e^2\boldsymbol{\Sigma}_\rho^{-1}\right)^{-1}{\bm \Psi}^\top\by, \\
&={\bm \psi}({\bf x})^{\top}\boldsymbol{\Sigma}_\rho{\bm \Psi}^{\top} \left({\bm \Psi}\boldsymbol{\Sigma}_\rho{\bm \Psi}^\top+\sigma_e^2 {\bf I}_N\right)^{-1}\by,  
\end{split}
\end{gather}
where have replaced the two possible expressions of $ \widehat{{\bm \rho}}$ in Eqs.~\eqref{Mean_Post_rho}--\eqref{Mean_Post_rho_last}, and 
\begin{eqnarray}
 \sigma_{f|y}^2(\x)&=&{\bm \psi}({\bf x})^{\top}  {\bm \Sigma}_{\rho|y}{\bm \psi}({\bf x})={\bm \psi}({\bf x})^{\top}  \left(\frac{1}{\sigma_e^2}{\bm \Psi}^\top{\bm \Psi}+\boldsymbol{\Sigma}_\rho^{-1}\right)^{-1}{\bm \psi}({\bf x}), \label{Eq41var} \\
 &=&{\bm \psi}({\bf x})^{\top}{\bm \Sigma}_\rho{\bm \psi}({\bf x})- {\bm \psi}({\bf x})^{\top}{\bm \Sigma}_\rho{\bm \Psi}^{\top}\left({\bm \Psi}{\bm \Sigma}_\rho{\bm \Psi}^{\top}+\sigma_e^2{\bf I}_N\right)^{-1}{\bm \Psi}{\bm \Sigma}_\rho{\bm \psi}({\bf x}). \nonumber
\end{eqnarray}
where we recall that ${\bm \psi}({\bf x})$ is an $N\times 1$ dimensional vector. For the last equality, see \ref{RVMvarianceAPP}.

\subsubsection{Interpolation with RVM}\label{InterpRVM}
 Considering \eqref{Mean_Post_rho}, if we have noisy-free observations $\sigma_e^2=0$, we obtain the following expression 
\begin{align}
\widehat{{\bm \rho}}&= \left({\bm \Psi}^\top{\bm \Psi}\right)^{-1}{\bm \Psi}^\top\by, \nonumberÊ\\
&={\bm \Psi}^{-1}\by, \label{interpolation_RHO_RVM}
\end{align}
and the resulting mean function 
$$
\widehat{f}(\x)={\bm \psi}({\bf x})^{\top} \widehat{{\bm \rho}}={\bm \psi}({\bf x})^{\top}{\bm \Psi}^{-1}\by,
$$ 
is an interpolant, satisfying the passing conditions $\widehat{f}(\x_n)=y_n$, for all $n=1,\ldots,N$, as described in Section \ref{GPinterp} and the linear system given in Eqs.~\eqref{SystemInterp}.  Let us begin with the simple case  $\boldsymbol{\Sigma}_\rho=\sigma_\rho^2 {\bf I}_N$. Then, we can define $\mbox{SNR}=\frac{\sigma_\rho^2}{\sigma_e^2}$ and ${\bf S}=\mbox{SNR} \cdot {\bf I}_N$.
Note that if $\mbox{SNR}= \infty$, i.e., if we have noise-free observations $\sigma_e^2=0$ or an uninformative prior $\sigma_\rho^2=\infty$,  in both cases we obtain Eq.~\eqref{interpolation_RHO_RVM}.
 
 {\Remark With RVM, we can obtain the interpolative solution $\widehat{{\bm \rho}}={\bm \Psi}^{-1}\by$ in Section \ref{GPinterp},  either with $\sigma_e^2=0$ (and a finite $\sigma_\rho^2$) or using  an uninformative prior over the weights, $\sigma_\rho^2=\infty$ (and $\sigma_e^2\neq 0$).
  }
 \newline
 In the opposite scenario,  with $\mbox{SNR}=0$ (if $\sigma_e^2=\infty$ or $\sigma_\rho^2=0$),  we have $\widehat{{\bm \rho}}={\bf 0}$.  Regarding the predictive variance $ \sigma_{f|y}^2(\x)$, when $\sigma_e^2=0$ we obtain 
\begin{eqnarray}
 \sigma_{f|y}^2(\x)&=&{\bm \psi}({\bf x})^{\top}{\bm \Sigma}_\rho{\bm \psi}({\bf x})- {\bm \psi}({\bf x})^{\top}{\bm \Sigma}_\rho{\bm \Psi}^{\top}\left({\bm \Psi}{\bm \Sigma}_\rho{\bm \Psi}^{\top}\right)^{-1}{\bm \Psi}{\bm \Sigma}_\rho{\bm \psi}({\bf x}). \nonumber \\
 &=& {\bm \psi}({\bf x})^{\top}{\bm \Sigma}_\rho{\bm \psi}({\bf x})- {\bm \psi}({\bf x})^{\top}{\bm \Sigma}_\rho{\bm \psi}({\bf x})= 0,
\end{eqnarray}
where we have used $\left({\bm \Psi}{\bm \Sigma}_\rho{\bm \Psi}^{\top}\right)^{-1}=\left({\bm \Psi}^{\top}\right)^{-1} \left({\bm \Psi}{\bm \Sigma}_\rho\right)^{-1}$.
Namely, in the interpolation scenario, the predictive variance of RVM  is zero, i.e., $ \sigma_{f|y}^2(\x)=0$,  for all $\x\in \mathcal{X}$.
\subsubsection{Posterior density for the smoothing problem}

Considering the smoothing problem, the posterior of the vector  ${\bf f}={\bm \Psi}{\bm \rho}$ is a multivariate Gaussian pdf,
$p({\bf f}|{\bf y}) =\mathcal{N}({\bf f}| {\bm \mu}_{f|y}, {\bm \Sigma}_{f|y})$, where the mean vector is
\begin{eqnarray}
{\bm \mu}_{f|y}&=&\widehat{\f}={\bm \Psi} {\bm \mu}_{\rho|y} \nonumber \\
 &=& {\bm \Psi}\left({\bm \Psi}^\top{\bm \Psi}+\sigma_e^2\boldsymbol{\Sigma}_\rho^{-1}\right)^{-1}{\bm \Psi}^\top\by  \nonumber  \\
 &=&{\bm \Psi}\boldsymbol{\Sigma}_\rho{\bm \Psi}^\top \left({\bm \Psi}\boldsymbol{\Sigma}_\rho{\bm \Psi}^\top+\sigma_e^2 {\bf I}_N\right)^{-1}\by, \label{MuSmoothRVM}
 \end{eqnarray}
 and  covariance matrix is given by
 \begin{eqnarray}
{\bm \Sigma}_{f|y}&=&{\bm \Psi}{\bm \Sigma}_{\rho|y} {\bm \Psi}^{\top},  \nonumber  \\
 &=&{\bm \Psi}\left(\frac{1}{\sigma_e^2}{\bm \Psi}^\top{\bm \Psi}+\boldsymbol{\Sigma}_\rho^{-1}\right)^{-1}{\bm \Psi}^{\top},  \nonumber \\
 &=&\left[\left({\bm \Psi}{\bm \Sigma}_\rho{\bm \Psi}^{\top}  \right)^{-1}+\left(\sigma_e^2 {\bf I}_N \right)^{-1}\right]^{-1},  \nonumber   \\
 &=&{\bm \Psi}{\bm \Sigma}_\rho{\bm \Psi}^{\top}- {\bm \Psi}{\bm \Sigma}_\rho{\bm \Psi}^{\top}\left(\sigma_e^2{\bf I}_N+{\bm \Psi}{\bm \Sigma}_\rho{\bm \Psi}^{\top}\right)^{-1}{\bm \Psi}{\bm \Sigma}_\rho{\bm \Psi}^{\top}.  \label{VarSmoothRVM}
  \end{eqnarray} 
For more details, see \ref{RVMvarianceAPP}.

\subsubsection{Induced prior density over the underlying function}
  Given a test input $\x$ and the vector ${\bm \psi}({\bf x})$ (choosing and fixing the bases) and considering the random variable $f(\x)={\bm \psi}({\bf x})^{\top} {\bm \rho}$ (a scalar value), we can observe that 
\begin{align}\label{RVMprior1}
&p(f(\x))=\mathcal{N}(f(\x)|\mu_f(\x), \sigma_f^2(\x)), \quad \mbox{ with } \nonumber \\  
&\mu_f(\x)=0, \quad  \sigma_f^2(\x)={\bm \psi}({\bf x})^{\top} \boldsymbol{\Sigma}_\rho {\bm \psi}({\bf x}). 
\end{align}
Therefore, in this probabilistic approach, we directly impose a prior over the weights ${\bm \rho}$, and we also induce a prior density over the function $f(\x)$.   Clearly, if we consider the $N\times 1$ vector ${\bf f}={\bm \Psi}{\bm \rho}$, we have that 
\begin{equation}\label{RVMprior2}
p(\f)=\mathcal{N}(\f|{\bm \mu}_f,{\bm \Sigma}_f), \quad \mbox{ with } \quad  {\bm \mu}_f={\bf 0}, \quad {\bm \Sigma}_f={\bm \Psi} \boldsymbol{\Sigma}_\rho {\bm \Psi}^{\top}.
\end{equation}

\subsubsection{Why it is called a Relevance Vector Machine}
 Let us consider a prior covariance matrix over the weights of type
\begin{equation}
{\bm \Sigma}_\rho=
  \begin{bmatrix}
    1/\alpha_1 & 0 & 0 &  &0 \\
    0 & 1/\alpha_2 & 0 &\ldots & 0 \\
    0 & 0  & 1/\alpha_3 &\ldots & 0 \\
   \ldots & \ldots  & \ldots &\ldots & \ldots \\ 
    0 & 0  & 0 &\ldots & 1/\alpha_N 
  \end{bmatrix}
\end{equation}
i.e.,  ${\bm \Sigma}_\rho$ is diagonal with elements $[1/\alpha_1,1/\alpha_2, \ldots,1/\alpha_N]$ in its diagonal.
The idea is to use a hierarchical approach considering  that  also the hyper-parameters $\alpha_i$  are unknown coefficients to be learned. 
 As an example of learning procedure, we can maximize the marginal likelihood with respect to these hyper-parameters.
It is possible to show that a significant proportion of the $\{\alpha_i\}$ diverge to infinity. As a consequence,   the mean of the posterior in Eq.~\eqref{Mean_Post_rho} of the weights $\{\rho_i\}$  corresponding to these ``divergent'' $\{\alpha_i\}$  is close to zero (with negligible variance). Hence, the basis functions $\psi_i$ associated with these weights $\rho_i$ are virtually pruned out, and the function $\widehat{f}(\x)$ depends only on a few bases. Then, the result is a {\it sparse} model. As the bases are localized around particular training inputs $\x_i$, this learning procedure can be also interpreted as a way of selecting {\it relevant} inputs. 
Therefore, RVM can be considered a Bayesian sparse kernel
technique.  RVM typically leads
to much sparser models than the well-known Support Vector Machine (SVM), resulting in correspondingly faster performance on test data \cite[Section 7.2, page 345]{bishop2006pattern}. However, the training of the hyper-parameters is often slower than SVM.


\subsection{Random functions according to RVM models}\label{RVMgeneration}
\begin{figure}[!h]
	\centering
\centerline{	
	\subfigure[]{\includegraphics[width=7cm]{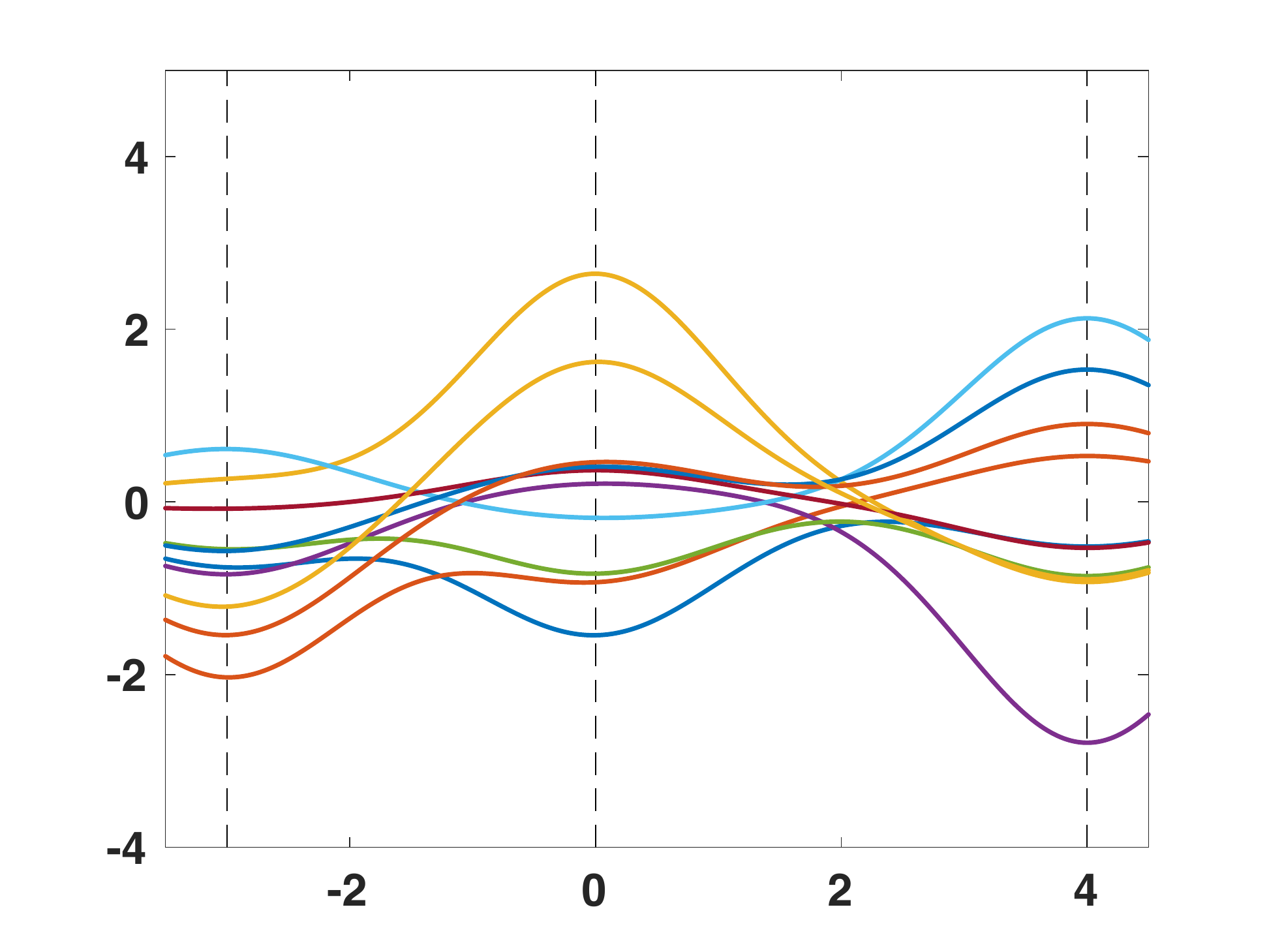}}
	\hspace{-0.9cm}
	\subfigure[]{\includegraphics[width=7cm]{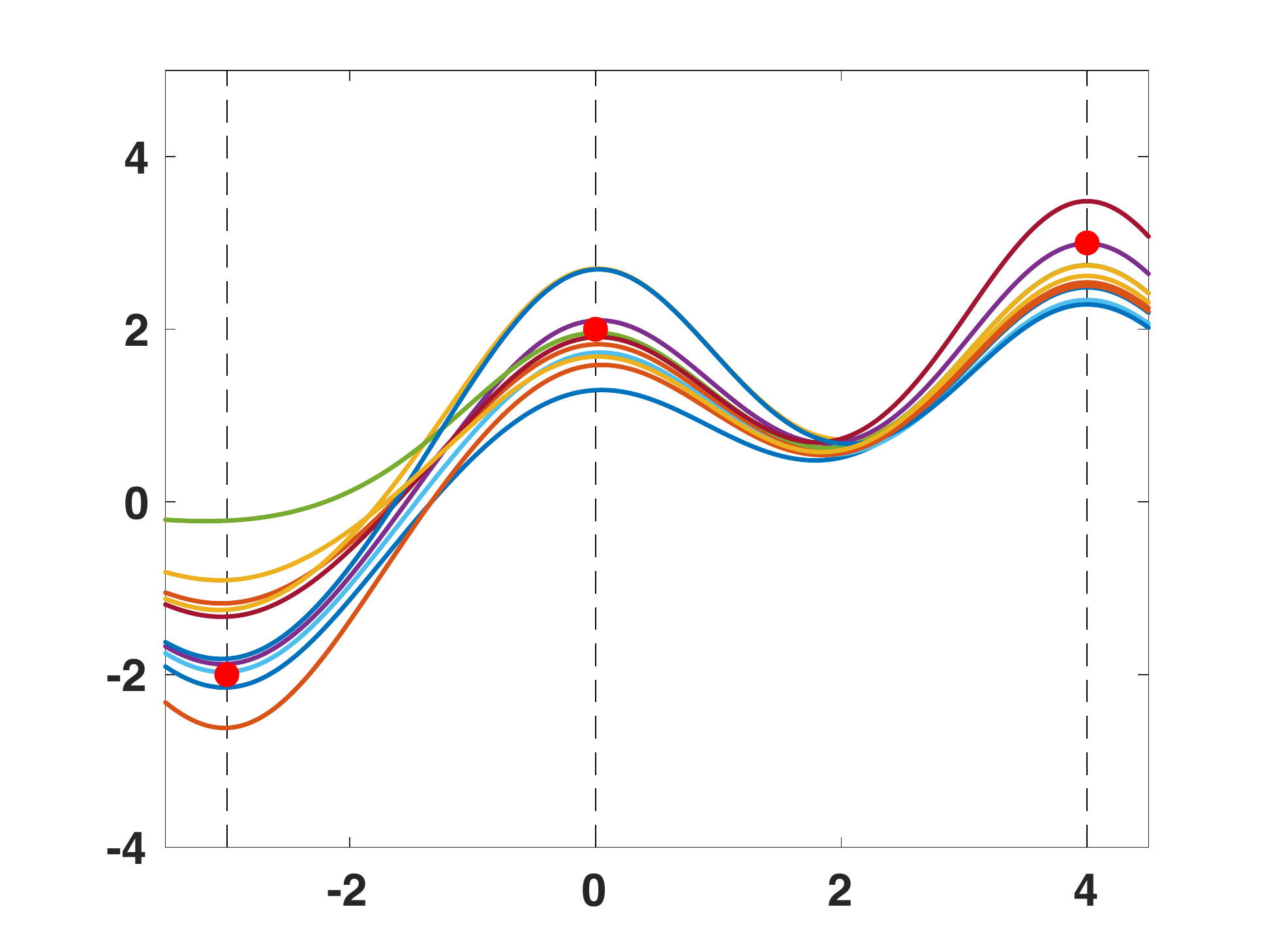}}
	\hspace{-0.9cm}	
	\subfigure[]{\includegraphics[width=7cm]{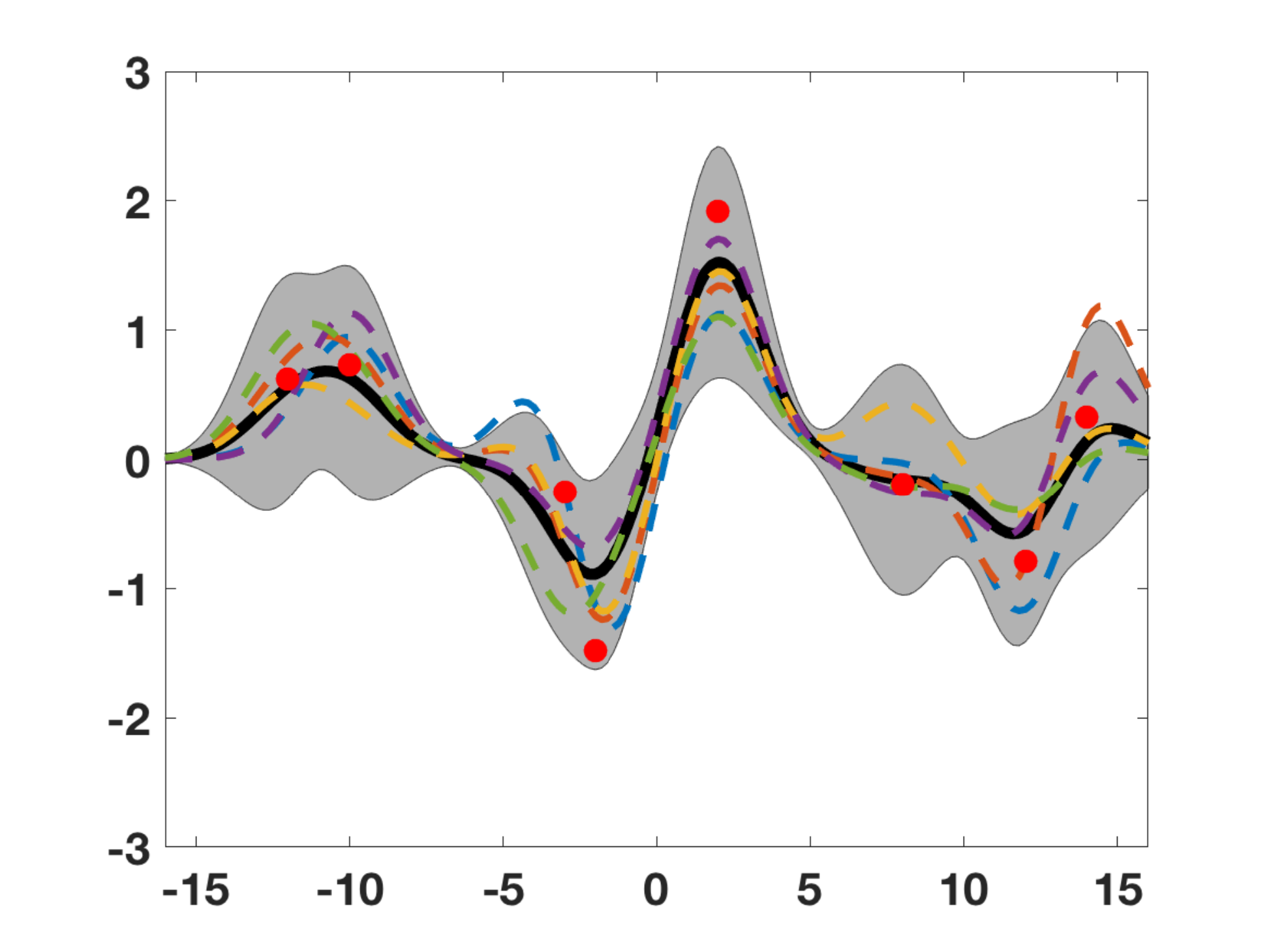}}
}
	\caption{Random functions $\widehat{f}^{(s)}(\x)$ with $s=1,\ldots,10$  {\bf (a)} from a RVM prior over $f(\x)$ and {\bf (b)} from a RVM posterior (after knowing the $N=3$ data points), with $N=3$ Gaussian bases with the mean location depicted by the dashed lines (and bandwidth $\lambda=2$). We have considered a diagonal covariance matrix ${\bm \Sigma}_\rho$ with all the elements in the diagonal equal to $2.25$. {\bf (c)} Example of RVM mean and variance with $N=8$ data points and $S=5$ random functions from the posterior depicted with dashed lines ($\sigma_e=0.5$, $\lambda=4$,  ${\bm \Sigma}_\rho$ diagonal  all the elements equal to $1$). The black solid line shows the mean $ \mu_{f|y}(\x)=\widehat{f}(\x)$ and the boundary of the grey area corresponds to $\widehat{f}(\x)\pm2 \sigma_{f|y}^2(\x)$ (i.e., $\approx 95\%$ of the probability).}
	\label{figRandomRVM}
\end{figure}

 Note that (also in this approach) random functions can be generated from the prior and posterior densities. Indeed, we show different generating procedures in order to draw from the prior and posterior pdf $f(\x)$.
  \newline
 \newline
 {\bf Draw functions from the prior.}  Let us consider the following procedure for generating $S$ random functions from the prior pdf:
\newline
For $s=1,\ldots,S$:
\begin{enumerate}
\item Draw a vector ${\bm \rho}^{(s)}=[ \rho_1^{(s)},\ldots,\rho_N^{(s)}]^{\top} \sim \mathcal{N}({\bm \rho}|{\bf 0}, {\bm \Sigma}_\rho)$.
\item Then, set
\begin{eqnarray}\label{EqotraVezSiempre}
f^{(s)}({\bf x})=\sum_{n=1}^N \rho_n^{(s)} \psi_n({\bf x},{\bf x}_n), \qquad  \forall  {\bf x}\in \mathcal{X}.
\end{eqnarray}
\end{enumerate}
Note that the procedure above takes into account the correlation (among different $\x$) induced by the model assumptions. See below and Section \ref{SectDualK} for further details regarding the induced correlation. 
 \newline
  \newline
 {\bf Draw functions from the posterior.} In order to draw functions from the posterior pdf, we can use the following steps:
\newline
For $s=1,\ldots,S$:
\begin{enumerate}
\item Draw a vector ${\bm \rho}^{(s)}=[ \rho_1^{(s)},\ldots,\rho_N^{(s)}]^{\top} \sim \mathcal{N}({\bm \rho}|{\bf \vmu}_{\rho|y}, {\bm \Sigma}_{\rho|y})$ where mean and variance are given in Eqs.~\eqref{Mean_Post_rho}--\eqref{Var_Post_rho}.
\item Then, set
\begin{eqnarray}
f^{(s)}({\bf x})=\sum_{n=1}^N \rho_n^{(s)} \psi_n({\bf x},{\bf x}_n), \qquad  \forall  {\bf x}\in \mathcal{X}.
\end{eqnarray}
\end{enumerate}
Thus, the generation of random function from the prior and posterior density is also possible in RVMs. Hence, this possibility is not only a prerogative of the Gaussian Process (GP) approach in Section \ref{GPsect}, as is often hinted in the literature. See Figure \ref{figRandomRVM} for some example of random functions drawn from a RVM model.
\newline
\newline
 {\bf Alternative sampling schemes.} We describe two alternative procedures from drawing from the RVM prior and posterior equivalent to the procedures above. For instance, in order to draw from the RVM prior, we can consider the Eqs.~\eqref{RVMprior1}-\eqref{RVMprior2}. Let us consider $P$ test points $\x^{(1)},\ldots,\x^{(P)}$.
Then, the following procedure generates $S$ random functions from a RVM  prior:
\begin{enumerate}
\item Compute  the $N\times P$ matrix ${\bf V}=[{\bm \psi}({\bf x}^{(1)}),\ldots, {\bm \psi}({\bf x}^{(P)})]$. Recall that ${\bm \psi}({\bf x})=[\psi_1({\bf x},{\bf x}_1),\ldots,\psi_N({\bf x},{\bf x}_N)]^{\top}$ is the $N\times 1$ design vector.
\item Compute  the  $P\times P$ covariance matrix of the vector 
 ${\bf f}_P=[f(\x^{(1)}),\ldots,f(\x^{(P)})]^{\top}$, i.e.,
$$
{\bf C}={\bf V} \boldsymbol{\Sigma}_\rho {\bf V}^{\top}.
$$
\item Draw  $S$  vectors ${\bf f}_P^{(s)}=[f^{(s)}(\x^{(1)}),\ldots,f^{(s)}(\x^{(P)})]^{\top}$ from a multivariate Gaussian, i.e., 
$$
{\bf f}_P^{(s)}\sim\mathcal{N}({\bf f}_P|{\bf 0},{\bf C}), \quad s=1,\ldots,S,
$$
 where ${\bf 0}$ is a $P\times 1$ null vector and ${\bf C}$ is given above.  
\end{enumerate}
In the same fashion, considering again $P$ test inputs $\x^{(1)},\ldots,\x^{(P)}$ and the Eqs.~\eqref{RVMsolMean}--\eqref{Eq41var} and Eqs.~\eqref{MuSmoothRVM}--\eqref{VarSmoothRVM}, we can consider the following procedure for sampling from the RVM posterior:
 \begin{enumerate}
\item Compute  the $N\times P$ matrix ${\bf V}=[{\bm \psi}({\bf x}^{(1)}),\ldots., {\bm \psi}({\bf x}^{(P)})]$. Recall that ${\bm \psi}({\bf x})=[\psi_1({\bf x},{\bf x}_1),\ldots,\psi_N({\bf x},{\bf x}_N)]^{\top}$ is the $N\times 1$ design vector.
\item Compute  the  $P\times P$ covariance matrix of the vector ${\bf f}_P=[f(\x^{(1)}),\ldots,f(\x^{(P)})]^{\top}$,
$$
{\bf C}={\bf V} \boldsymbol{\Sigma}_\rho {\bf V}^{\top}.
$$
\item Draw  $S$  vectors ${\bf f}_P^{(s)}=[f^{(s)}(\x^{(1)}),\ldots,f^{(s)}(\x^{(P)})]^{\top}$ from a multivariate Gaussian, i.e., 
$$
{\bf f}_P^{(s)}\sim\mathcal{N}({\bf f}_P|{\bm \mu}, {\bf \Sigma}), \quad s=1,\ldots,S,
$$
 where
  \begin{align*}
 {\bm \mu}&={\bf V}\boldsymbol{\Sigma}_\rho{\bm \Psi}^\top \left({\bm \Psi}\boldsymbol{\Sigma}_\rho{\bm \Psi}^\top+\sigma_e^2 {\bf I}_N\right)^{-1}\by, \qquad\qquad\qquad \mbox{is a $P\times 1$ vector and}, \\
 {\bf \Sigma}
&= {\bf C}- {\bf V}{\bm \Sigma}_\rho{\bm \Psi}^{\top}\left(\sigma_e^2{\bf I}_N+{\bm \Psi}{\bm \Sigma}_\rho{\bm \Psi}^{\top}\right)^{-1}{\bm \Psi}{\bm \Sigma}_\rho{\bf V}^{\top}, \quad \mbox{is a $P\times P$ covariance matrix}.
 \end{align*} 
\end{enumerate} 

\section{Probabilistic Kernel Ridge Regression: the Quasi GP model}\label{ProbKRR}

Quasi Gaussian Process (Q-GP) model  is an intermediate model between RVM and GP, which can be also useful for achieving a better understanding of both.  The Q-GP model represents the probabilistic version of the so-called {\it Kernel Ridge Regression} \cite{bishop2006pattern,murphy12}. Therefore, Q-GP is a probabilistic version of the Kernel Ridge regression, sometimes called {\it Bayesian Ridge Regression}.  Q-GP is also related to the so-called {\it regularization networks} in the literature \cite{Poggio20}. 
We highlight in advance the  connections with RVM for helping the reader. 


{\Remark  Q-GP is a special case of RVM with a specific choice of ${\bm \Sigma}_\rho={\bm \Psi}^{-1}$ (i.e., the covariance prior over ${\bm \rho}$).
Note that, we need  ${\bm \Psi}={\bm \Psi}^{\top}$, unlike in RVM.  }

{\Remark  In the smoothing problem, Q-GP  can be also derived with a specific choice of ${\bm \Psi}$ as covariance prior over $\f$. }




\subsection{Q-GP solution for regression} \label{QGPsectReg}
We consider the same classical Bayesian approach used for RVMs.
Namely, the observation model is again $y_i= {\bm \psi}({\bf x}_i)^{\top} {\bm \rho}+e_i$,
as in Eq.~\eqref{aquiEqPsiVect}. 
 However, in this case, we assume the following prior over the vector of weights,
\begin{eqnarray}
p({\bm \rho}) =\mathcal{N}({\bm \rho}|{\bf 0}, {\bf \Psi}^{-1}) \propto \exp(- {\bm \rho}^{\top}{\bf \Psi} {\bm \rho} ),
\end{eqnarray}
i.e., ${\bm \Sigma}_\rho={\bf \Psi}^{-1}$. This is possible just if ${\bf \Psi}^{-1}$ is a covariance matrix, so that it must be positive definite and symmetric, hence   ${\bm \Psi}={\bm \Psi}^{\top}$. The rest of formulas can be obtained replacing ${\bm \Sigma}_\rho={\bf \Psi}^{-1}$ and ${\bm \Psi}={\bm \Psi}^{\top}$, in the RVM expressions. For instance, replacing  ${\bm \Sigma}_\rho={\bf \Psi}^{-1}$ in Eq.\eqref{Mean_Post_rho_last}, we have  $p({\bm \rho}|{\bf y})=\mathcal{N}({\bm \rho}|{\bm \mu}_{\rho|y}, {\bm \Sigma}_{\rho|y})$,   
where 
\begin{eqnarray}\label{Mean_Post_rho_QGP}
 \widehat{{\bm \rho}}={\bm \mu}_{\rho|y}&=& \left({\bm \Psi}+\sigma_e^2 {\bf I}_N\right)^{-1}\by \nonumber  \\ 
 {\bm \Sigma}_{\rho|y}&=&{\bf \Psi}^{-1}-\left({\bm \Psi}+\sigma_e^2{\bf I}_N\right)^{-1},
\end{eqnarray}
The marginal likelihood is $p(\y)=\mathcal{N}({\bf y}|{\bf 0}, {\bm \Psi}+\sigma_e^2{\bf I}_N)$, and the posterior of $f(\x)$ in a generic $\x \in \mathcal{X}$, is also Gaussian, 
$$
p(f(\x)|{\bf y}) =\mathcal{N}\left(f(\x)| \mu_{f|y}(\x), \sigma_{f|y}^2(\x)\right),
$$
 with
\begin{gather}
\begin{split}
\mu_{f|y}(\x)= \widehat{f}(\x)={\bm \psi}({\bf x})^{\top} \left({\bm \Psi}+\sigma_e^2 {\bf I}_N\right)^{-1}\by,  
\end{split}
\end{gather}
and 
\begin{eqnarray}
 \sigma_{f|y}^2(\x)={\bm \psi}({\bf x})^{\top}{\bf \Psi}^{-1}{\bm \psi}({\bf x})- {\bm \psi}({\bf x})^{\top}\left({\bm \Psi}+\sigma_e^2{\bf I}_N\right)^{-1}{\bm \psi}({\bf x}).
\end{eqnarray}
 {\Remark We will see, that the mean solution $\widehat{f}(\x)$ of Q-GP coincides perfectly with the standard GP solution, that we will describe later. However, Q-GP  and GP differ for the expression of variance $\sigma_{f|y}^2(\x)$, for a generic $\x$  (as we will show later).  For further details, see below and \cite{Kimeldorf70}.}


\subsection{Q-GP for smoothing}\label{QGP_smoothing}
For obtaining the expressions for the   smoothing scenario, we can replace the vector ${\bm \psi}({\bf x})^{\top}$ with the matrix ${\bf \Psi}$ in the formulas above.
However, in the smoothing case, Q-GP can be directly derived assuming a particular prior over $f(\x)$. Although the formulas of the  Q-GP for smoothing can be obtained as particular case of the expressions above, we repeat the derivation since it can be useful for understanding  the classical GP derivation (in the next section).
We will use again the previous standard Bayesian approach, but now we focus on removing noise of the observations $y_n$ at the inputs $\x_n$ obtaining $\widehat{\f}$, and we will consider a specific covariance prior over $\f$. More specifically, we consider a Gaussian prior over the vector $\f$ (i.e., a prior over the {\it  hidden function}),
\begin{equation}
 \f \sim \mathcal{N}(\f|{\bf 0},{\bm \Psi})\propto \exp(- {\bf f}^{\top}{\bf \Psi}^{-1} {\bf f} ),
\end{equation}
where ${\bm \Psi}$ is exactly the design matrix in Eq.~\eqref{psiMAT}. This is possible only if ${\bm \Psi}$ can represent a covariance matrix, then ${\bm \Psi}$ must be positive definite and symmetric, ${\bm \Psi}={\bm \Psi}^{\top}$. 
Therefore, we need that $\psi_n(\x|{\bf z})=\psi_n({\bf z}|\x)$. Recall that the observation model has the form
\[
 \y= \f+{\bf e}.
\]
We are interested in inferring $\f$ given $\y$ under the assumption ${\bf e}\sim \mathcal{N}({\bf e}|{\bf 0},\sigma_e^2 {\bf I}_N)$. Then, the likelihood is
  \begin{equation}
 p(\y|\f)=\mathcal{N}({\bf e}|\f,\sigma_e^2 {\bf I}_N).
\end{equation}
 Hence, the marginal likelihood is
  \begin{equation}
 p(\y)=\int_{\mathbb{R}^N}p(\y|\f)p(\f) d\f= \mathcal{N}({\bf y}|{\bf 0},{\bm \Psi}+\sigma_e^2 {\bf I}_N).
\end{equation}
The posterior of the vector $\f$ is 
\begin{eqnarray}
p(\f|\y)&=&\frac{1}{ p(\y)} p(\y|\f)p(\f)\propto  p(\y|\f)p(\f) \\
&=&\mathcal{N}({\bf f}|{\bm \mu}_{f|y},{\bm \Sigma}_{f|y}),
\end{eqnarray}
where the vector mean ${\bm \mu}_{f|y}=\widehat{\f}$ and covariance matrix are
\begin{eqnarray}
{\bm \mu}_{f|y}&=&\widehat{\f}={\bf \Psi}({\bf \Psi}+\sigma_e^2 {\bf I}_N)^{-1} \y, \quad \mbox{and} \label{QGP_smooth1} \\
{\bm \Sigma}_{f|y}&=&  \left[\left({\bf \Psi} \right)^{-1}+\left(\sigma_e^2 {\bf I}_N \right)^{-1}\right]^{-1} \nonumber \\
&=& {\bf \Psi}- {\bf \Psi}\left({\bf \Psi}+\sigma_e^2 {\bf I}_N\right)^{-1}{\bf \Psi}. \label{QGP_smooth2}
\end{eqnarray}

{\Remark The solution above of Q-GP for smoothing, represented by Eqs.~\eqref{QGP_smooth1}--\eqref{QGP_smooth2},  coincides perfectly with the standard GP solution for smoothing, that we will describe below. See Eqs.~\eqref{SmoothGP1}--\eqref{SmoothGP2}. }


\section{Gaussian Processes (GPs)}\label{GPsect} 


\subsection{Definition}

For simplicity, let us consider ${\bf \psi}_n(\x,\x_n)={\bf \psi}(\x,\x_n)$. 
Moreover, let us assume that the nonlinearity ${\bf \psi}$ is chosen such that (a) ${\bf \psi}(\x,\x)>0$, (b) the design matrix ${\bm \Psi}={\bm \Psi}^{\top}$ (symmetric) and (c) ${\bm \Psi}$ is a positive-definite matrix. In this case, we can interpret the design matrix ${\bm \Psi}$ as a covariance matrix (as we have assumed in Section \ref{QGPsectReg}). As in Section \ref{QGP_smoothing}, we can consider a Gaussian prior over the vector $\f=[f(\x_1),\ldots,f(\x_N)]^{\top}$, e.g.,
\begin{equation}
\label{EqPriorGPoverf}
 \f \sim \mathcal{N}(\f|{\bf 0},{\bm \Psi})\propto \exp(- {\bf f}^{\top}{\bf \Psi}^{-1} {\bf f} ),
\end{equation}
assuming a zero mean vector for the sake of simplicity. We can generalize the idea above, for two generic inputs $\x$ and ${\bf z}$, which are not (necessarily) in the input dataset $\{\x_1,\ldots,\x_N\}$. Let consider that the function ${\bf \psi}$ is symmetric, i.e., ${\bf \psi}(\x,{\bf z})={\bf \psi}({\bf z},\x)$ for all $\x,{\bf z} \in \mathcal{X}$ and represents a {\it covariance function} \cite{rasmussen2003gaussian,bishop2006pattern} (see below). 

{\Remark We are assuming that function $\psi(\x,{\bf z})$ represents the covariance function between the random variables $f(\x)$ and $f({\bf z})$, at two {\it generic} inputs $\x$ and ${\bf z}$. Namely, 
\begin{equation}
\psi(\x,{\bf z})=E[(f(\x)- \mu(\x))(f({\bf z})-\mu({\bf z}))],
\end{equation}
 where we have assumed for simplicity $\mu(\x)=0$, $\mu({\bf z})=0$.  Thus,  $\psi(\x,\x)=E[(f(\x)- \mu(\x))^2]$ is the variance the random variable $f(\x)$, i.e, $p(f(\x)) \sim \mathcal{N}(0, \psi(\x,\x))$.}
\newline
\newline
Thus, we can assume that the two random variables $f(\x)$ and $f({\bf z})$ are jointly Gaussian, with mean $[0,0]^{\top}$ and $2\times 2$ covariance matrix
 \begin{equation}
{\bf C}(f(\x),f({\bf z}))=
\begin{bmatrix}
\psi(\x,\x) & \psi({\bf z}, \x)  \\
\psi(\x,{\bf z})& \psi({\bf z},{\bf z}) 
\end{bmatrix}.
\end{equation}
Considering 3 generic inputs $\x, {\bf z}, {\bf t}\in \mathcal{X}$, we have the following covariance matrix
 \begin{equation}
{\bf C}(f(\x),f({\bf z}), f({\bf t}))=
\begin{bmatrix}
\psi(\x,\x) & \psi({\bf z},\x) & \psi({\bf t},\x)  \\
\psi(\x,{\bf z})& \psi({\bf z},{\bf z})  & \psi({\bf t},{\bf z}) \\
\psi(\x,{\bf t})& \psi({\bf z},{\bf t})  & \psi({\bf t},{\bf t})
\end{bmatrix}.
\end{equation}
Moreover, considering all the input dataset $\{\x_n\}_{n=1}^N$, we have that 
$$
{\bf C}(f(\x_1),f(\x_2),\ldots,f(\x_N))={\bm \Psi}.
$$
{\bf Marginal Likelihood.} Recalling that $\y={\bf f}+{\bf e}$,  then  the marginal likelihood is again
\begin{equation}
p({\bf y}) =\mathcal{N}({\bf y}|{\bf 0},{\bm \Psi} +\sigma_e^2 {\bf I}_N).
\end{equation}
where we have the sum of two independent multivariate Gaussian random variables  $\f \sim \mathcal{N}(\f|{\bf 0},{\bm \Psi})$ in Eq.~\eqref{EqPriorGPoverf} and ${\bf e}\sim \mathcal{N}({\bf e}|{\bf 0},\sigma_e^2 {\bf I}_N)$.

{\Remark  If the kernel function is stationary $\psi(\x,{\bf z})=\psi(||\x-{\bf z}||)$, we are converting the distance between the inputs $\x$ and ${\bf z}$, into an priori covariance/correlation information. Often, we associate small correlation to high distances and  high correlation to small distances. }
\newline
\newline
{\bf Definition.}  ``A Gaussian process (GP) is a collection of random variables, any finite number of which have a joint Gaussian distribution'' \cite{rasmussen2003gaussian}.  A GP is completely specified by its mean function $m(\x): \mathcal{X}\rightarrow \mathbb{R}$ (that we have assumed $m(\x)=0$, for simplicity) and its covariance function ${\bf \psi}(\x,{\bf z}):\mathcal{X} \times \mathcal{X} \rightarrow \mathbb{R}$.

\begin{figure}[!h]
	\centering
\centerline{	
	\includegraphics[width=12cm]{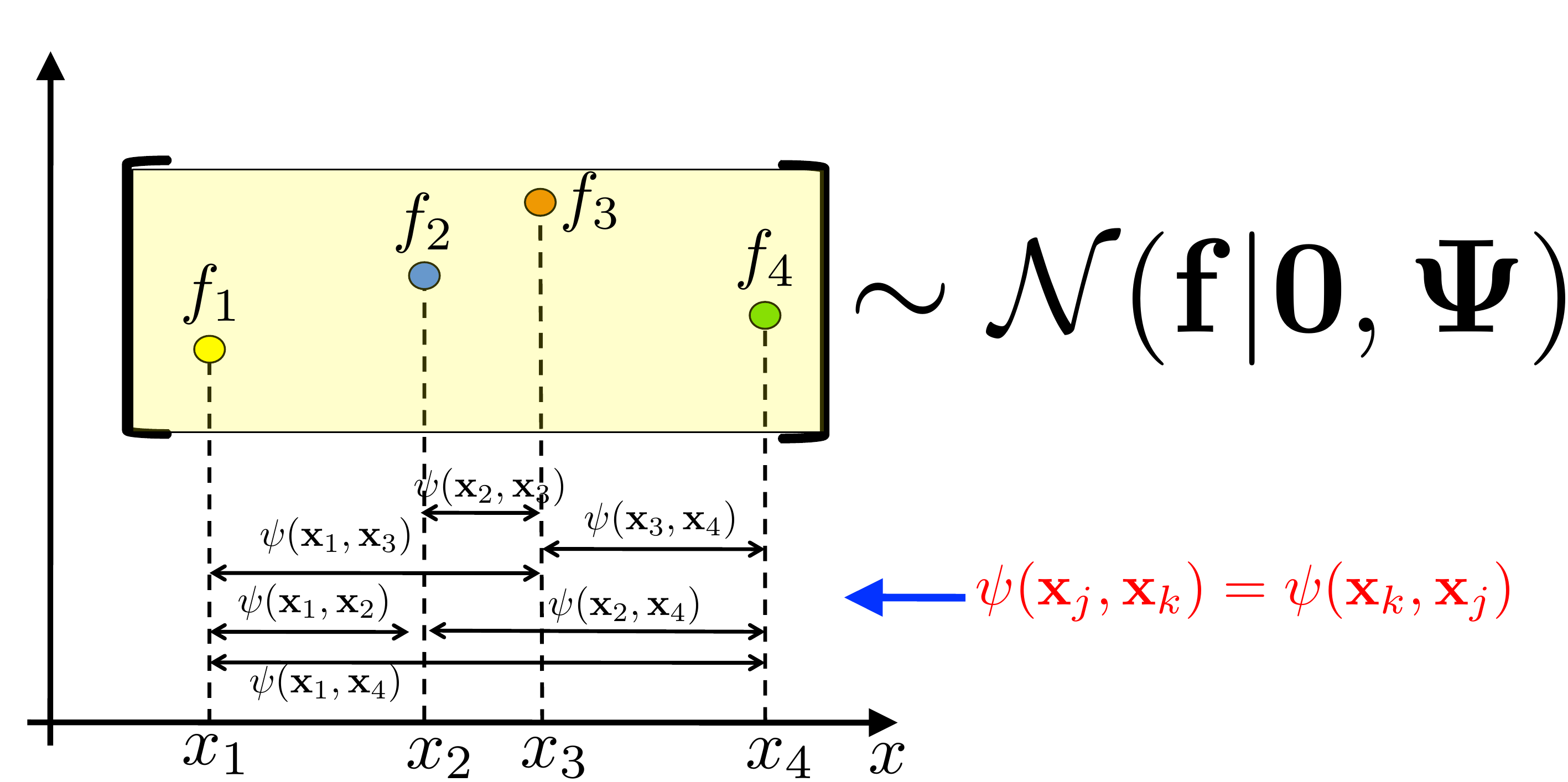}
}
	\caption{Graphical representation of a GP prior idea with $N=4$ and ${\bm \Psi}={\bf C}(\x_1,\x_2,\x_3,\x_4)$.}
	\label{figGPprior}
\end{figure}

A graphical representation of a GP prior considering a vector of dimension $N=4$ is given in Figure \ref{figGPprior}. 

\subsection{Posterior density of a GP in regression}
Let us continue the derivation of the main GP formulas for regression, considering the joint probability $p(\y, f(\x))$ where $\x\in \mathcal{X}$ is a generic input and $\y={\bf f}+{\bf e}$ (recall that $\f=[f(\x_1),\ldots,f(\x_N)]^{\top}$).  By the GP definition,  
\begin{eqnarray}\label{EqyFx}
p(\y, f(\x))=\mathcal{N}([\y, f(\x)]^{\top}| {\bm \mu}_{\texttt{joint}},  {\bf C}_{\texttt{joint}})
\end{eqnarray}
where ${\bm \mu}_{\texttt{joint}}=[0,\ldots,0]^{\top}$ is a null vector of dimension $(N+1)$, and 
\begin{eqnarray}
\label{psiMAT_gp}
{\bf C}_{\texttt{joint}}={\bf C}(\y,f(\x)) =
\begin{bmatrix}
{\bm \Psi} +\sigma_e^2 {\bf I}_N & {\bm \psi}({\bf x})  \\
{\bm \psi}({\bf x})^{\top} & \psi(\x,\x)
\end{bmatrix},
\end{eqnarray}
is a $(N+1)\times (N+1)$ covariance matrix. We recall that 
$$
{\bm \psi}({\bf x})=[\psi({\bf x},{\bf x}_1),\ldots,\psi({\bf x},{\bf x}_N)]^{\top},
$$
 is  the $N\times 1$ design vector. The first bock in the diagonal of ${\bf C}_{\texttt{joint}}$ is the covariance  $N\times N$ matrix of $\y$ i.e., ${\bf C}(\y,\y)={\bm \Psi} +\sigma_e^2 {\bf I}_N$, and the last element in the diagonal is $\mbox{var}[f(\x)]=\psi(\x,\x)$. The covariance of each element $y_n$  of the vector $\y=[y_1,\ldots,y_N]^{\top}$,  and the random variable $f(\x)$ is $\psi({\bf x}_n,{\bf x})=\psi({\bf x},{\bf x}_n)$. All those $N$ covariances are contained in the vector ${\bm \psi}({\bf x})$.  We will use the following property of the Gaussian distributions,
\begin{eqnarray}
p({\bf a},{\bf b})\sim \mathcal{N}\left([{\bf a},{\bf b}]^{\top}\Big | [{\bm \mu}_a, {\bm \mu}_b]^{\top},  
\begin{bmatrix}
{\bf C}_a  & {\bm \Lambda } \\
 {\bm \Lambda } & {\bf C}_b
\end{bmatrix},
\right)
\end{eqnarray}
then the conditional pdf $p({\bf b}|{\bf a})=\mathcal{N}({\bf b}| {\bm \mu}_{b|a}, {\bf C}_{b|a})$ has the following mean and variance,
\begin{eqnarray}
{\bm \mu}_{b|a}={\bm \mu}_{b}+{\bm \Lambda}^{\top}  {\bf C}_a^{-1}({\bf a}-{\bm \mu}_a), \quad \quad {\bf C}_{b|a}={\bf C}_b-{\bm \Lambda}^{\top} {\bf C}_a^{-1} {\bm \Lambda}.
\end{eqnarray}
Hence, given the joint probability in \eqref{EqyFx}, now we can obtain the mean and variance of posterior pdf 
$$
p(f(\x)|\y)=\mathcal{N}\left(f(\x)|\mu_{f|y}(\x), \sigma_{f|y}^2(\x)\right).
$$
Thus, in this case, we have ${\bf a}=\y$, ${\bm \mu}_{a}$, ${\bm \mu}_{b}$ are zero, ${\bf C}_a={\bm \Psi} +\sigma_e^2 {\bf I}_N$, ${\bm \Lambda }={\bm \psi}({\bf x})$ and ${\bf C}_b=\psi(\x|\x)$ (i.e., a scalar in this case), 
\begin{eqnarray}
\mu_{f|y}(\x)&=& \widehat{f}(\x)={\bm \psi}({\bf x})^{\top}({\bm \Psi} +\sigma_e^2 {\bf I}_N)^{-1}\y, \label{GPmu1} \\
\sigma_{f|y}^2(\x)&=&\psi(\x,\x)-{\bm \psi}({\bf x})^{\top}({\bm \Psi} +\sigma_e^2 {\bf I}_N)^{-1}{\bm \psi}({\bf x}). \label{GPvar1}
\end{eqnarray}
 {\Remark Note that, also in this case, $\widehat{f}({\bf x})$ can be expressed as Eq.~\eqref{aquiF2}, i.e., $\widehat{f}({\bf x})={\bm \psi}({\bf x})^{\top} \widehat{{\bm \rho}}$ where 
\begin{equation}
\label{RhoGPs}
 \widehat{{\bm \rho}}=({\bm \Psi} +\sigma_e^2 {\bf I}_N)^{-1}\y.
 \end{equation}
 }
 {\Remark Also in the GP formulation, with noise-free data $\sigma_e^2=0$  (interpolation), we come back to $\widehat{{\bm \rho}}={\bm \Psi}^{-1}\by$, as expected. }
 \newline 
 \newline
{\bf Posterior of several test points.} The formulas \eqref{GPmu1}--\eqref{GPvar1} can be easily generalized when we consider the posterior distribution of the hidden function in $P$ different generic test points 
$$
f(\x^{(1)}), f(\x^{(2)}),\ldots,f(\x^{(P)}).
$$
 In this case,  we have to replace above the $N\times 1$ vector ${\bm \psi}({\bf x})$ with the $N\times P$ matrix ${\bf V}=[{\bm \psi}({\bf x}^{(1)}),\ldots., {\bm \psi}({\bf x}^{(P)})]$, and the scalar value $\psi(\x|\x)$ with the $P\times P$ covariance matrix
$$
{\bf C}={\bf C}(f(\x^{(1)}),\ldots,f(\x^{(P)}))=
\begin{bmatrix}
 \psi(\x^{(1)},\x^{(1)}) &  \cdots & \psi(\x^{(P)},\x^{(1)})  \\
 \psi(\x^{(1)},\x^{(2)}) &    \cdots & \psi(\x^{(P)},\x^{(2)})  \\
  \vdots &    \cdots &  \vdots  \\ 
 \psi(\x^{(1)},\x^{(P)}) &  \cdots & \psi(\x^{(P)},\x^{(P)})  \\
\end{bmatrix}.
$$
Then, the posterior pdf is a multivariate Gaussian with the following $P\times 1$ mean vector and $P\times P$ covariance matrix
\begin{eqnarray}
{\bm \mu}_{f|y}&=&{\bf V}^{\top}({\bm \Psi} +\sigma_e^2 {\bf I}_N)^{-1}\y, \label{GPmu3} \\
{\bf \Sigma}_{f|y}&=&{\bf C}-{\bf V}^{\top}({\bm \Psi} +\sigma_e^2 {\bf I}_N)^{-1}{\bf V}. \label{GPvar4}
\end{eqnarray}
where ${\bm \mu}_{f|y}=[\widehat{f}(\x^{(1)}),\ldots,\widehat{f}(\x^{(P)})]^{\top}$.
\newline
\newline
 {\bf Smoothing case.} If we consider the training inputs $\x_1,\ldots,\x_N$, then 
 $$
 {\bf C}(f(\x_1),\ldots,f(\x_N))={\bf \Psi}.
 $$ 
 The posterior of the vector $\f$ is 
$$
p(\f|\y)=\mathcal{N}(\f|{\bm \mu}_{f|y}, {\bm \Sigma}_{f|y}),
$$
where 
\begin{eqnarray}
{\bm \mu}_{f|y}&=& \widehat{\f}={\bf \Psi}({\bm \Psi} +\sigma_e^2 {\bf I}_N)^{-1}\y, \label{SmoothGP1}  \\
 {\bm \Sigma}_{f|y}&=&{\bf \Psi}-{\bf \Psi}({\bm \Psi} +\sigma_e^2 {\bf I}_N)^{-1}{\bf \Psi}.  \label{SmoothGP2}
\end{eqnarray}
Note that we have used ${\bf \Psi}={\bf \Psi}^{\top}$. 
 {\Remark The expressions \eqref{SmoothGP1}--\eqref{SmoothGP2} are exactly the same as in Eqs.~\eqref{QGP_smooth1}--\eqref{QGP_smooth2} of the Q-GP. }
\subsection{Generation of random functions according to GP models}\label{GPgeneration}
\begin{figure}[!h]
	\centering
\centerline{	
	\subfigure[]{\includegraphics[width=7cm]{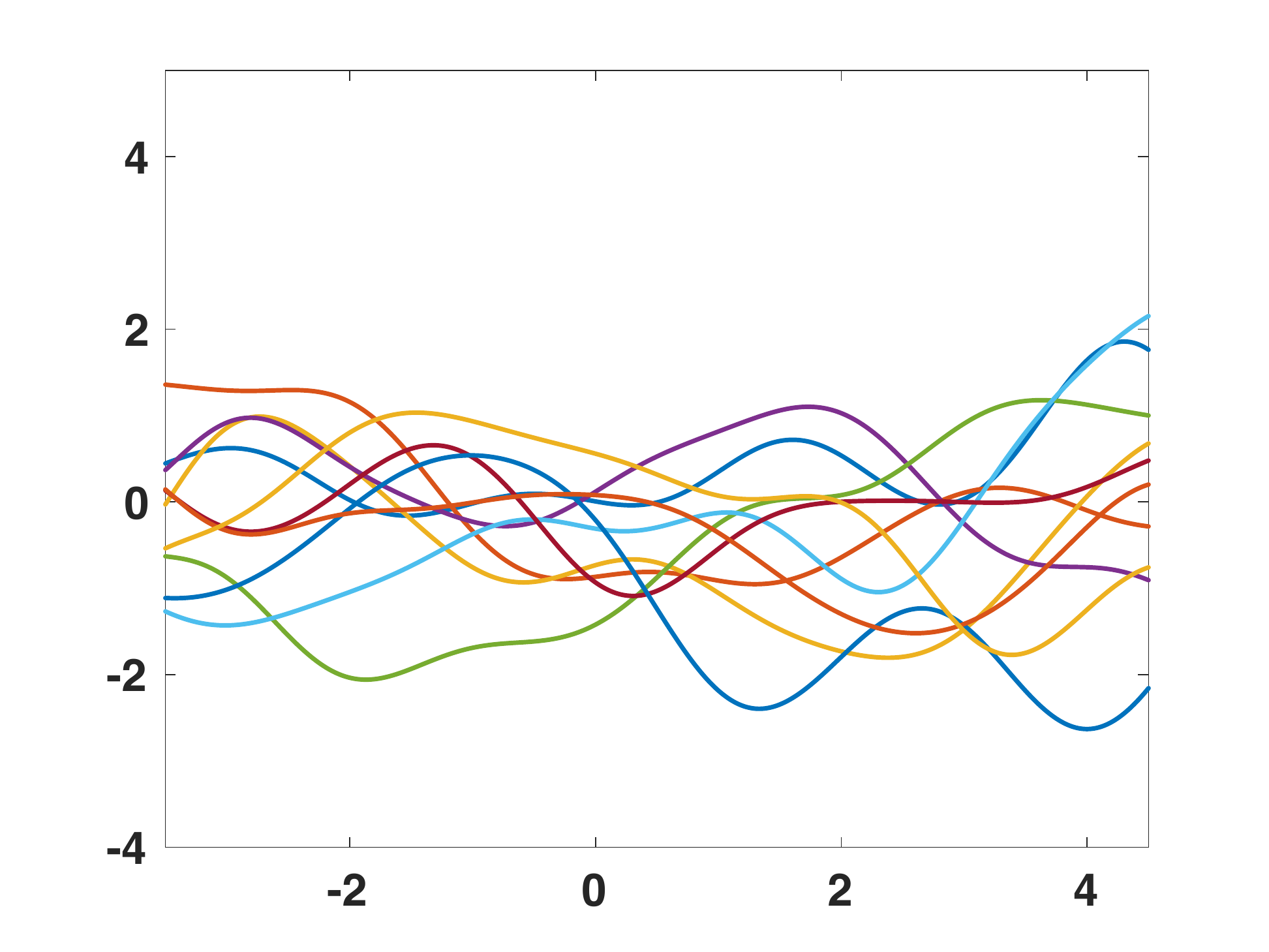}}
	\hspace{-0.9cm}
	\subfigure[]{\includegraphics[width=7cm]{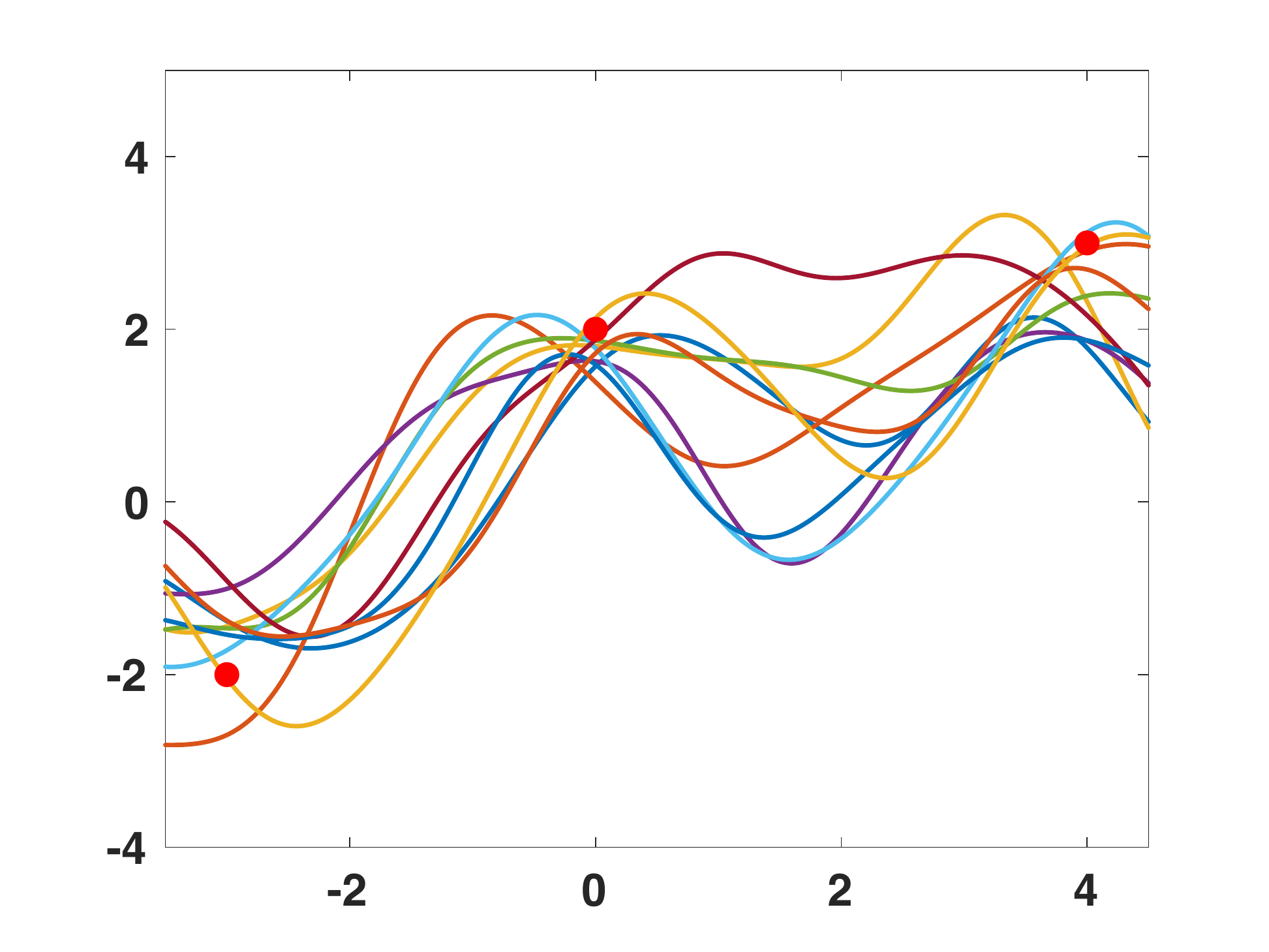}}
	\hspace{-0.9cm}	
	\subfigure[]{\includegraphics[width=7cm]{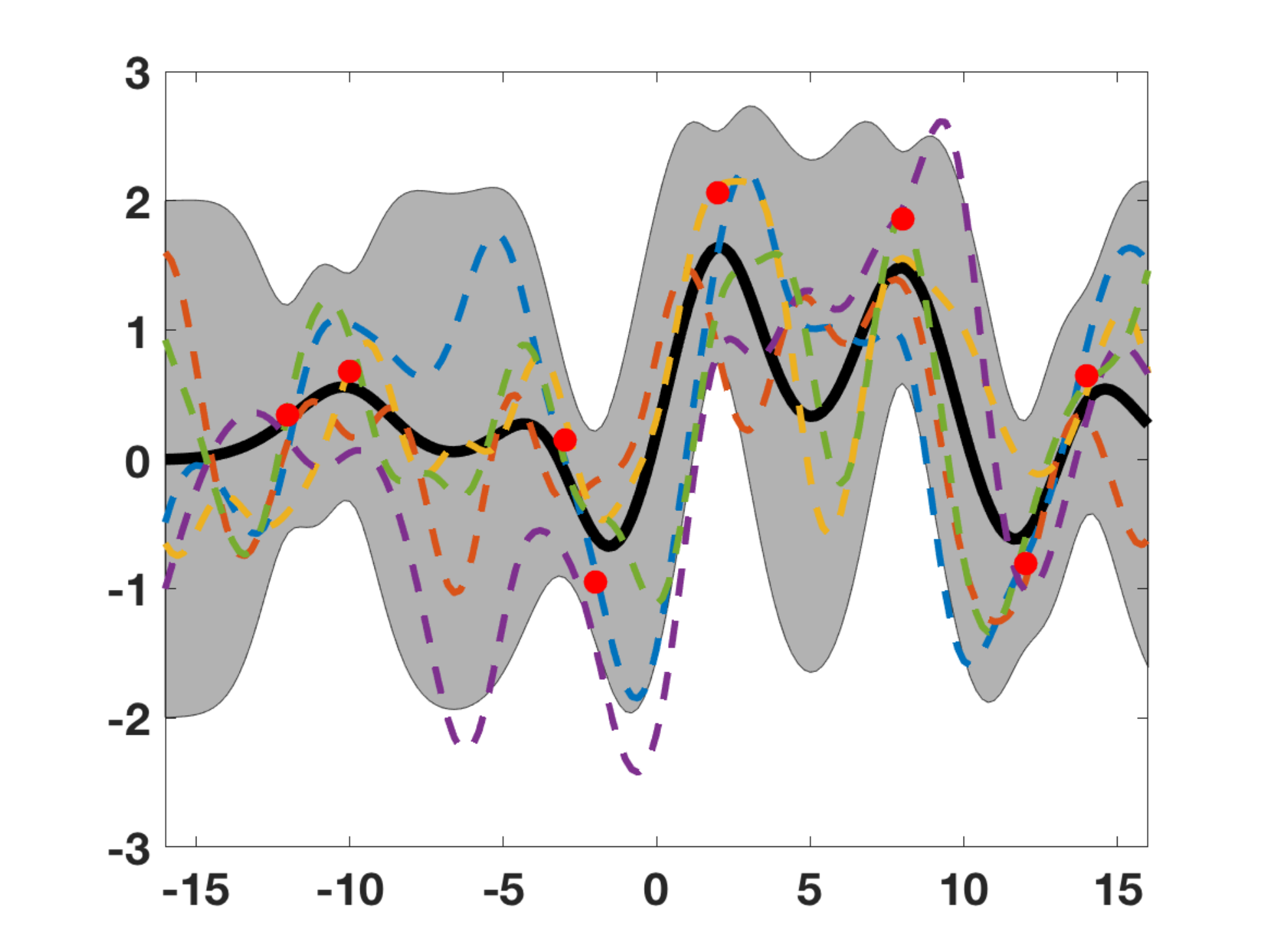}}
}
	\caption{Random functions $\widehat{f}^{(s)}(\x)$ with $\x=xÊ\in \mathbb{R}$ scalar, and $s=1,\ldots,10$  {\bf (a)} from a GP prior over $f(\x)$ and {\bf (b)} from a GP posterior (after knowing the $N=3$ data points), with $N=3$ Gaussian kernels and bandwidth $\lambda=2$, $\sigma_e=0.5$. {\bf (c)} Example of GP mean and variance with $N=8$ data points and $S=5$ random functions from the posterior depicted with dashed lines ($\sigma_e=0.5$, $\lambda=4$). The black solid line shows the mean $ \mu_{f|y}(\x)=\widehat{f}(\x)$ and the boundary of the grey area corresponds to $\widehat{f}(\x)\pm2 \sigma_{f|y}^2(\x)$ (i.e., $\approx 95\%$ of the probability).}
	\label{figRandomGP}
\end{figure}

{\bf Drawing functions from the GP prior.}  Let us consider that we desire to know (and then, e.g., to plot) a random function $f(\x)$ in the input points $\x^{(1)},\ldots,\x^{(P)}$.
Then, the following procedure generates $S$ random ``functions'' (represented as random vectors) from a GP prior with kernel function $\psi(\x,{\bf z})$:
\begin{enumerate}
\item Compute the $P\times P$ covariance matrix
\begin{equation*}
{\bf C}={\bf C}(f(\x^{(1)}),\ldots,f(\x^{(P)}))=
\begin{bmatrix}
 \psi(\x^{(1)},\x^{(1)}) & \vdots &   \psi(\x^{(P)},\x^{(1)}) \\
\vdots& \ddots &   \vdots \\
 \psi(\x^{(1)},\x^{(P)})  &\cdots  &  \psi(\x^{(P)},\x^{(P)}) 
\end{bmatrix}.
\end{equation*}
Recall that, for simplicity, we have assumed ${\bm \mu}=[\mu(\x^{(1)}),\ldots,\mu(\x^{(P)})]^{\top}=[0,\ldots,0]^{\top}$.
\item Draw  $S$  vectors ${\bf f}_P^{(s)}=[f^{(s)}(\x^{(1)}),\ldots,f^{(s)}(\x^{(P)})]^{\top}$ from a multivariate Gaussian with zero mean and covariance matrix ${\bf C}$ above, i.e., 
$$
{\bf f}_P^{(s)}\sim\mathcal{N}({\bf f}_P|{\bf 0},{\bf C}), \quad s=1,\ldots,S.
$$
 
\end{enumerate}
{\bf Drawing functions from the GP posterior.}  Let us consider the test points $\x^{(1)},\ldots,\x^{(P)}$.
Then, the following procedure generates $S$ random ``functions'' (that actually are random vectors) from a GP prior with kernel function $\psi(\x,{\bf z})$:
\begin{enumerate}
\item Compute  the $N\times P$ matrix ${\bf V}=[{\bm \psi}({\bf x}^{(1)}),\ldots, {\bm \psi}({\bf x}^{(P)})]$. Recall that ${\bm \psi}({\bf x})=[\psi_1({\bf x},{\bf x}_1),\ldots,\psi_N({\bf x},{\bf x}_N)]^{\top}$ is the $N\times 1$ design vector.
\item Compute  the  $P\times P$ covariance matrix
$$
{\bf C}={\bf C}(f(\x^{(1)}),\ldots,f(\x^{(P)}))=
\begin{bmatrix}
 \psi(\x^{(1)},\x^{(1)}) &  \cdots & \psi(\x^{(P)},\x^{(1)})  \\
 \psi(\x^{(1)},\x^{(2)}) &    \cdots & \psi(\x^{(P)},\x^{(2)})  \\
 \vdots & \vdots&  \vdots \\
 \psi(\x^{(1)},\x^{(P)}) &  \cdots & \psi(\x^{(P)},\x^{(P)})  \\
\end{bmatrix}.
$$
\item Draw  $S$  vectors ${\bf f}_P^{(s)}=[f^{(s)}(\x^{(1)}),\ldots,f^{(s)}(\x^{(P)})]^{\top}$ from a multivariate Gaussian, i.e., 
$$
{\bf f}_P^{(s)}\sim\mathcal{N}({\bf f}_P|{\bm \mu},{\bf \Sigma}), \quad s=1,\ldots,S,
$$
 where 
 \begin{align*}
 {\bm \mu}&={\bf V}^{\top}({\bm \Psi} +\sigma_e^2 {\bf I}_N)^{-1}\y \quad\quad\quad \mbox{is a $P\times 1$ mean vector, and} \\
 {\bf \Sigma}&={\bf C}-{\bf V}^{\top}({\bm \Psi} +\sigma_e^2 {\bf I}_N)^{-1}{\bf V} \quad \mbox{is a $P\times P$ covariance matrix}.
 \end{align*}
\end{enumerate}
Figure \ref{figRandomGP} shows some examples of random functions, predictive mean and variance of GP model with a Gaussian kernel function.

\subsection{Interpretation of the hyper-parameters} \label{SuperSECTGhahramani11}
The kernel hyper-parameters can be learned from data, maximizing the marginal likelihood or by a cross-validation (CV) approach. In some cases, they are also interpretable in statistical terms \cite{Ghahramani11}.
As an example, let $x\in \mathbb{R}$ and consider the following  exponential kernel function 
\begin{equation}\label{kernelcompleto}
k(x_i, x_j)=a \exp\left(-\frac{|x_i- x_j|^\beta}{\lambda}\right)+v_1+v_2 \cdot \delta_{ij}
\end{equation}
where $a,\lambda, v_1,v_2,\beta>0$. Moreover, $\delta_{ij}=1$ when $i=j$ and zero otherwise. The statistical interpretation of each parameter is:
\begin{itemize}
\item $a:$ {\it a-priori signal variance,} i.e., the prior variance that the random function $f(x)$ has following the user's belief, without knowing any data. In regions where there are no data points, the posterior/predictive variance will be $a$.
\item $\lambda:$ {\it lengthscale.} This parameter determines the oscillations that the solution has. The optimal $\lambda$ becomes usually smaller as the number of data points grows and becoming closer and closer.  
\item  $\beta:$ {\it roughness.} This parameter determines the derivability and the smoothness of the resulting solution.
\item $v_1:$ {\it variance of bias.}
\item $v_2:$ {\it additional noise power.} Since we consider the parameter $\sigma_e^2$, we can avoid the use of $v_2$.
\end{itemize}

\subsection{Relevant GP special cases}
\begin{figure*}[htb]
	\centering
	\centerline{    
		\subfigure[]{\includegraphics[width=7cm]{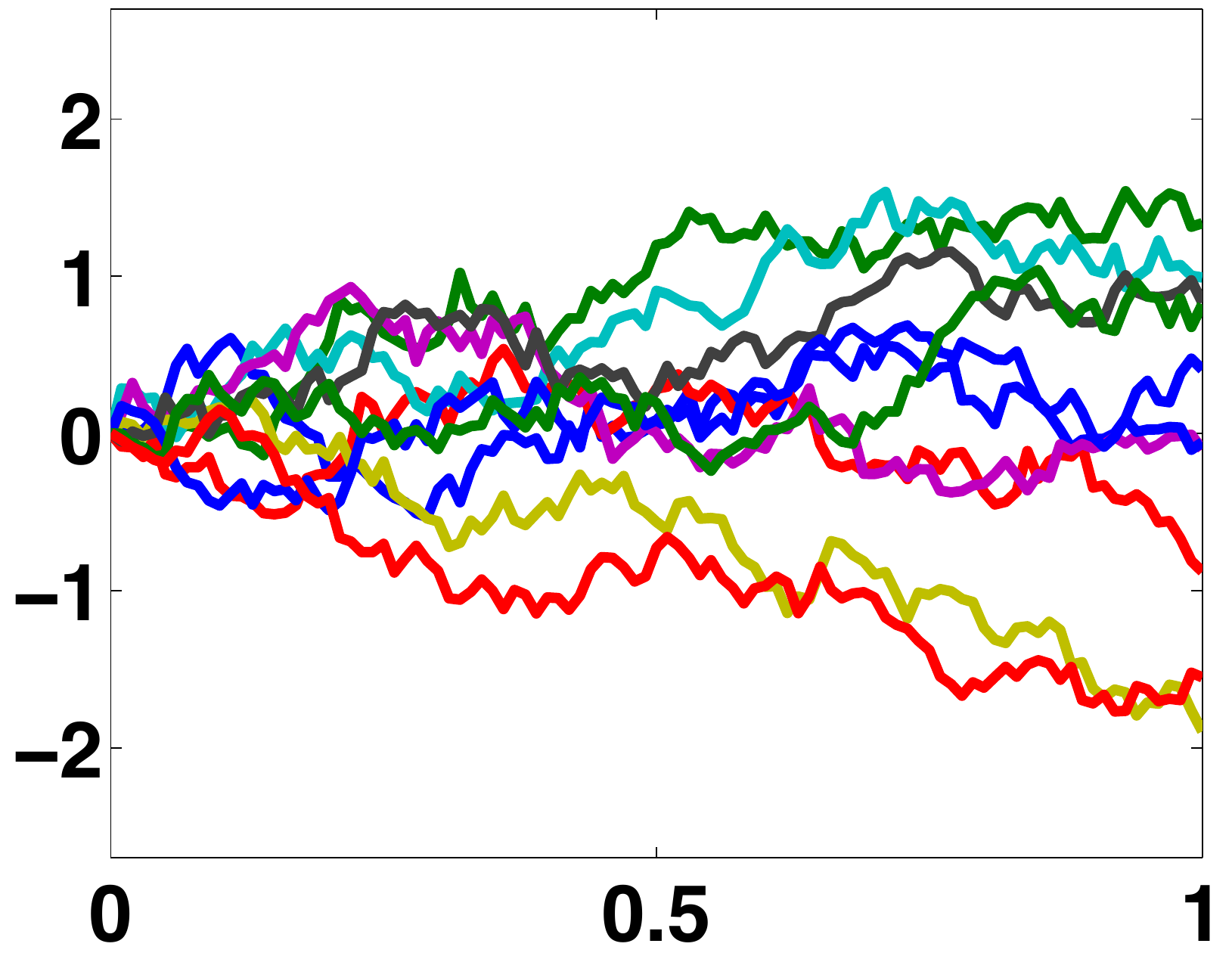}}
		\subfigure[]{\includegraphics[width=7cm]{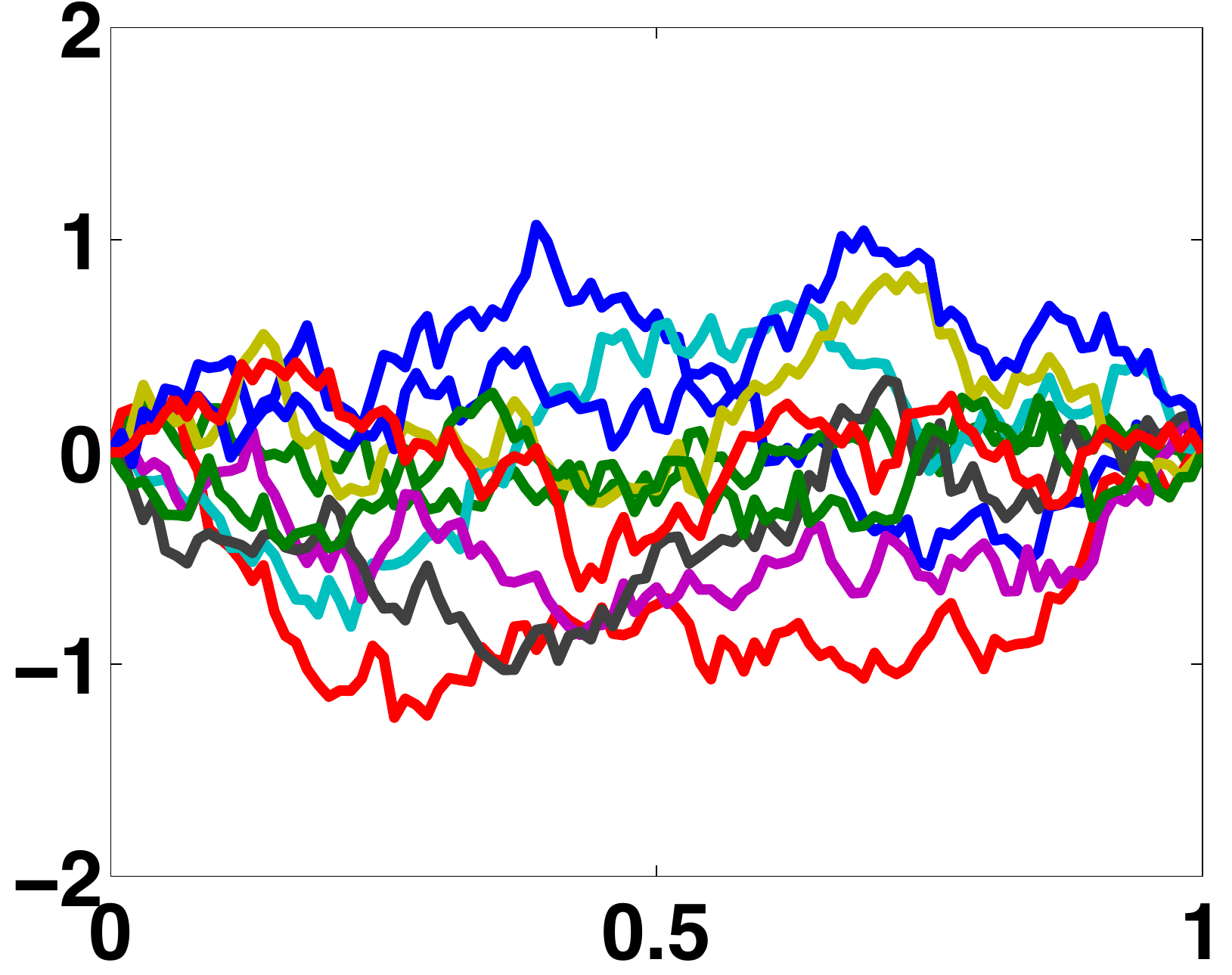}}
	}
\caption{Ten independent realizations {\bf (a)} of a Wiener process and {\bf (b)} of a  a standard Brownian bridge.  } 
\label{FigWP}
\end{figure*}

For simplicity, let us assume again a scalar input, $x\in \mathbb{R}$. Different well-known stochastic processes are GPs \cite[Chapter 6]{BookLuca}. Table \ref{TableGP} provides some examples, clarifying the specific choice of the mean $\mu(x)$ and covariance function $\psi(x,z)$ of the GP prior. Recall that, in this work, we have always considered $\mu(x)=0$ for the sake of simplicity. Figure \ref{FigWP} depicts ten realizations of a Wiener process and of a standard Brownian bridge. 

\begin{table}[!hbt]
\begin{center}
\caption{Special cases of Gaussian processes. }
\label{TableGP}
\vspace{0.2cm}
\begin{tabular}{|c|c|c|}
\hline
{\bf Type} & {\bf Mean} $\mu(x)$ & {\bf Covariance function} $\psi(x,z)$\\
\hline
\hline
{\it Wiener process} & $0$   & $\min\{x,z\}$ \\
\hline
{\it  Standard Brownian} &\multirow{2}{*}{$0$}   & \multirow{2}{*}{$\min\{x,z\}-xz$} \\
 {\it bridge} & &  \\
\hline
{\it   Ornstein-Uhlenbeck} & \multirow{2}{*}{$e^{-\theta x} \mu_0+\nu (1- e^{-\theta x})$} &  \multirow{2}{*}{$\frac{\sigma^2}{2\theta} e^{-\theta (x+z)} \left(e^{2 \theta \min\{x,z\}}-1\right)$} \\
 {\it process} & &  \\
\hline
\end{tabular}
\end{center}
\end{table}

\section{Dual representation of RVM -  Dual Gaussian Process} \label{DualRepres}
In this section, we show that RVM method can be seen also a GP model with a specific choice of the kernel function. Namely, RVM is also a GP. However,
the fact of choosing {\it just indirectly} the kernel function can provide undesirable behavior in the predictive variance, 
as we discuss in Section \ref{MostImportSect}. Below, we obtain the covariance function (i.e., the kernel function) of RVM and related vectors and matrices.

\subsection{Covariance function of RVM} \label{SectDualK}
We can compute the covariance between the random variables $f=f(\x)$ and $f'=f(\x')$. Indeed, recalling that $f(\x)={\bm \psi}(\x)^{\top}{\bm \rho}$, we can write
\begin{eqnarray}
\label{AquiKeq}
k_{\texttt{dual}}(\x,\x')&=&E[f(\x)-\mu_f,f(\x')-\mu_f]   \nonumber  \\
&=&E[f(\x),f(\x')]   \nonumber  \\
&=&E[{\bm \psi}(\x)^{\top}{\bm \rho}, {\bm \psi}(\x')^{\top}{\bm \rho}]  \nonumber  \\
&=&{\bm \psi}(\x)^{\top}E[{\bm \rho},{\bm \rho}]{\bm \psi}(\x'),  \nonumber  \\
&=&{\bm \psi}(\x)^{\top}{\bm \Sigma}_{\rho}{\bm \psi}(\x'): \mathcal{X}\times\mathcal{X} \rightarrow \mathbb{R}.
\end{eqnarray}
where we have also used the fact that $E[{\bm \rho}]=0$, hence $E[{\bm \rho},{\bm \rho}]={\bm \Sigma}_{\rho}$.
The function $k_{\texttt{dual}}(\x,\x')$ is called {\it dual kernel function}, and it is also symmetric, i.e.,
\begin{eqnarray}
k_{\texttt{dual}}(\x,\x')&=&{\bm \psi}(\x)^{\top}{\bm \Sigma}_{\rho}{\bm \psi}(\x')   \nonumber \\
&=&{\bm \psi}(\x')^{\top}{\bm \Sigma}_{\rho}{\bm \psi}(\x)=k_{\texttt{dual}}(\x',\x).  \nonumber 
\end{eqnarray}
Moreover, we define the vector
\begin{eqnarray}
{\bf k}_{\texttt{dual}}(\x)&=&[k_{\texttt{dual}}(\x,\x_1),k_{\texttt{dual}}(\x,\x_2),\ldots,k_{\texttt{dual}}(\x,\x_N) ]^{\top} : \mathcal{X} \rightarrow \mathbb{R}^{N\times 1}  \nonumber  \\
&=&{\bm \psi}(\x)^{\top}{\bm \Sigma}_{\rho}{\bm \Psi}^{\top},\label{AquiKvect}
\end{eqnarray}
where we have considered the training inputs $\{\x_n\}_{n=1}^N$, and finally we introduce the $N\times N$ matrix
\begin{eqnarray}
{\bf K}_{\texttt{dual}}&=&[{\bf k}_{\texttt{dual}}(\x_1),{\bf k}_{\texttt{dual}}(\x_2),\ldots,{\bf k}_{\texttt{dual}}(\x_N)]^{\top},  \nonumber  \\
&=&{\bm \Psi}{\bm \Sigma}_{\rho}{\bm \Psi}^{\top}. \label{AquiKmat}
\end{eqnarray}
Therefore, the probabilistic model of RVM {\it indirectly} assumes {\it correlation} between two inputs $\x$ and ${\bf z}$,  
\begin{eqnarray}
\mbox{corr}(f(\x),f({\bf z}))=\frac{{\bm \psi}(\x)^{\top}{\bm \Sigma}_{\rho}{\bm \psi}({\bf z})}{\sqrt{{\bm \psi}(\x)^{\top}{\bm \Sigma}_{\rho}{\bm \psi}({\bf x})} \sqrt{{\bm \psi}({\bf z})^{\top}{\bm \Sigma}_{\rho}{\bm \psi}({\bf z})}}.
\end{eqnarray}
where $|\mbox{corr}(f(\x),f({\bf z}))| \leq 1$ (i.e., it is a normalized covariance). 

\subsection{GP formulation of RVM - Dual Gaussian Process} 
Using the results above, the RVM formulas in Eqs.~\eqref{RVMsolMean}--\eqref{Eq41var} can be rewritten in some way  such that they coincide with the corresponding GP solutions (dual GP formulation), when $k_{\texttt{dual}}(\x,\x')$ in Eq.~\eqref{AquiKeq} is used as a kernel function (see Section \ref{GPsect}).
 If we replace the expressions \eqref{AquiKeq}--\eqref{AquiKvect}--\eqref{AquiKmat} within the predictive-posterior RVM distribution  
$$
p(f(\x)|{\bf y}) =\mathcal{N}(f(\x)| \mu_{f|y}, \sigma_{f|y}^2),
$$
i.e., in Eqs.~\eqref{RVMsolMean}--\eqref{Eq41var}, we obtain the following expressions for the corresponding mean and variance functions,
\begin{gather}
\begin{split}
\mu_{f|y}&= \widehat{f}(\x)
= {\bf k}_{\texttt{dual}}(\x)\left({\bf K}_{\texttt{dual}}+\sigma_e^2 {\bf I}_N\right)^{-1}\by, \\
 \sigma_{f|y}^2
 &=k_{\texttt{dual}}(\x,\x)-{\bf k}_{\texttt{dual}}(\x) \left({\bf K}_{\texttt{dual}}+\sigma_e^2{\bf I}_N\right)^{-1}{\bf k}_{\texttt{dual}}(\x)^{\top}.
\end{split}
\end{gather}
We can observe that these formulas coincide with the mathematical expressions of the GP solutions in Eqs.~\eqref{GPmu1}--\eqref{GPvar1}. 
Replacing the expressions \eqref{AquiKeq}-\eqref{AquiKvect}-\eqref{AquiKmat} within the smoothing solution,  $p({\bf f}|{\bf y}) =\mathcal{N}({\bf f}| {\bm \mu}_{f|y}, {\bm \Sigma}_{f|y})$, i.e., in  Eqs.~\eqref{MuSmoothRVM}--\eqref{VarSmoothRVM}, we obtain
\begin{eqnarray}
{\bm \mu}_{f|y}&=&\widehat{\f}={\bf K}_{\texttt{dual}} \left({\bf K}_{\texttt{dual}}+\sigma_e^2 {\bf I}_N\right)^{-1}\by,  \nonumber  \\
{\bm \Sigma}_{f|y} &=&{\bf K}_{\texttt{dual}}- {\bf K}_{\texttt{dual}}^{\top}\left({\bf K}_{\texttt{dual}}+\sigma_e^2{\bf I}_N\right)^{-1}{\bf K}_{\texttt{dual}}.
 \end{eqnarray}
 Again, the formulas above coincide with the mathematical form of the GP solutions in Eqs.~\eqref{SmoothGP1}--\eqref{SmoothGP2}.  Then, a RVM can be interpreted as a GP using $k_{\texttt{dual}}(\x,\x')$ in Eq.~\eqref{AquiKeq} as kernel function. However, this implicit choice of  $k_{\texttt{dual}}(\x,\x')$ in general does not provide a good behavior of the predictive variance, as we remark in Section \ref{MostImportSect}.
 
\begin{figure}[!h]
	\centering
\centerline{	
	\subfigure[$N=5$]{\includegraphics[width=7cm]{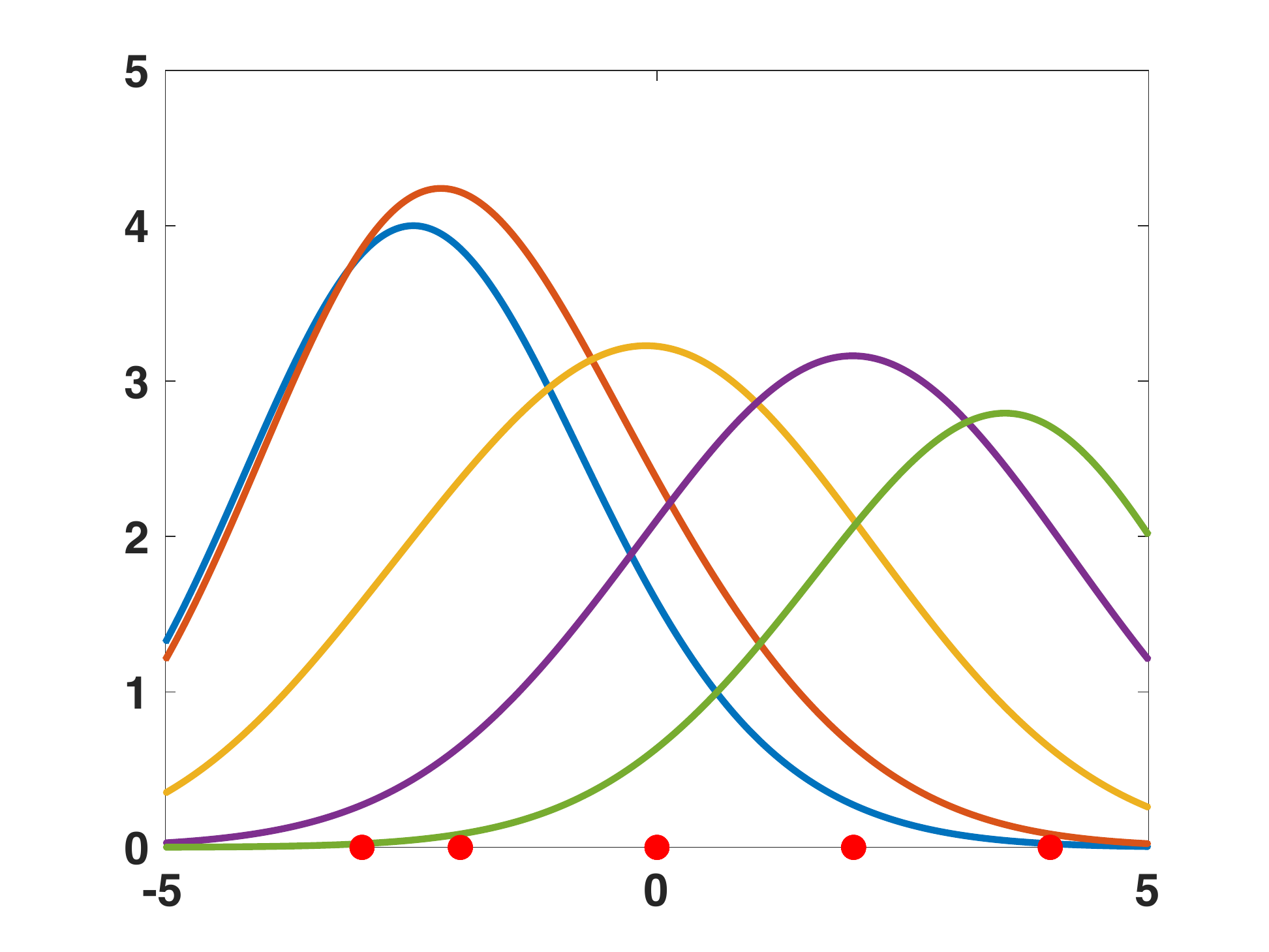}}
	\subfigure[$N=7$]{\includegraphics[width=7cm]{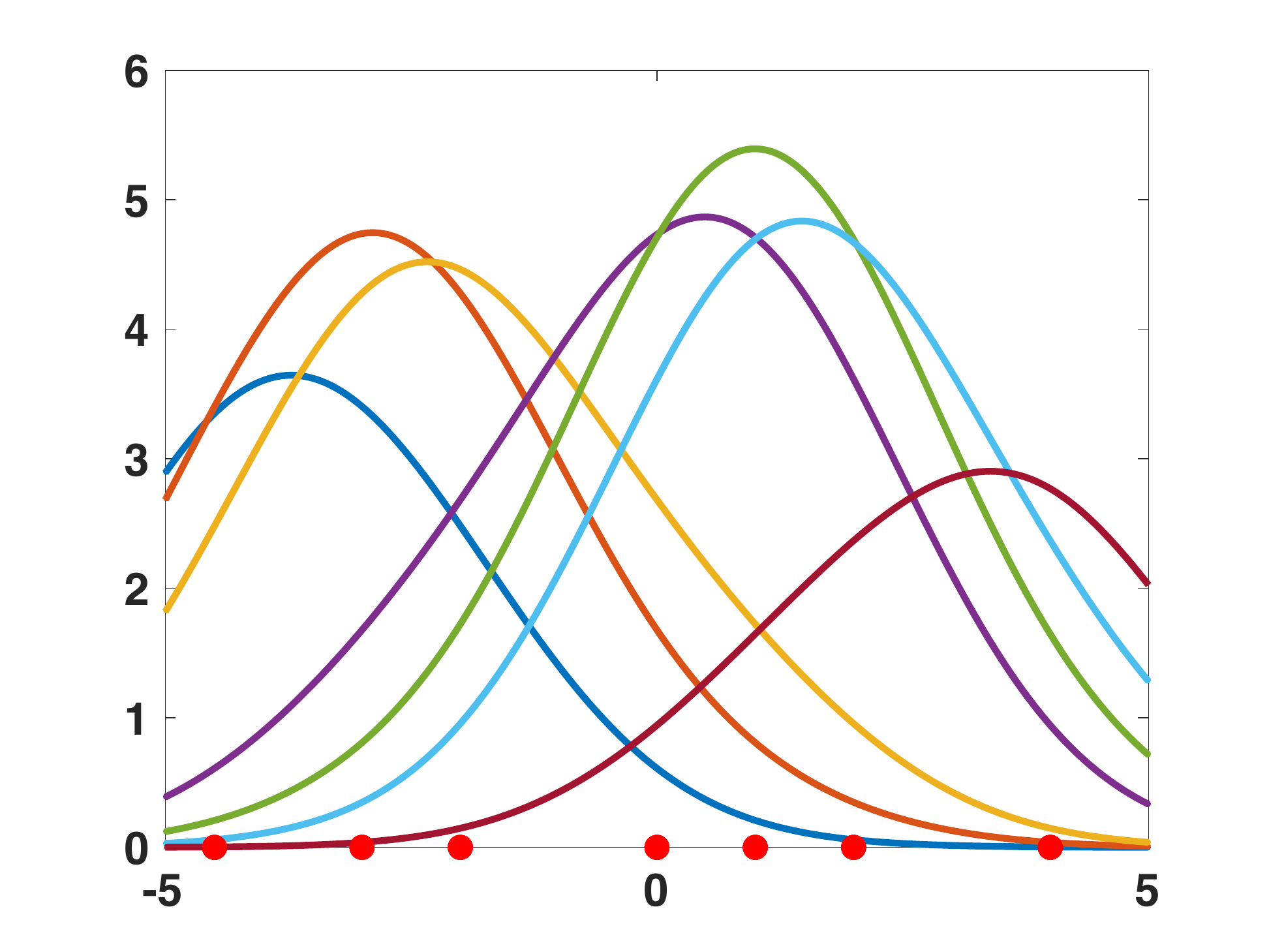}}	
}
	\caption{Examples of dual kernels $k_{\texttt{dual}}(x,x_n)$ obtained as in Eq.~\eqref{AquiKeq}, where {\bf (a)} $x_n\in\{-3, -2, 0,  2,  4 \}$, {\bf (b)}  $x_n\in\{-4.5,-3, -2, 0, 1, 2,  4 \}$  (shown with dots), and the bases $\ph(x,x_n)$ are Gaussian (with mean at $x_n$ and bandwidth $\lambda=5$). We have considered a diagonal covariance matrix ${\bm \Sigma}_\rho$ with all the elements in the diagonal equal to $2.25$. }
	\label{figDualKernel}
\end{figure}

 {\Remark Generally, the dual kernel function $k_{\texttt{dual}}(\x,\x')$ is not stationary. The choice of the bases functions $\psi(\x,\x')$ such that $k_{\texttt{dual}}(\x,\x')$ be stationary  is not straightforward. }
 \newline
 \newline
 Some examples of dual kernel functions are provided in Figure \ref{figDualKernel}. In the next section, we  summarize all the relationships and differences highlighted so far. Moreover, we provide more details regarding the predictive variance and the uncertainty analysis.

\section{Summary of relationships among RVMs, Q-GPs, and GPs} \label{MostImportSect}

\begin{figure}[!h]
	\centering
	\includegraphics[width=13cm]{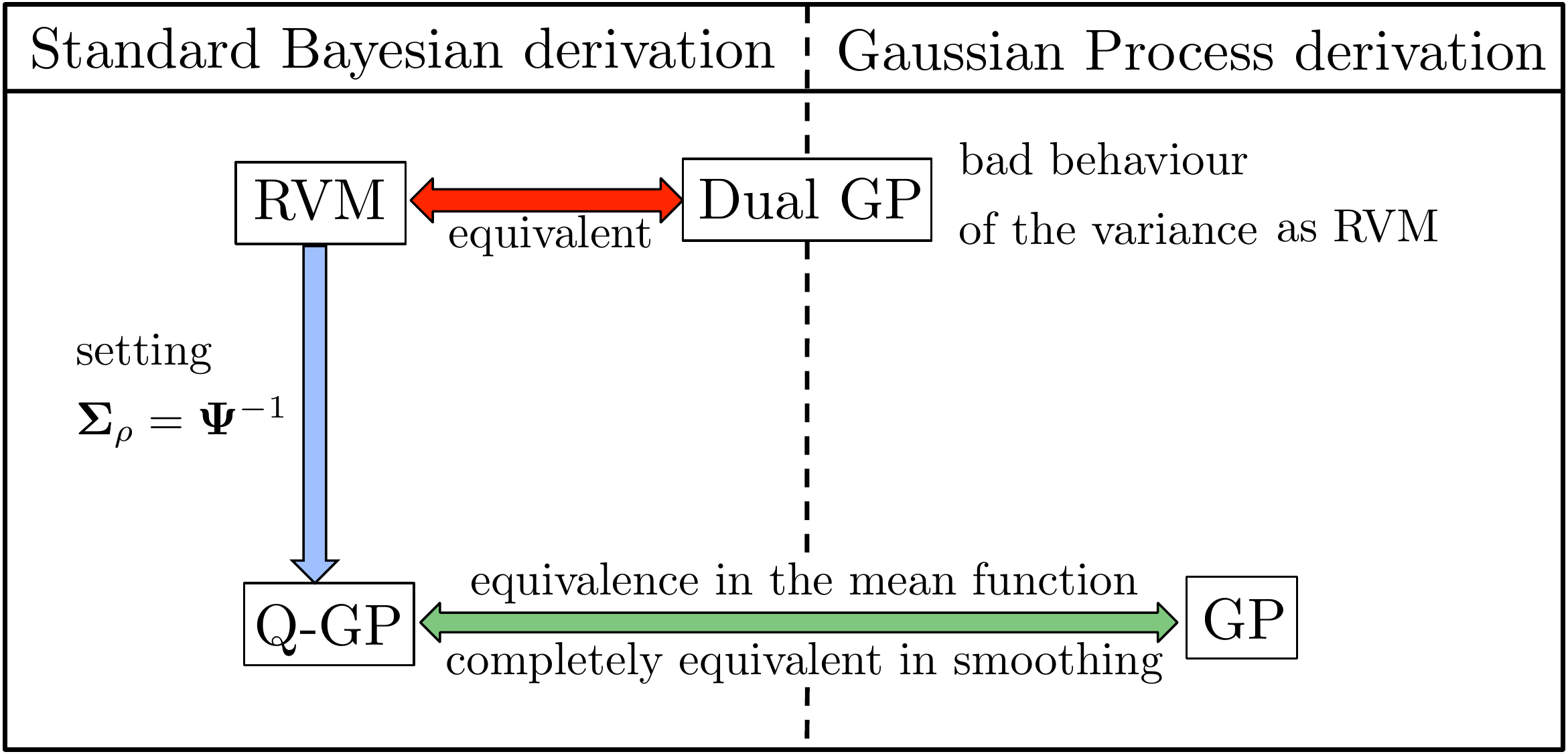}
	\caption{Graphical representation of the relationships among  the  different methods. }
	\label{fig1TEO}
\end{figure}

In all the considered methods, we have a complete characterization of posterior of the function $f(\x)$ for all $\x$. Moreover, in all the methods, we have  shown the generation of random functions from (direct or induced) priors and/or posteriors. 
In this section, we recall the connections among all the methods. Below, we enumerate some important observations highlighted so far:
\begin{itemize}
\item Q-GP is a special case of RVM setting ${\bm \Sigma}_p={\bm \Psi}^{-1}$.
\item The posterior mean function $\widehat{f}(\x)$ of Q-GP always coincides  with the posterior mean function $\widehat{f}(\x)$ of a standard GP.  
\item Given the previous point, in regression, Q-GP and standard GP differs only for the expression of their posterior-predictive variance (see Table \ref{table_SummaryMethods}).  
\item In the  smoothing scenario, i.e., considering only the data inputs $\{\x_n\}_{n=1}^N$,  Q-GP and GP are perfectly equivalent, i.e., they have  the same posterior/predictive distributions (Gaussian with the same mean and variance). See Table \ref{table_SummaryMethods2}, for further details.
\item In Q-GP and GP, the design matrix ${\bm \Psi}$ must be symmetric, i.e., ${\bm \Psi}={\bm \Psi}^{\top}$, and positive definite. This is not required in RVM. 
\end{itemize}
 These relationships among the methods are graphically represented in Figure \ref{fig1TEO}.  Summaries of the main formulas for regression, smoothing and interpolation are given also in Tables \ref{table_SummaryMethods},  \ref{table_SummaryMethods2} and \ref{table_SummaryMethods3}, respectively. 
In Table \ref{table_SummaryMethods0}, we summarize the expressions related to ${\bm \rho}$. It is important remark that GP follows a different mathematical derivation (without assuming a prior over ${\bm \rho}$) with respect to Q-GP and RVM.  The next remark is a consequence of this fact.  
 {\Remark All the random functions generated from RVM priors and/or RVM posteriors (see Section \ref{RVMgeneration}) can be always expressed as linear combinations of the basis functions, as in Eq. \eqref{aquiF}. In the GP case, the random functions (see Section \ref{GPgeneration}) cannot be written as in Eq. \eqref{aquiF}, and only the GP posterior mean can be expressed as in Eq. \eqref{aquiF}.
 }
 \newline
  \newline
Finally, note that Q-GP and standard GP does not induce sparsity, unlike RVM (when we learn the sclae parameters of the prior).  Q-GP is a special case of RVM (which can induce sparsity) but the covariance matrix of the prior is set ${\bm \Sigma}_\rho={\bm \Psi^{-1}}$. Therefore, in order to keep the sparsity inducing property, we should employ a kernel/basis function which such that satisfies ${\bm \Psi}={\bm \Psi}^{\top}$ and provides different elements in principal diagonal, and these values could change after learning  the hyper-parameters of the kernel/basis functions.

{\Remark\label{REM_VAR_QGPGP} The difference between the predictive variances of Q-GP and GP in regression is only the first part, ${\bm \psi}({\bf x})^{\top}{\bf \Psi}^{-1}{\bm \psi}({\bf x})$ for Q-GP, and $\psi(\x,\x)$ for GP (see Table \ref{table_SummaryMethods}). The second factor, i.e.,  
$-{\bm \psi}({\bf x})^{\top}({\bm \Psi} +\sigma_e^2 {\bf I}_N)^{-1}{\bm \psi}({\bf x})$, is common in both expressions. }

\begin{table}[!h]
\centering
\small
\caption{Posterior mean and covariance matrix of ${\bm \rho}$. N.B. In Q-GP and GP, we have ${\bm \Psi}={\bm \Psi}^\top$. }
\vspace{-0.3cm}
\begin{center}
	\begin{tabular}{|l|c|c| }
    \hline
\multirow{2}{*}{{\bf Method}} & \multirow{2}{*}{{\bf Mean} $\widehat{{\bm \rho}}={\bm \mu}_{\rho|y}$}  & \multirow{2}{*}{{\bf Covariance matrix} ${\bm \Sigma}_{\rho|y}$} \\
 &  & \\
\hline
\hline
& & \\ 
RVM  &$\boldsymbol{\Sigma}_\rho{\bm \Psi}^{\top} \left({\bm \Psi}\boldsymbol{\Sigma}_\rho{\bm \Psi}^\top+\sigma_e^2 {\bf I}_N\right)^{-1}\by$  &  ${\bm \Sigma}_\rho-{\bm \Sigma}_\rho{\bm \Psi}^{\top}\left({\bm \Psi}{\bm \Sigma}_\rho{\bm \Psi}^{\top}+\sigma_e^2{\bf I}_N\right)^{-1}{\bm \Psi}{\bm \Sigma}_\rho$ \\
& & \\ 
\hline
& & \\ 
Q-GP& $\left({\bm \Psi}+\sigma_e^2 {\bf I}_N\right)^{-1}\by$   & ${\bf \Psi}^{-1}-\left({\bm \Psi}+\sigma_e^2{\bf I}_N\right)^{-1}$ \\
& & \\ 
\hline
& & \\ 
GP & $\left({\bm \Psi}+\sigma_e^2 {\bf I}_N\right)^{-1}\by$ &  ----- \\ 
& &   \\
   \hline 
\end{tabular}
\end{center}
\label{table_SummaryMethods0}
\end{table}


\begin{table}[!h]
\centering
\small
\caption{Regression formulas. N.B. In Q-GP and GP, we have ${\bm \Psi}={\bm \Psi}^\top$. }
\vspace{0.1cm}
	\begin{tabular}{|l|l|l| }
    \hline
\multirow{2}{*}{{\bf Method}} & \multirow{2}{*}{{\bf Mean} $\widehat{f}(\x)=\mu_{f|y}(\x)$}  & \multirow{2}{*}{{\bf Variance} $\sigma^2_{f|y}(\x)$} \\
 & & \\
\hline
\hline
 & & ${\bm \psi}({\bf x})^{\top}{\bm \Sigma}_\rho{\bm \psi}({\bf x})-$ \\
RVM & 
${\bm \psi}({\bf x})^{\top}\boldsymbol{\Sigma}_\rho{\bm \Psi}^{\top} \left({\bm \Psi}\boldsymbol{\Sigma}_\rho{\bm \Psi}^\top+\sigma_e^2 {\bf I}_N\right)^{-1}\by$ & 
${\bm \psi}({\bf x})^{\top}{\bm \Sigma}_\rho{\bm \Psi}^{\top}\left({\bm \Psi}{\bm \Sigma}_\rho{\bm \Psi}^{\top}+\sigma_e^2{\bf I}_N\right)^{-1}{\bm \Psi}{\bm \Sigma}_\rho{\bm \psi}({\bf x})$ \\
 & & \\
Q-GP  & 
${\bm \psi}({\bf x})^{\top}({\bm \Psi} +\sigma_e^2 {\bf I}_N)^{-1}\y $ & ${\bm \psi}({\bf x})^{\top}{\bf \Psi}^{-1}{\bm \psi}({\bf x})- {\bm \psi}({\bf x})^{\top}\left({\bm \Psi}+\sigma_e^2{\bf I}_N\right)^{-1}{\bm \psi}({\bf x})$ \\
& & \\
GP  & 
${\bm \psi}({\bf x})^{\top}({\bm \Psi} +\sigma_e^2 {\bf I}_N)^{-1}\y $ &
$\psi(\x,\x)-{\bm \psi}({\bf x})^{\top}({\bm \Psi} +\sigma_e^2 {\bf I}_N)^{-1}{\bm \psi}({\bf x})$ \\
   \hline 
\end{tabular}
\label{table_SummaryMethods}
\end{table}

\begin{table}[!h]
\centering
\small
\caption{Smoothing formulas. N.B. In Q-GP and GP, we have ${\bm \Psi}={\bm \Psi}^\top$.}
\vspace{0.1cm}
	\begin{tabular}{|l|l|l| }
    \hline
\multirow{2}{*}{{\bf Method}} & \multirow{2}{*}{{\bf Mean vector} $\widehat{\f}={\bm \mu}_{f|y}$}  & \multirow{2}{*}{{\bf Covariance matrix} ${\bm \Sigma}_{f|y}$} \\
 & & \\
\hline
\hline
 & &Ê\\
RVM & 
${\bm \Psi}\boldsymbol{\Sigma}_\rho{\bm \Psi}^\top \left({\bm \Psi}\boldsymbol{\Sigma}_\rho{\bm \Psi}^\top+\sigma_e^2 {\bf I}_N\right)^{-1}\by$ & 
${\bm \Psi}{\bm \Sigma}_\rho{\bm \Psi}^{\top}- {\bm \Psi}{\bm \Sigma}_\rho{\bm \Psi}^{\top}\left(\sigma_e^2{\bf I}_N+{\bm \Psi}{\bm \Sigma}_\rho{\bm \Psi}^{\top}\right)^{-1}{\bm \Psi}{\bm \Sigma}_\rho{\bm \Psi}^{\top}$ \\
 & & \\
Q-GP & 
${\bm \Psi}({\bm \Psi} +\sigma_e^2 {\bf I}_N)^{-1}\y$ & ${\bf \Psi}-{\bf \Psi}\left({\bm \Psi}+\sigma_e^2{\bf I}_N\right)^{-1}{\bf \Psi}$ \\
 & & \\
GP & 
${\bm \Psi}({\bm \Psi} +\sigma_e^2 {\bf I}_N)^{-1}\y$ &
 ${\bf \Psi}-{\bf \Psi}\left({\bm \Psi}+\sigma_e^2{\bf I}_N\right)^{-1}{\bf \Psi}$ \\
   \hline 
\end{tabular}
\label{table_SummaryMethods2}
\end{table}

\begin{table}[!h]
\centering
\small
\caption{Interpolation formulas. N.B. In Q-GP and GP, we have ${\bm \Psi}={\bm \Psi}^\top$.}
\vspace{0.1cm}
	\begin{tabular}{|l|l|l| }
    \hline
\multirow{2}{*}{{\bf Method}} & \multirow{2}{*}{{\bf Mean} $\widehat{f}(\x)$}  & \multirow{2}{*}{{\bf Variance} $\sigma^2_{f|y}(\x)$} \\
 & & \\
\hline
\hline
 & &  \\
RVM & 
${\bm \psi}({\bf x})^{\top}{\bm \Psi}^{-1}\y$ & 
$0$ for all $\x$\\
 & & \\
Q-GP  & 
${\bm \psi}({\bf x})^{\top}{\bm \Psi}^{-1}\y$ & $0$ for all $\x$  \\
& & \\
GP  & 
${\bm \psi}({\bf x})^{\top}{\bm \Psi}^{-1}\y $ &
$\psi(\x,\x)-{\bm \psi}({\bf x})^{\top}{\bm \Psi}^{-1}{\bm \psi}({\bf x})$ \\
   \hline 
\end{tabular}
\label{table_SummaryMethods3}
\end{table}

\subsection{Advantages and weaknesses}
	The advantage of RVM is that we have fewer restrictions in choosing the bases $\psi_n$, i.e., we have more flexibility in this choice. With Q-GP and GP, we need some function $\psi$ such that the matrix ${\bm \Psi}$ be invertible and positive definite, since ${\bm \Psi}$ must be interpreted as a covariance matrix. For instance, in RVM, we can employ directly $N$ different bases $\psi_n$, each one with a different analytical form and different parameters. Whereas, with Q-GP and GP, first of all we should check if the resulting matrix ${\bm \Psi}$ is positive definite,  for any possible values of the entries and the parameters. In all cases, RVM, Q-GP and GP,  the mean solution can be expressed as linear combination of $N$ nonlinearities. Hence, in both cases, the flexibility of the solution grows with the number of data (i.e., $N$).

{\Remark The main benefit of the RVM approach is that we have fewer restrictions in the choice  of the nonlinearities $\psi_n$ that can be also different for each data input $\x_n$, since we do not need that ${\bm \Psi}$ be symmetric. }
 
{\Remark The main advantage of GP with respect to the other methods, is that we directly decide the covariance function ensuring, for instance, stationary and other statistical properties (when required). This has also another important consequence:  the behavior of GP predictive variance  has a natural/intuitive behavior (as we shown below), unlike the predictive variances of RVM and Q-GP \cite{Szeliski1987RegularizationUF,Poggio20,Silverman85,RasmussenQuinonero05}. 
}
 


\subsection{Variance behavior}

\begin{figure}[!h]
	\centering
\centerline{	
	\subfigure[]{\includegraphics[width=7cm]{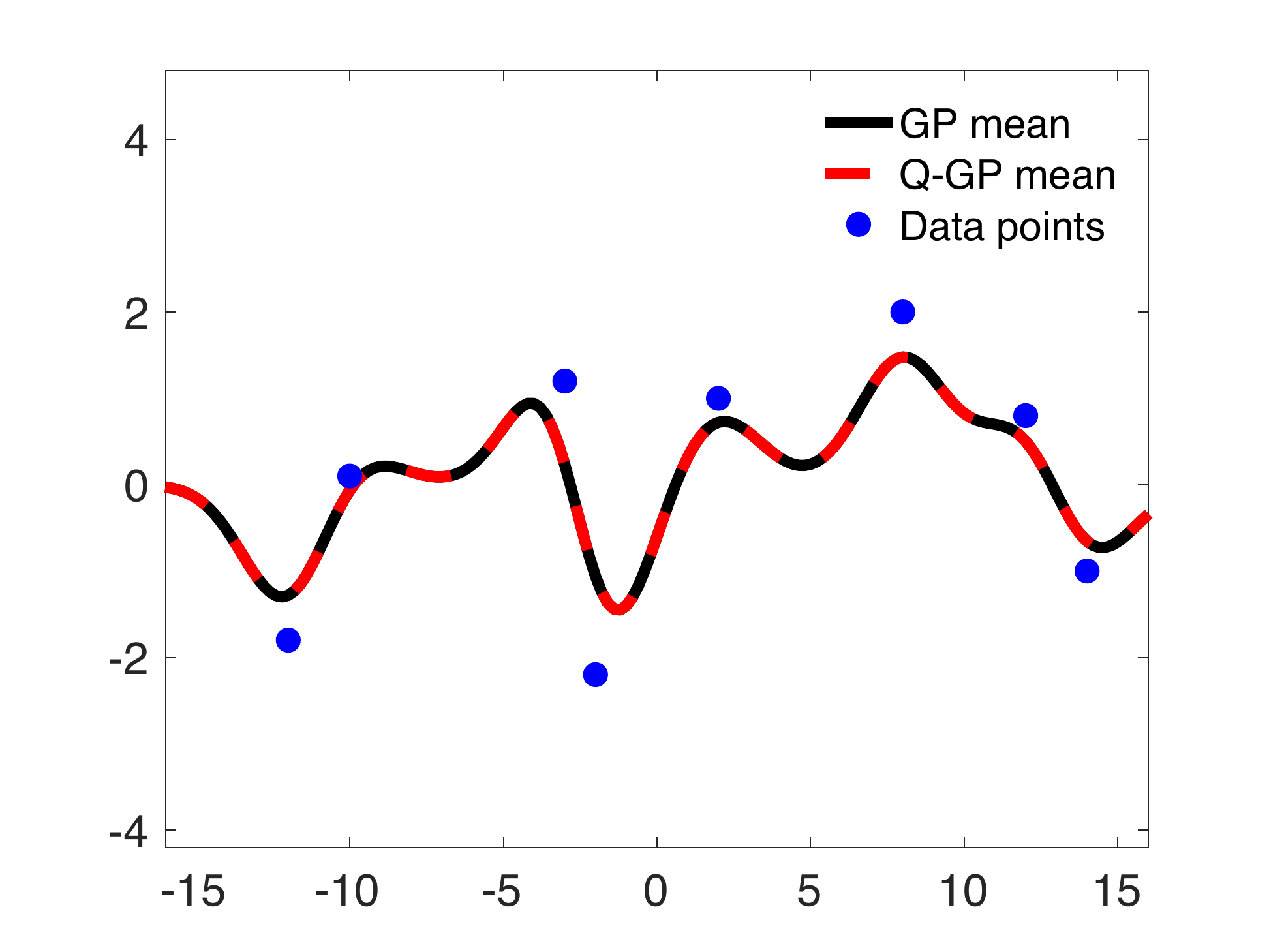}}
	\subfigure[]{\includegraphics[width=7cm]{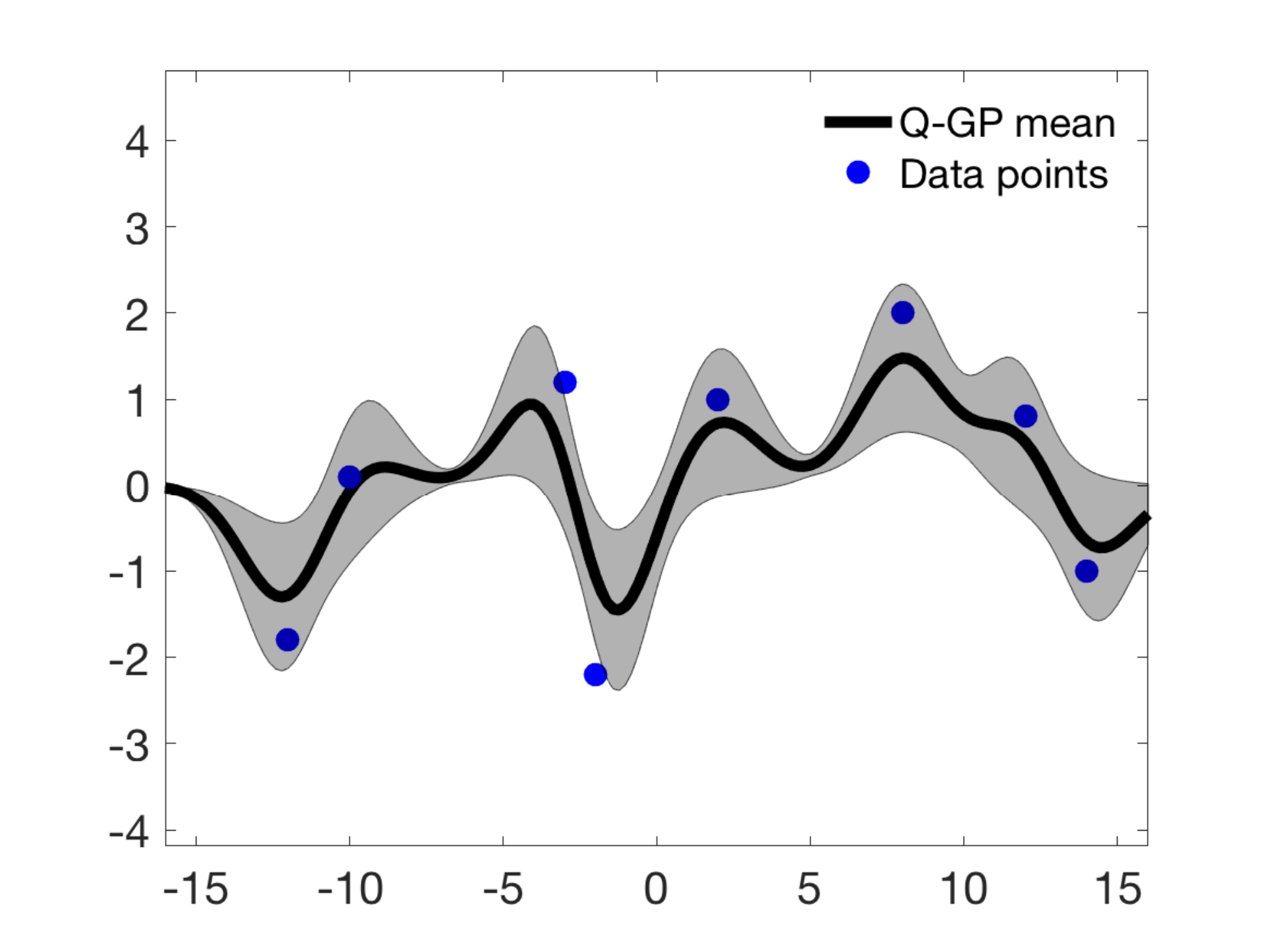}}
}
\centerline{	
	\subfigure[]{\includegraphics[width=7cm]{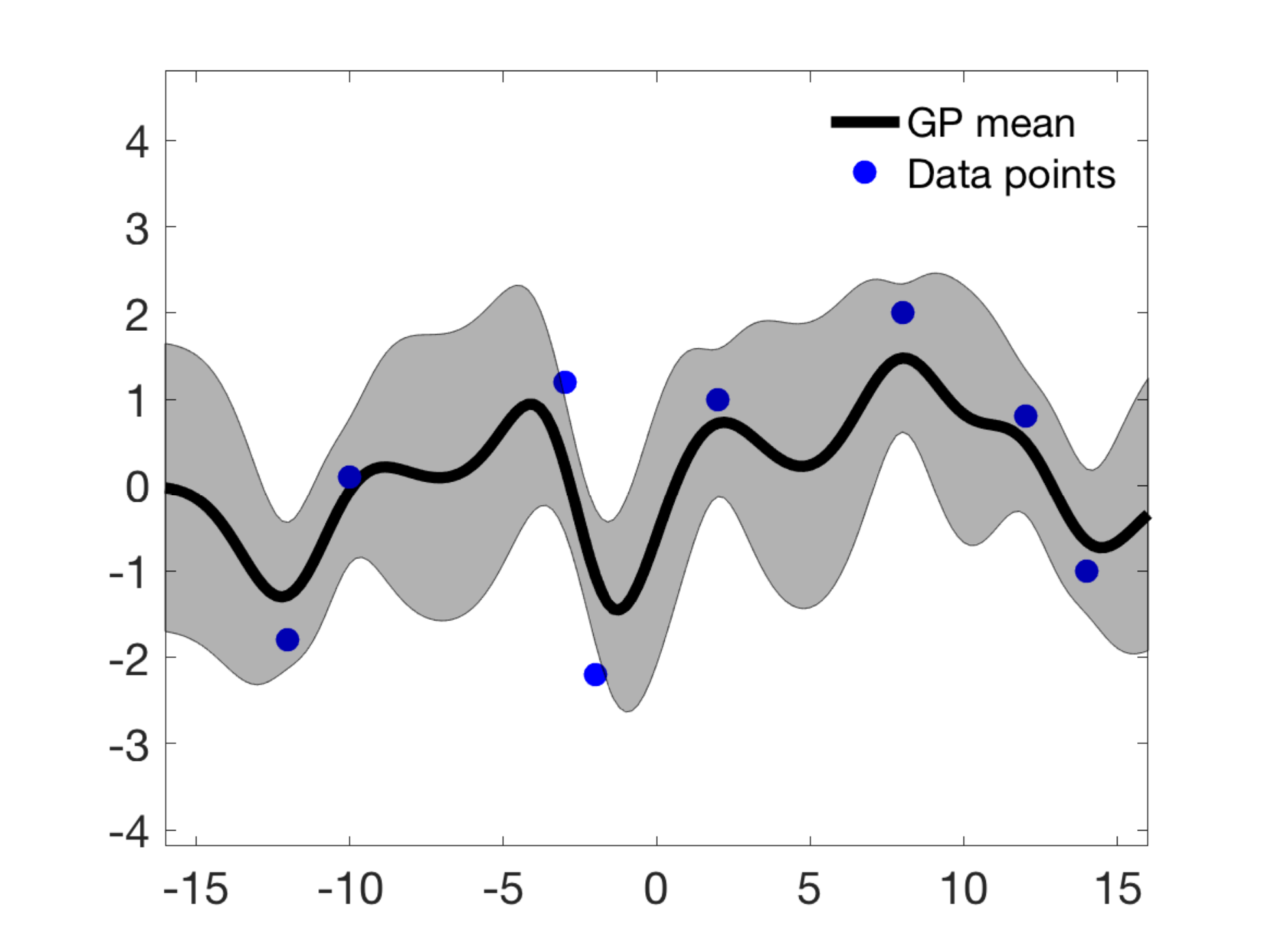}}
	\subfigure[]{\includegraphics[width=7cm]{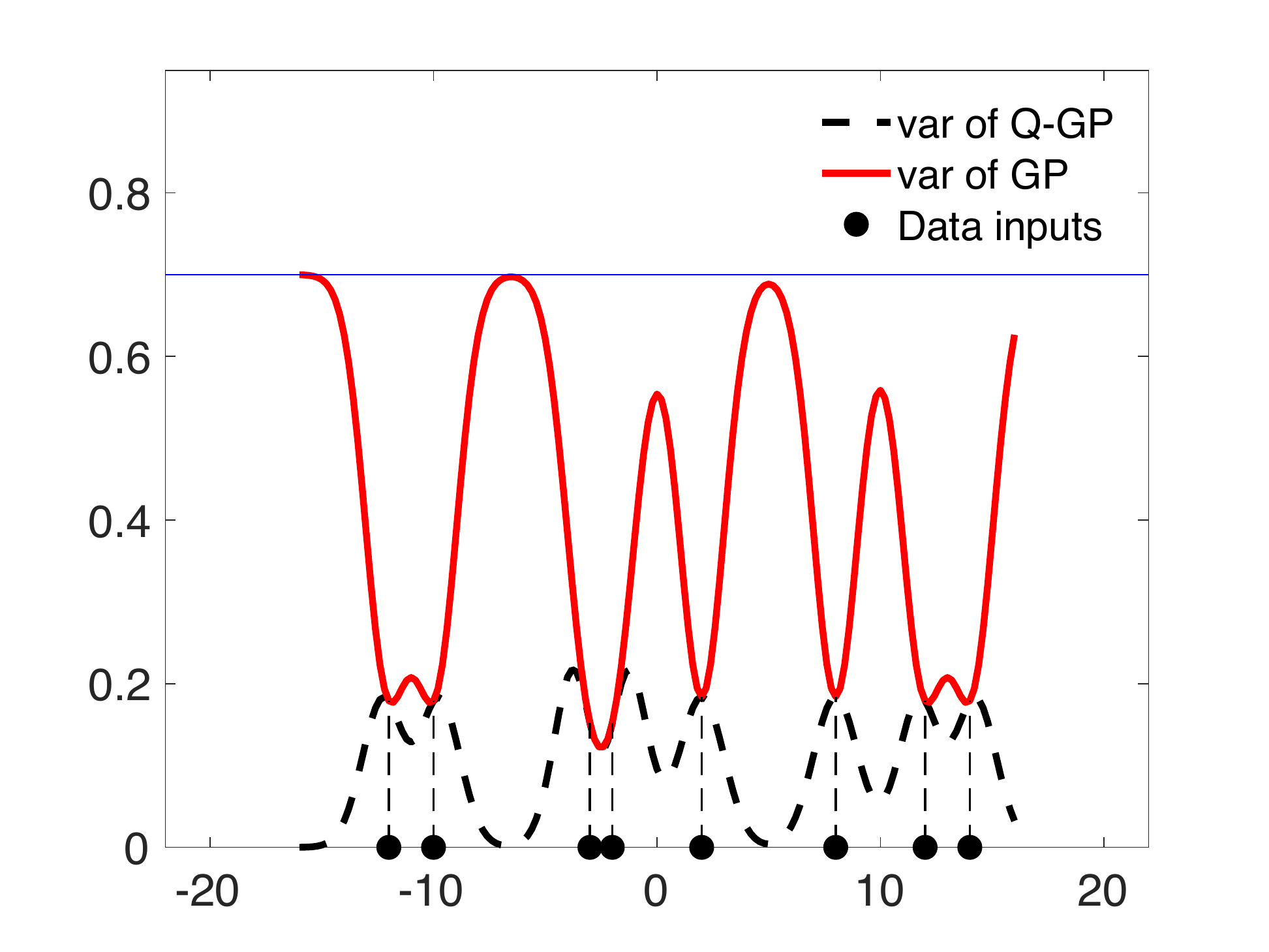}}
}
	\caption{Variance Behaviour (with $\sigma_e=0.5$). {\bf (a)-(b)-(c)} show the posterior (predictive) mean $\widehat{f}(x)$ of GP and Q-GP. The shaded areas represents $\widehat{f}(x)\pm 2 \sqrt{\sigma_{f|y}^2(x)}$. Note the posterior means coincides. {\bf (d)} shows the corresponding variances $\sigma_{f|y}^2(x)$ as function of $x$. The horizontal blue line denotes the value of $a=0.7$. }
	\label{fig2QGPvsGP}
\end{figure}

\begin{figure}[!h]
	\centering
\centerline{	
	\subfigure[RVM and Q-GP]{\includegraphics[width=8cm]{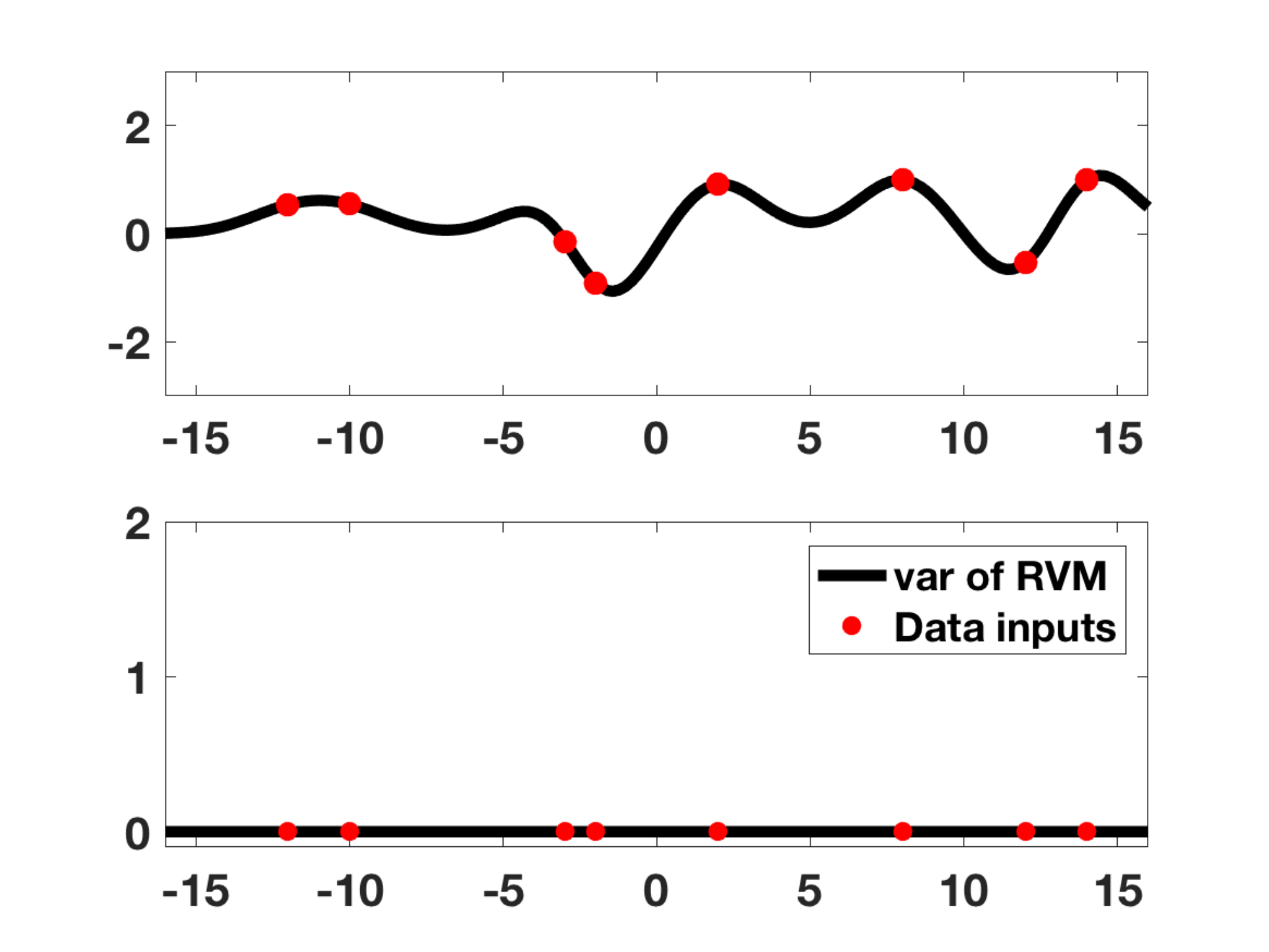}}
	\subfigure[GP]{\includegraphics[width=8cm]{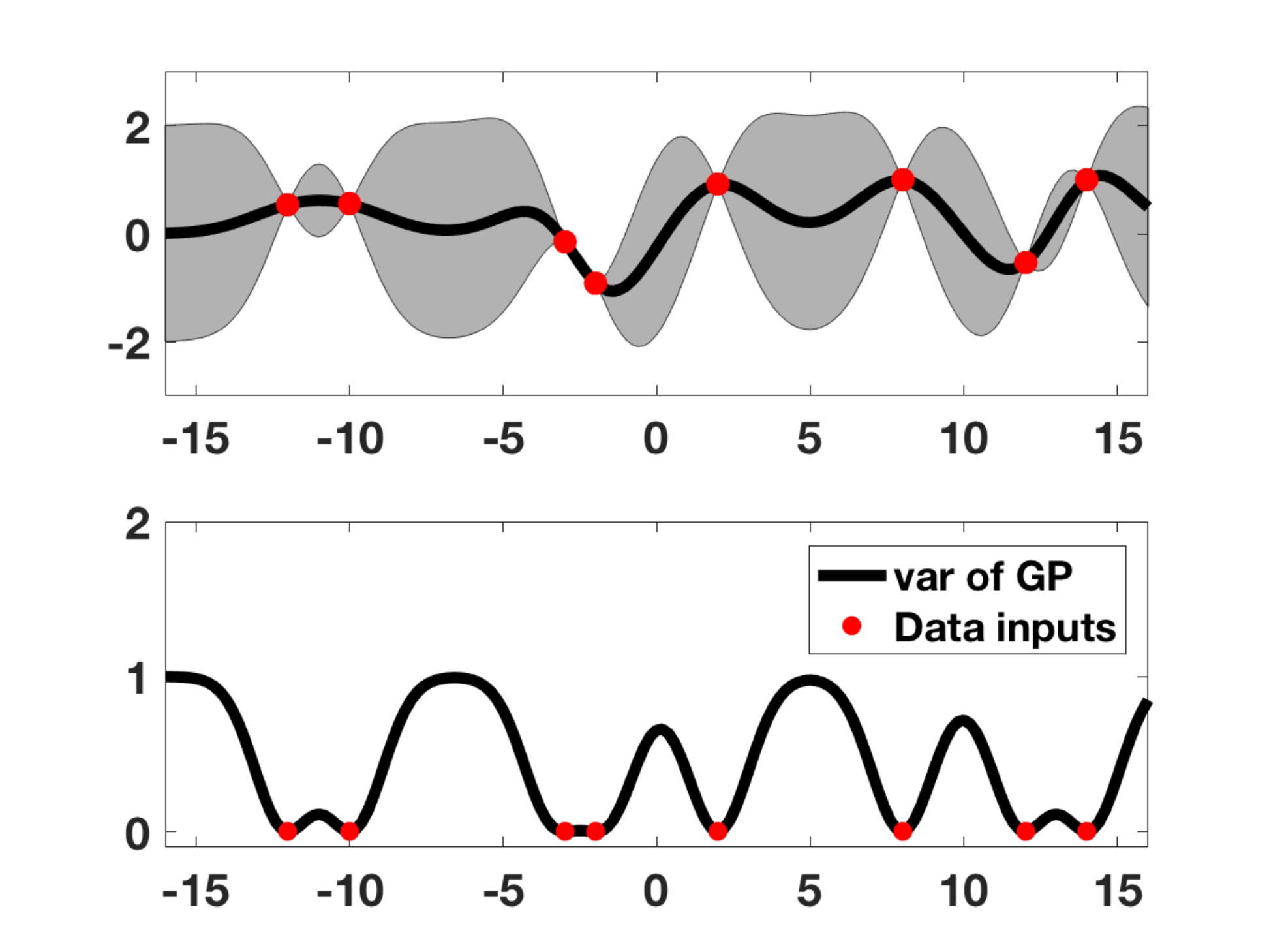}}
}
	\caption{ Interpolation case, i.e., $\sigma_e=0$ (and $a=1$).  {\bf (a)} In this case, RVM and Q-GP provides the same solution with a null predictive variance, i.e., $\sigma_{f|y}^2(x)=0$.  The posterior/predictive mean  $\widehat{f}(x)$ of RVM/Q-GP is also given. {\bf (b)} Predictive  mean and variance of a GP when $\sigma_e=0$ (interpolation).  The shaded area represents $\widehat{f}(x)\pm 2 \sqrt{\sigma_{f|y}^2(x)}$.  The corresponding variances $\sigma_{f|y}^2(x)$ as function of $x$ is also given below. Note that $\sigma_{f|y}^2(x)$ is zero only at the data inputs.  }
	\label{figInterpVarRVM_GP}
\end{figure}

A intuitive and natural behavior of the predictive variance is the following: it should be smaller at $\x$ close to the data inputs $\x_n$, and greater far away from the data points. This usually happens with a GP with a reasonable choice of the kernel function.  With RVM (and Q-GP) and  localized bases, the behavior is arguably non-intuitive: the predictive variance is greater close to $\x_n$, and smaller far away from the data inputs.
\newline
\newline
{\bf Comparison of Q-GP and GP variances.} We know that Q-GP is a special case of RVM, which coincides with GP in the mean of te posterior/predictive function. In order to provide a comparison between Q-GP and GP, consider a one dimensional example, $x\in\mathbb{R}$, with the following basis/kernel
\begin{equation}\label{aquiEQa}
k(x, z)=a \exp\left(-\frac{(x-z)^2}{\lambda}\right),
\end{equation}
with $a=0.7$ and $\lambda=2$. Given a set of data points, we have applied the  Q-GP and GP methods, considering $\sigma_e=0.5$ (in Figure \ref{fig2QGPvsGP}). In Figure \ref{fig2QGPvsGP}(a), we show the data points and the  posterior means of Q-GP and GP. As expected, they perfectly coincide in both cases. Figure~\ref{fig2QGPvsGP}(b) depicts the mean of Q-GP $\widehat{f}(x)$ and $\widehat{f}(x)\pm 2 \sqrt{\sigma_{f|y}^2(x)}$ with a shaded area.  Figure  \ref{fig2QGPvsGP}(c) depicts the mean $\widehat{f}(x)$ of the standard GP and  $\widehat{f}(x)\pm 2 \sqrt{\sigma_{f|y}^2(x)}$ with a shaded area. The corresponding variances $\sigma_{f|y}^2(x)$ as function of $x$ are shown in Figure \ref{fig2QGPvsGP}(d). We can observe that the the variance of the GP is smaller closer to the data points (as expected). The opposite occurs with Q-GP. The two variance functions coincide exactly in the data points $\{x_n\}_{n=1}^N$. Indeed, in the smoothing scenario, Q-GP and GP are perfectly equivalent. Furthermore, observe that the variance of GP is always greater than the variance of Q-GP. Finally, let us compare  Figures  \ref{fig2QGPvsGP}(d) and \ref{figInterpVarRVM_GP}, for instance.   Note that, when $\sigma_e \rightarrow 0$, the variance of Q-GP vanishes to zero for all  $x$, i.e.,  $\sigma_{f|y}^2(x) \rightarrow 0$ $\forall x$. Whereas,  when $\sigma_e \rightarrow 0$, the variance of GP goes to zero only in  $\{x_n\}_{n=1}^N$, i.e., in the data points $\sigma_{f|y}^2(x_n)=0$ for all $n$.  The interpolation case is given in Figure \ref{figInterpVarRVM_GP} (see also Table \ref{table_SummaryMethods3}). 

\begin{figure}[!h]
	\centering
	
\centerline{	
	\subfigure[GP.]{\includegraphics[width=7cm]{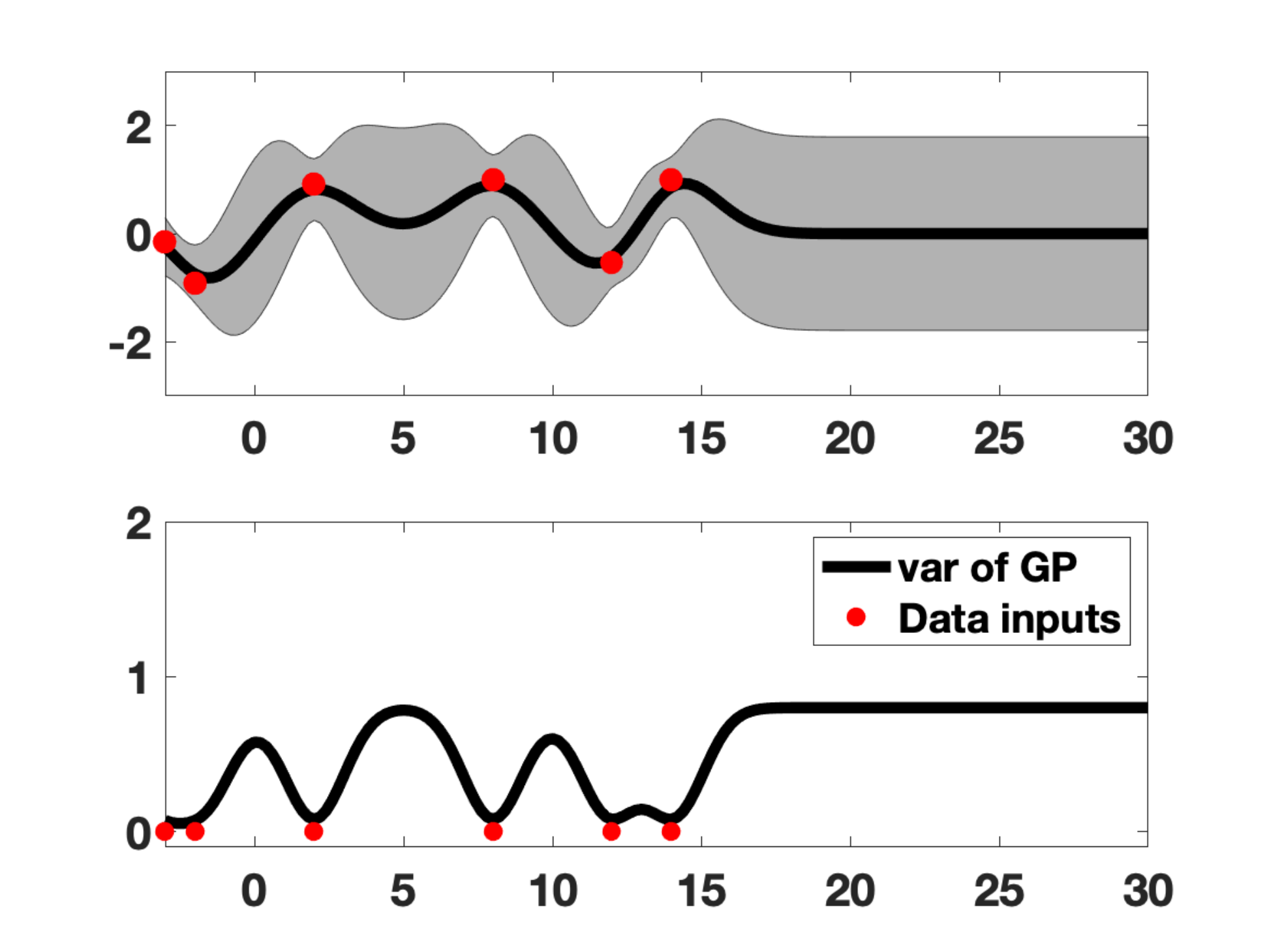}}
	\subfigure[Q-GP.]{\includegraphics[width=7cm]{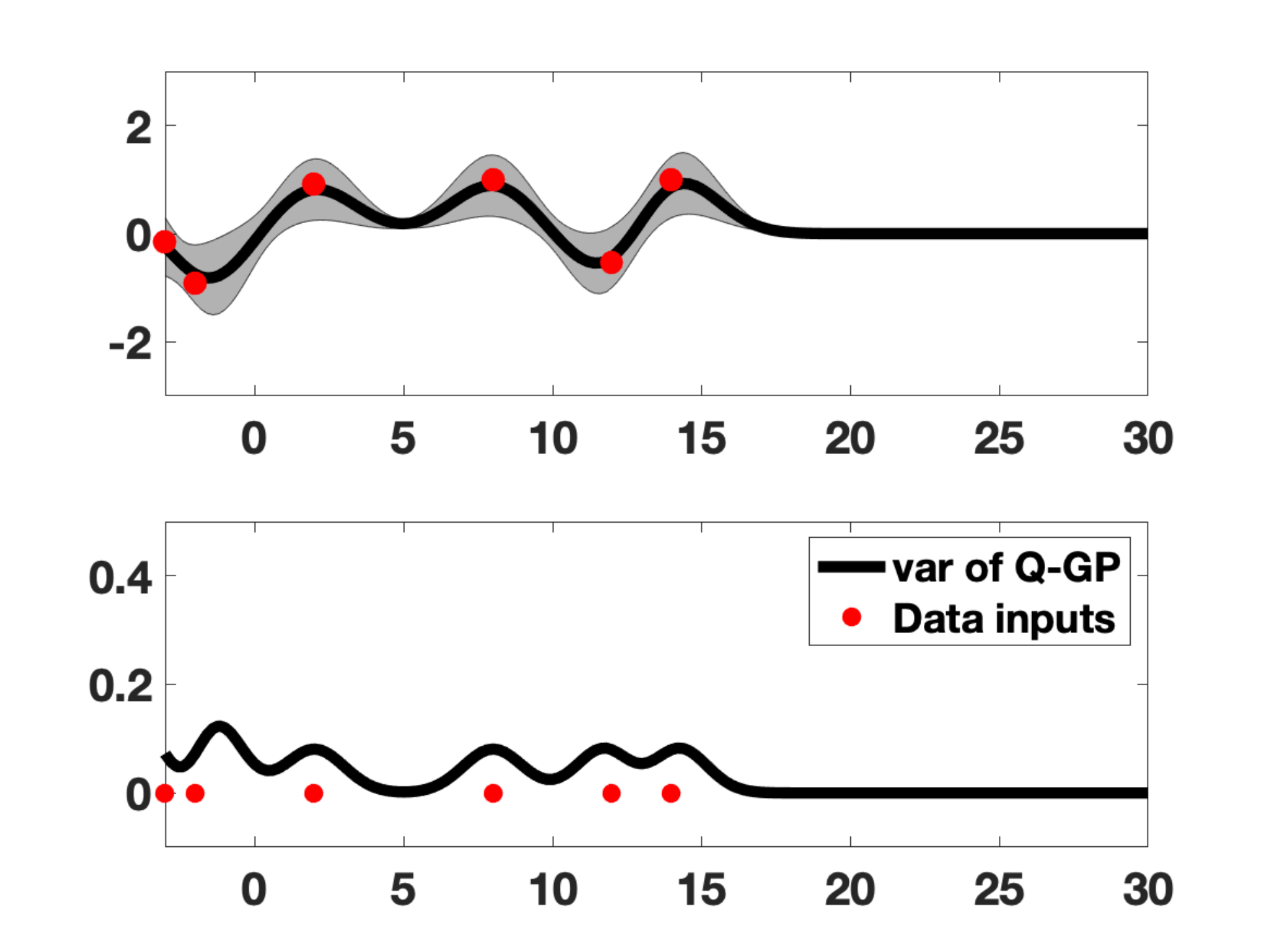}}
}
	\caption{The behavior of variance at places on the space far from the training points.  (with $\sigma_e=0.3$ and $a=0.8$), {\bf (a)} for GP and for {\bf (b)} Q-GP.  We can observe that, on the right side of the figures, the GP mean converges to the prior mean (that we assume to be zero), and the GP variance approaches the {\it a-priori signal variance} $a=0.8$ (see Section \ref{SuperSECTGhahramani11}). Whereas, in the Q-GP case,  both mean and variance converge to zero. }
	\label{fig2QGPvsGP_prediction}
\end{figure}

{\Remark Note that, when the value $x$ is far from the input points $x_i$, the posterior-predictive variance of the GP, $\sigma_{f|y}^2(x)$, tends to the value $a$ in Eq. \eqref{aquiEQa}, which is the maximum value reached by the $k(x, z)$ and represents the assumed a-priori variance.}
\newline
\newline
 The mathematical reason of these different variance behaviors among RVM/Q-GP and GP requires additional further studies. Moreover, in the example of Figure \ref{figInterpVarRVM_GP}, we can observe that the variance of Q-GP is  smaller (or equal in $\x=\x_i$) than  the variance of GP, Namely, in this scenario, it seems that $\psi(x,x)\geq  {\bm \psi}(x)^{\top}{\bf \Psi}^{-1}{\bm \psi}(x)$ (see Remark \ref{REM_VAR_QGPGP}). However, this is just a conjecture since it could depend on the type of kernel employed. 
\newline
\newline
{\bf Predictive variance far from training points.} Let us study the behavior of GP and Q-GP at places on the space far from the training points (e.g., for future predictions in a time series). In Figure \ref{fig2QGPvsGP_prediction}(a), we can observe that the GP mean $\widehat{f}(\x)$ converges to the prior mean, that we assume to be zero, and the GP variance $\sigma_{f|y}^2(x)$ approaches the {\it a-priori signal variance}, which is $a=0.8$ in this figure. See Section \ref{SuperSECTGhahramani11}, for the interpretation of the GP hyper-parameters. Whereas, in Figure \ref{fig2QGPvsGP_prediction}(b),   the Q-GP mean and Q-GP variance converge to zero.

\subsection{The legend of infinite bases}
Several authors show that the squared exponential kernel function, defined (scalar $x \in \R$ for simplicity) as 
 $$
  \psi(x_i,x_j)=\exp\left(-\frac{(x_i-x_j)^2}{2\sqrt{2}\lambda}\right),
 $$
 can also be obtained by expanding the input into a feature space represented by an 
{\it infinite} network defined by Gaussian-shaped basis functions  $\phi_c(x)=\exp\left(-\frac{(x-c)^2}{2\lambda}\right)$,
where $c$ denotes the centre of the basis function. Let us consider $M$ bases centered in different values $c$. Assuming the covariance matrix of the prior density as ${\bm\Sigma}_p=\sigma_p^2 {\bf I}_M$, the induced kernel can be written as 
 $$
  k_M(x_i,x_j)=\sigma_p^2 \sum_{c=1}^M  \phi_c(x_i) \phi_c(x_j).
 $$
 It is possible to show that $\lim\limits_{M\rightarrow \infty}   k_M(x_i,x_j)\propto \psi(x_i,x_j)$  (see, e.g., \cite{rasmussen2003gaussian}).
\newline
This result can lead to misleading conclusions. For instance, one could state that {\it`` to obtain the GP flexibility and performance, with a standard Bayesian formulation, we need infinite bases''}. This statement is not true, or only partially true.  Indeed, regarding the posterior/predictive mean function, we know that we can obtain exactly the GP solution with  a standard Bayesian formulation setting $M=N$ (finite) and  using  a {\it suitable} covariance prior over the weights,  ${\bm\Sigma}_p={\bm \Psi}^{-1}$. On the other hand, we cannot obtain the same posterior/predictive variance, at least not with a finite number of bases.  


\subsection{Marginal likelihood and parameter learning}  \label{Aqui_ML}
 The marginal likelihood (a.k.a., Bayesian evidence) is  defined as 
 \begin{equation}
p({\bf y}) =\int_{\mathbb{R}^{N}} p(\y|{\bf f})p({\bf f}) {\bf d}{\bf f}.
\end{equation}
where $p({\bf f})$ is the prior over the $N\times 1$ vector $\f$ given the considered probabilistic model. The marginal likelihood represents the probability of data given the model $\mathcal{M}$ and its parameters, $p({\bf y}) =p({\bf y}|\mathcal{M})$, hence it is useful for model selection purposes. 
 For instance, it  can be used in order to tune of the parameters and hyper-parameters of the model denoted as ${\bm \theta}$. In the vector  ${\bm \theta}=[{\bm \lambda},\sigma_e^2]$ we include all the parameters ${\bm \lambda}$ of the nonlinearities $\psi_n$,  as well as the  parameters need for defining the prior densities and the power of the noise perturbation $\sigma_e^2$.  Therefore, in this case, a more complete notation is the following
 \begin{equation}
p({\bf y}|{\bm \theta}) =\int_{\mathbb{R}^{N}} p(\y|{\bf f},{\bm \theta})p({\bf f}|{\bm \theta}) {\bf d}{\bf f}.
\end{equation}
The marginal likelihoods  of different methods studied so far can be computed analytically, and are given below
 \begin{eqnarray*}
\mbox{RVM: }\quad p({\bf y}|{\bm \theta}) &=&\mathcal{N}({\bf y}|{\bf 0}, {\bm \Psi} {\bm \Sigma}_\rho{\bm \Psi}^{\top}+\sigma_e^2 {\bf I}_N), \\
 \mbox{Q-GP, GP: }\quad  p(\y|{\bm \theta})&=&\mathcal{N}({\bf y}|{\bf 0},{\bm \Psi}+\sigma_e^2 {\bf I}_N).
\end{eqnarray*}

{\Remark Since Q-GP and GP have the same marginal likelihood, then they have the same estimator $\widehat{\bm \theta}$ of the hyper-parameters (e.g., maximum likelihood, MAP, MMSE etc.).
}
\newline
\newline
Note that, in all cases, we have a multivariate Gaussian density
$$
p(\y)=\mathcal{N}({\bf y}|{\bf 0},{\bf C}_{yy}),
$$
with ${\bf C}_{yy}= {\bm \Psi} {\bm \Sigma}_\rho{\bm \Psi}^{\top} +\sigma_e^2 {\bf I}_N$ in RVM, and ${\bf C}_{yy}={\bm \Psi}+\sigma_e^2 {\bf I}_N$  in  Q-GP and GP.
 Then, we can write the full negative log-marginal likelihood as 
\begin{eqnarray}\label{CostFunLogLike}
-\log p({\bf y}|{\bm \theta})= \frac{1}{2}{\bf y}^{\top}{\bf C}_{yy}^{-1}{\bf y} + \frac{1}{2}\log \left[\det{\bf C}_{yy}\right]   +const.
\end{eqnarray}
 The first term $\frac{1}{2}{\bf y}^{\top}{\bf C}_{yy}^{-1}{\bf y} $ in Eq.~\eqref{CostFunLogLike} can be considered a fitting term,  the second one  $\frac{1}{2}\log \left[\det{\bf C}_{yy}\right]$ plays the role of a regularizer, i.e., a penalty on the model complexity. 
\newline
We can maximize $p({\bf y}|{\bm \theta})$ obtaining a possible choice $\widehat{{\bm \theta}}$ (i.e., a possible estimator). This approach is also called  {\em type-II maximum likelihood procedure} (a.k.a., {\it empirical Bayes}). Note that, in this way, we  avoid the use of a cross-validation (CV) procedure. Alternatively, one can also consider  a prior $p({\bm \theta})$ over ${\bm \theta}$ and study the  posterior $p({\bm \theta}|{\bf y})\propto p({\bf y}|{\bm \theta})p({\bm \theta})$ using for instance Monte Carlo methods \cite{BookLuca,Robert04}. Different possible point estimators $\widehat{{\bm \theta}}$ are possible such as the maximum, the expected value or  the median of the posterior $p({\bm \theta}|{\bf y})$. Additionally, studying the posterior $p({\bm \theta}|{\bf y})$, we can obtain credible intervals of each parameter and approximate its marginal distribution, for instance.  An additional alternative is the use of a full Bayesian approach, described in the next section.

\section{Uncertainty analysis with GPs}\label{UncertGP_SECT}
Although the predictive GP variance has a good intuitive behavior, its analytical form depends {\em explicitly} just on the inputs $\{\x_n\}_{n=1}^N$ (not on the outputs $\{y_n\}_{n=1}^N$), i.e.,
$\sigma_{f|y}^2(\x)=\psi(\x,\x)-{\bm \psi}({\bf x})^{\top}({\bm \Psi} +\sigma_e^2 {\bf I}_N)^{-1}{\bm \psi}({\bf x})$.
\newline
Namely, fixing the hyper-parameters of the employed kernel, we could compute  $\sigma_{f|y}^2(\x)$ only knowing $\{\x_n\}_{n=1}^N$ and without any information of the signal values $\{y_n\}_{n=1}^N$.  In this case, since we can compute $\sigma_{f|y}^2(\x)$ before knowing the signal, it seems that $\sigma_{f|y}^2(\x)$ can not provide relevant information. However, we will learn  the hyper-parameters  given the data ${\bf y}=[y_1,\ldots,y_N]^{\top}$ and the choice of the hyper-parameters affects (a) the value $\psi(\x,\x)$ (if we have a multiplicative parameter $a$ as in Eq.~\eqref{kernelcompleto}), (b) the  vector ${\bm \psi}({\bf x})$ and (c) the matrix ${\bm \Psi}$. Hence, we can assert that the variance $\sigma_{f|y}^2(\x)$ depends on $\{y_n\}_{n=1}^N$  through the hyper-parameters learning. More information can be obtained by performing a full Bayesian study. 
\newline  
\newline 
{\bf Full Bayesian solution.}  A full Bayesian solution can provide more information for a proper uncertainty analysis, as we show below.
Let us assume also a prior $p({\bm \theta})$ over the hyper-parameters ${\bm \theta}$.  A full Bayesian analysis considers the complete joint posterior, which can be expressed as  
\begin{equation}
\label{compPost}
p({\bf f},{\bm \theta} |\y)=\frac{p({\bf f},{\bm \theta},\y)}{p(\y)}=\frac{p({\bf y}|{\bf f},{\bm \theta}) p({\bf f}|{\bm \theta})p({\bm \theta})}{p(\y)}.
\end{equation}
This is a more complete and proper approach from a Bayesian point of view. However, moments and other features of  $p({\bf f},{\bm \theta} |\y)$ are not analytically available, so that the application of computational algorithms (such as Monte Carlo methods) is required  \cite{BookLuca,Robert04}.
 Indeed, so far we have considered
a  {\it conditional posterior}, i.e.,
\begin{align}\label{condPost}
p({\bf f}|\y,{\bm \theta})  &= \frac{p({\bf f},{\bm \theta},\y)}{p({\bm \theta},\y)} =\frac{p(\y| {\bf f}, {\bm \theta})p({\bf f}|{\bm \theta})p({\bm \theta})}{p(\y| {\bm \theta})p({\bm \theta})}= \frac{p({\bf y}|{\bf f},{\bm \theta}) p({\bf f}|{\bm \theta})}{p({\bf y}|{\bm \theta})},
\end{align}
where the conditional marginal likelihood is  $p({\bf y}|{\bm \theta})=\int_{\mathbb{R}^N}   p(\y| {\bf f}, {\bm \theta})p({\bf f}|{\bm \theta}) d\f$. 
One {\it marginal posterior} of ${\bm \theta}$ is given as
\begin{align}\label{margPost}
p({\bm \theta}|{\bf y})&=\frac{p({\bf y}|{\bm \theta})p({\bm \theta})}{p(\y)} \propto p({\bf y}|{\bm \theta})p({\bm \theta}), 
\end{align}
where 
$$
p(\y)=\int_{{\bm \Theta}}   p({\bf y}|{\bm \theta})p({\bm \theta}) d{\bm \theta}.
$$
which can be useful for comparing GP models using different kernels, for instance.
 Note that the relationship among the full posterior in Eq. \eqref{compPost}, the conditional posterior in Eq. \eqref{condPost}, and the marginal posterior in Eq. \eqref{margPost}, is give by
\begin{align}
\label{SuperIMPEq}
p(\f,{\bm \theta}|\y) &=p(\f|\y,{\bm \theta})p({\bm \theta}|\y).
\end{align}
Furthermore, the other  {\it marginal posterior} is 
\begin{align}
\label{SuperIMPEq2}
p(\f|\y) &=\int_{{\bm \Theta}} p(\f|\y,{\bm \theta})p({\bm \theta}|\y) d{\bm \theta}.
\end{align}
Note that the equation above resembles the so-called posterior predictive approach (see \cite[Section 7.4]{LlorenteREV19}).
In order to study the posterior  $p(\f,{\bm \theta}|\y)$, we can use a Monte Carlo approximation.
We can generate $N$ samples from the complete posterior $\{\f_s,{\bm \theta}_s\} \sim p(\f,{\bm \theta}|\y)$, where ${\bm \theta}_s\sim p({\bm \theta}|\y)$ and $\f_s \sim p(\f|\y,{\bm \theta}_s)$, with $s=1,...,S$. The difficult task is to draw from the marginal posterior $p({\bm \theta}|\y)$, whereas the conditional  posterior $p(\f|\y,{\bm \theta}_s)$ is a Gaussian pdf with known mean and covariance matrix (for any possible value of ${\bm \theta}_s$). Thus, the Monte Carlo approximation of  marginal posterior $p(\f|\y)$ in Eq. \eqref{SuperIMPEq2}, is a mixture of Gaussians,
\begin{align}
\label{SuperIMPEq3}
p(\f|\y) &\approx \frac{1}{S}\sum_{s=1}^S p(\f|\y,{\bm \theta}_s), \qquad {\bm \theta}_s\sim p({\bm \theta}|\y).
\end{align} 
{\bf Dependence on the choice of the hyperparameters.} Here, we discuss a more complete study related to the uncertainty analysis of the solutions. 
Let us draw $S$ samples ${\bm \theta}_1,...,{\bm \theta}_S$ from the marginal posterior $p({\bm \theta}|{\bf y})$. Since the conditional posterior mean $\mu_{f|y}(\x|{\bm \theta}_s)=\widehat{f}(\x|{\bm \theta}_s)$ and variance $\sigma_{f|y}^2(\x|{\bm \theta}_s)$ depend on  the hyper-parameters ${\bm \theta}_s$, we also have $S$ mean and variance values (for each $\x$), i.e., $\widehat{f}(\x|{\bm \theta}_1)$,....,$\widehat{f}(\x|{\bm \theta}_S)$
and $\sigma_{f|y}^2(\x|{\bm \theta}_1)$,....,$\sigma_{f|y}^2(\x|{\bm \theta}_S)$.
Then we can calculate the approximate averaged solution as
\begin{eqnarray*}
{\bar f}(\x)&=&\frac{1}{S}\sum_{s=1}^S \widehat{f}(\x|{\bm \theta}_s)\approx \int_{{\bm \Theta}}  \widehat{f}(\x|{\bm \theta}) p({\bm \theta}|{\bf y}) d{\bm \theta} \\
{\bar \sigma}^2(\x)&=&\frac{1}{S}\sum_{s=1}^S \sigma_{f|y}^2(\x|{\bm \theta}_s)\approx E_p\left[\sigma_{f|y}^2(\x|{\bm \theta})\right]= \int_{{\bm \Theta}}  \sigma^2_{f|y}(\x|{\bm \theta}) p({\bm \theta}|{\bf y}) d{\bm \theta}.
\end{eqnarray*}
Note that ${\bar f}(\x)$ is an approximation the expected value associated to marginal posterior $p(f(\x)|\y)$.
Thus, we can also compute
\begin{eqnarray*}
 V_f(\x)&=&\frac{1}{S}\sum_{s=1}^S\left(\widehat{f}(\x|{\bm \theta}_s)-{\bar f}(\x)\right)^2 \approx  \mbox{Var}\left[\widehat{f}(\x|{\bm \theta})\right]. 
\end{eqnarray*}
For the law of total variance, the variance associated to $p(f(\x)|\y)$ is
\begin{eqnarray}
\mbox{Var}[\widehat{f}(\x)]&=& \mbox{Var}\left[E_p[f(\x|{\bm \theta})]\right]+E_p\left[\mbox{Var}[f(\x|{\bm \theta})]\right],  \\
&=& \mbox{Var}\left[\widehat{f}(\x|{\bm \theta})\right]+E_p\left[\sigma_{f|y}^2(\x|{\bm \theta})\right], \\
&\approx& V_f(\x)+{\bar \sigma}^2(\x).
\end{eqnarray}
Therefore, the complete variance is the sum of the two terms ${\bar \sigma}^2(\x)$ and $V_f(\x)$. Moreover, it is interesting to analyze the term $V_f(\x)$ which provides the variation of the mean solution depending on the choice of the hyper-parameters ${\bm \theta}$ (i.e., a sensitivity analysis). 
\newline
 In the rest of the work, we describe methods that are also connected (in some sense) with RVM, GP and Q-GP.

\section{Linear kernel smoothers}\label{linearSmoothertodo}
RVMs and GPs belong to a more general class of regressors: the linear kernel smoothers. In this family of regression methods  the prediction $\widehat{f}({\bf x})$ at some input $\x$ is expressed as linear combination of the outputs $y_1,\ldots,y_N$. The weights of this combination vary with $\x$. We can interpret that is another way of {\it implicitly} modeling the correlation among the different outputs. Generally, the linear smoothers have not associated a probabilistic derivation (unlike RVMs and GPs), so that we focus on the approximation $\widehat{f}({\bf x})$.

\subsection{Definition and examples}

 A linear smoother is a regressor which combines linearly the observations $y_1,\ldots,y_N$ at each $\x$, i.e., 
  \begin{equation}
 \label{LinearSmootherEq_aquiSection8}
\widehat{f}({\bf x})=\sum_{n=1}^N  \varphi_n({\bf x},{\bf x}_n) y_n= {\bm \varphi}({\bf x})^{\top} {\bf y},
\end{equation}
where ${\bm \varphi}({\bf x})=[\varphi_1({\bf x},{\bf x}_1),\ldots,\varphi_N({\bf x},{\bf x}_N)]^{\top}$ and $\varphi_n({\bf x},{\bf x}_n): \mathcal{X}Ê\times \mathcal{X}\rightarrow  \mathbb{R}$  plays the role as a weight. 
 When the output of a regression technique can be expressed as in Eq.~\eqref{LinearSmootherEq_aquiSection8} then it is also called linear kernel smoother.  Equation \eqref{LinearSmootherEq_aquiSection8} shows that the estimator at any input ${\bf x}$, i.e., $\widehat{f}({\bf x})$, can be expressed as linear combination of the $N$ outputs  $y_1,\ldots,y_N$. We can observe that   the coefficients of the linear combination $ \varphi_n({\bf x},{\bf x}_n)$, with $n=1,\ldots,N$, depend on ${\bf x}$. 

\subsubsection{The case of RVM and GP}

\begin{figure}[!h]
	\centering
\centerline{	
	\subfigure[RVM ($\sigma_e=0.5$)]{\includegraphics[width=7cm]{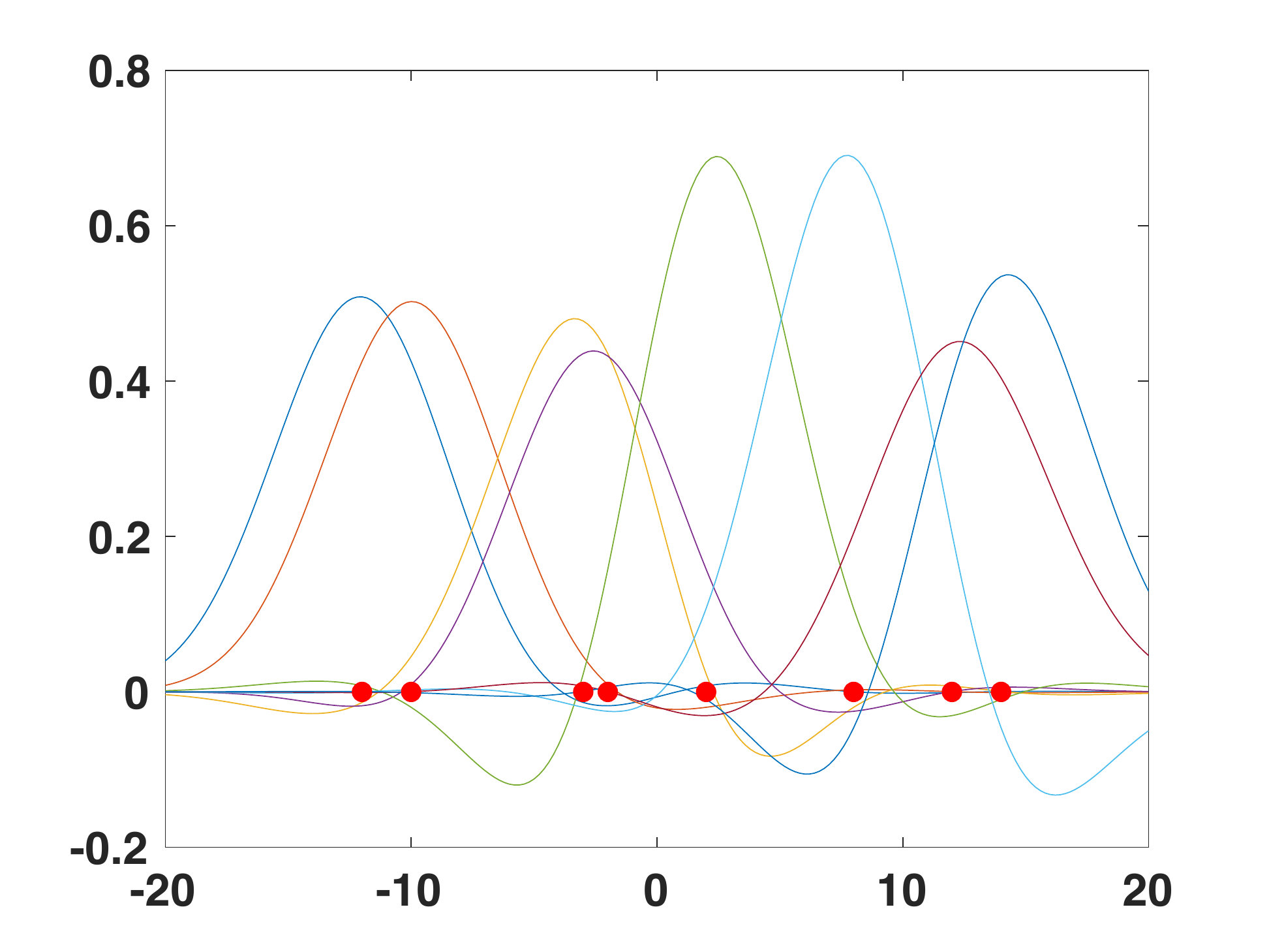}}
	\hspace{-0.8cm}	
	\subfigure[GP ($\sigma_e=0.5$)]{\includegraphics[width=7cm]{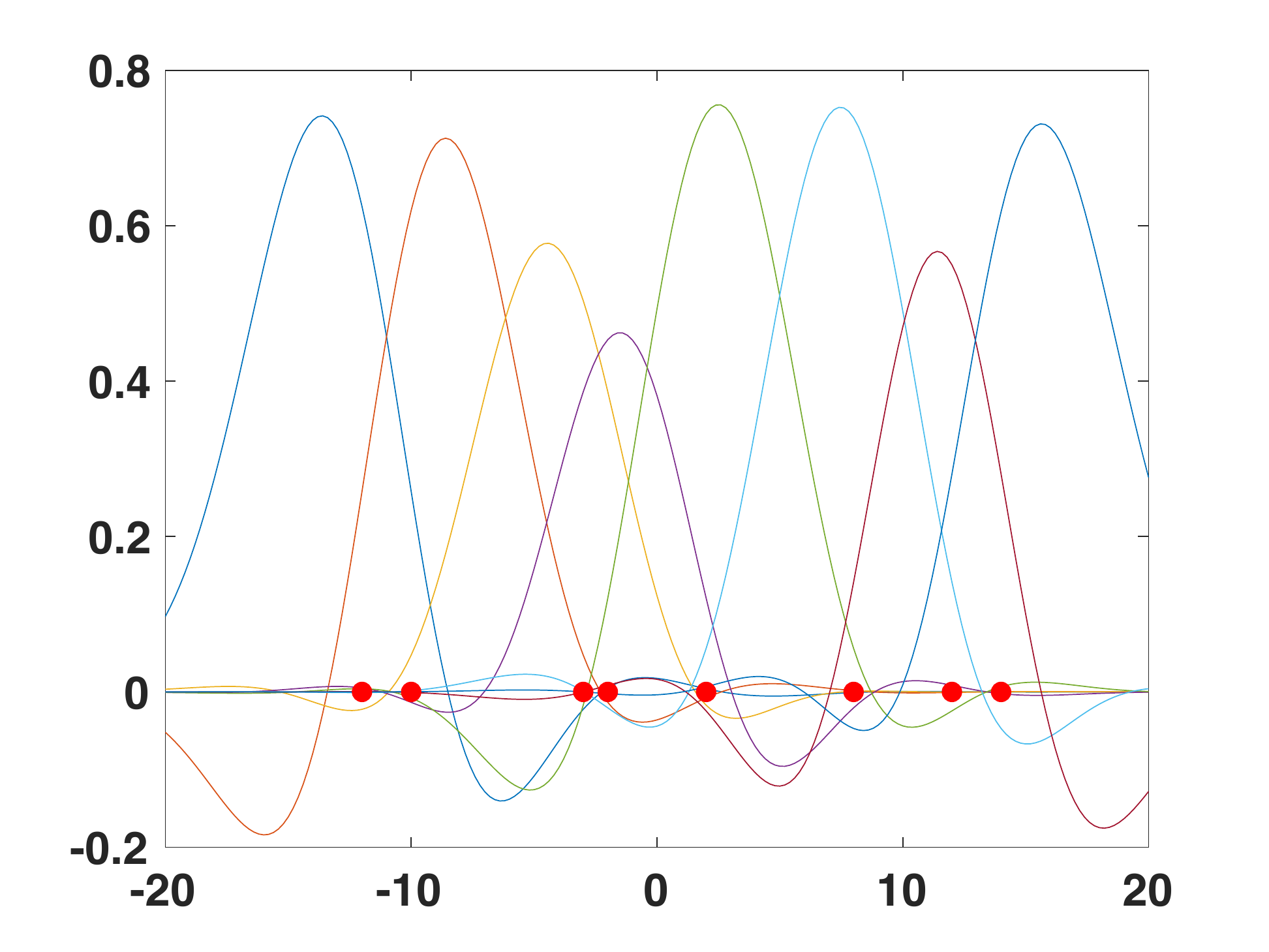}}	
	\hspace{-0.8cm}		
\subfigure[RVM, GP ($\sigma_e=0$)]{\includegraphics[width=7cm]{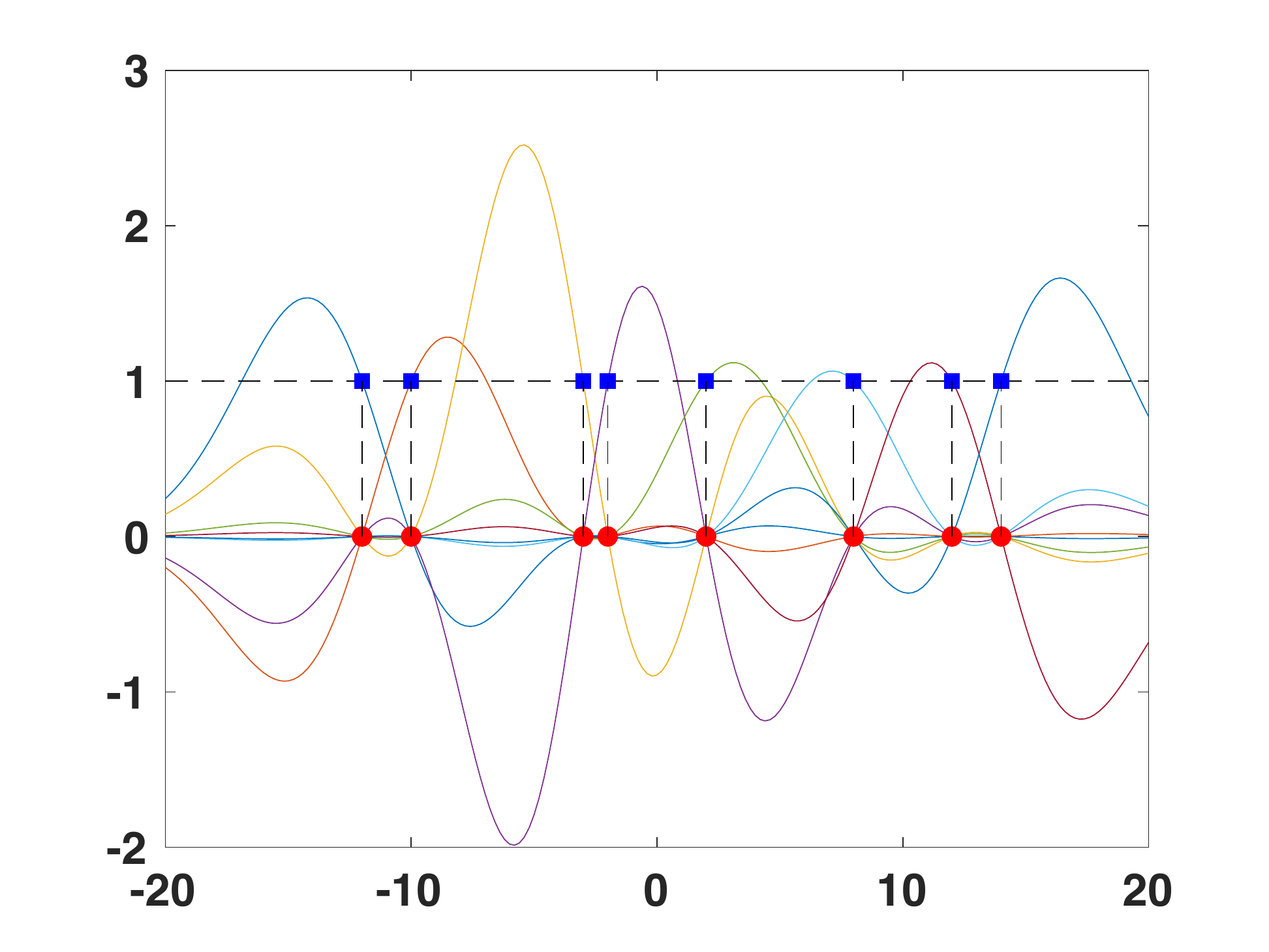}}	
}
	\caption{An example of weighting functions $ \varphi_n(x, x_n)$; {\bf (a)} in a RVM and {\bf (b)} in a GP. We have considered $N=8$ data inputs $x_n$ (shown with dots),  $\psi({\bf x},{\bf x}_n)=\exp\left(-\frac{(x-x_n)^2}{\lambda}\right)$ with $\lambda=25$, and $\sigma_e=0.5$. For RVM, we have also considered a diagonal covariance matrix ${\bm \Sigma}_\rho$ with all the elements in the diagonal equal to $1$. {\bf (c)} Weighting functions $ \varphi_n(x, x_n)$ of RVM and GP for the interpolation case, $\sigma_e=0$. In this scenario, at each data input $x_n$, all  $ \varphi_n(x, x_n)$ are zero except the $n$-th function where $ \varphi_n(x_n, x_n)=1$, i.e., $\varphi_n(x_j, x_n)=\delta_{jn}$.  
 }
	\label{figRVMGPLinearSmoother}
\end{figure}

 The RVM and Q-GP, GP  are linear kernel smoothers since the mean function of the posterior can be expressed as in Eq.~\eqref{LinearSmootherEq_aquiSection8}, setting
  \begin{align}
 \label{RVM_GP_LinearSmoother}
\mbox{RVM:} \quad {\bm \varphi}({\bf x})^{\top}&= {\bm \psi}({\bf x})^{\top}\boldsymbol{\Sigma}_\rho{\bm \Psi}^{\top} \left({\bm \Psi}\boldsymbol{\Sigma}_\rho{\bm \Psi}^\top+\sigma_e^2 {\bf I}_N\right)^{-1}, \\
\mbox{Q-GP,GP:} \quad {\bm \varphi}({\bf x})^{\top}&= {\bm \psi}({\bf x})^{\top} \left({\bm \Psi}+\sigma_e^2 {\bf I}_N\right)^{-1}. 
\end{align} 
Namely,  the weighting functions $ \varphi_n({\bf x},{\bf x}_n)$ depend also on the nonlinearities $\psi_n({\bf x},{\bf x}_n)$ as shown above. Figure \ref{figRVMGPLinearSmoother} shows some examples of the weighting functions $ \varphi_n({\bf x},{\bf x}_n)$ in RVM and GP cases.

\subsubsection{Normalized weighed functions}
\begin{figure}[!h]
	\centering
\centerline{	
	\subfigure[$\lambda=1$]{\includegraphics[width=7cm]{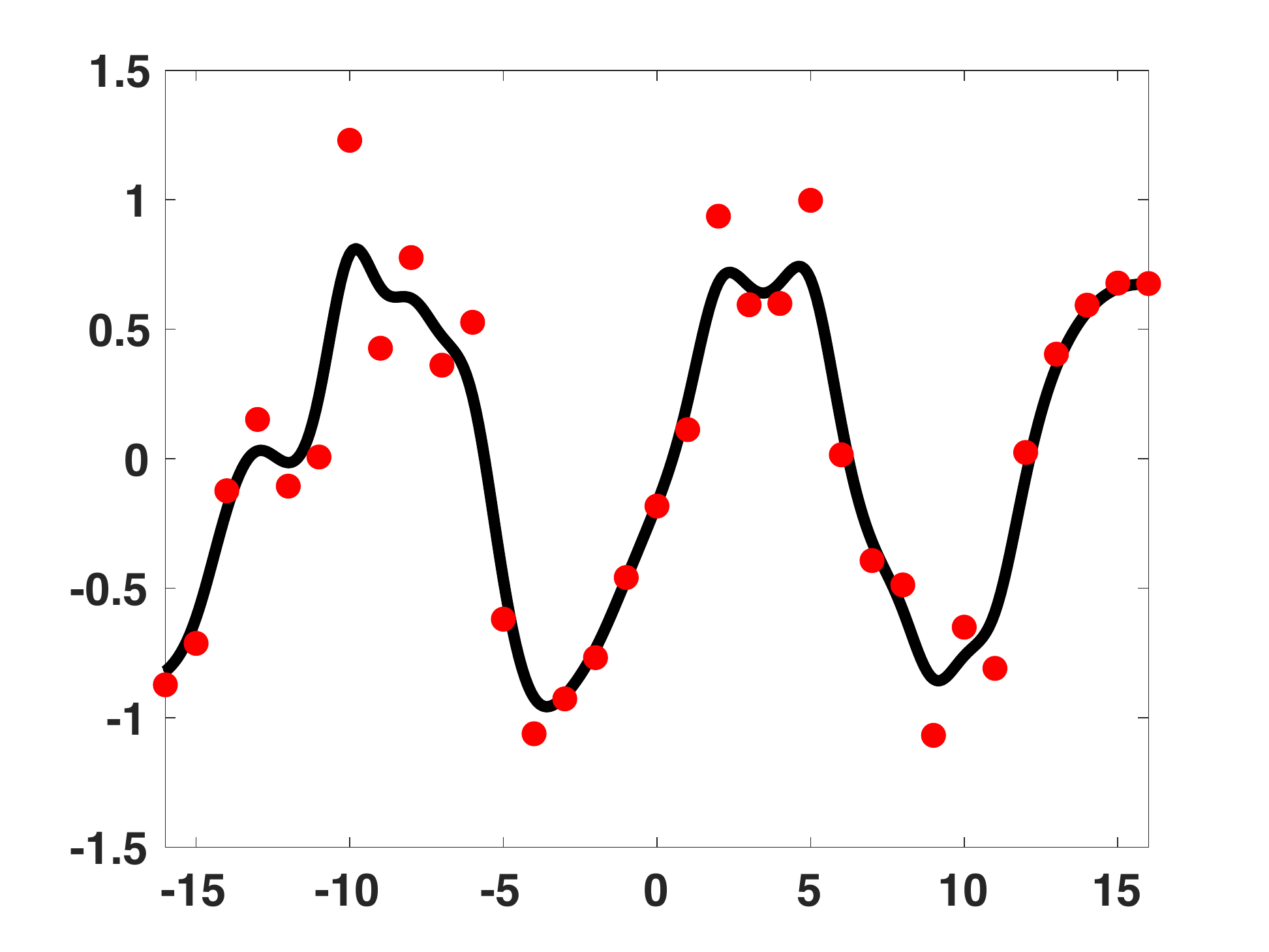}}
	\subfigure[$\lambda=3$]{\includegraphics[width=7cm]{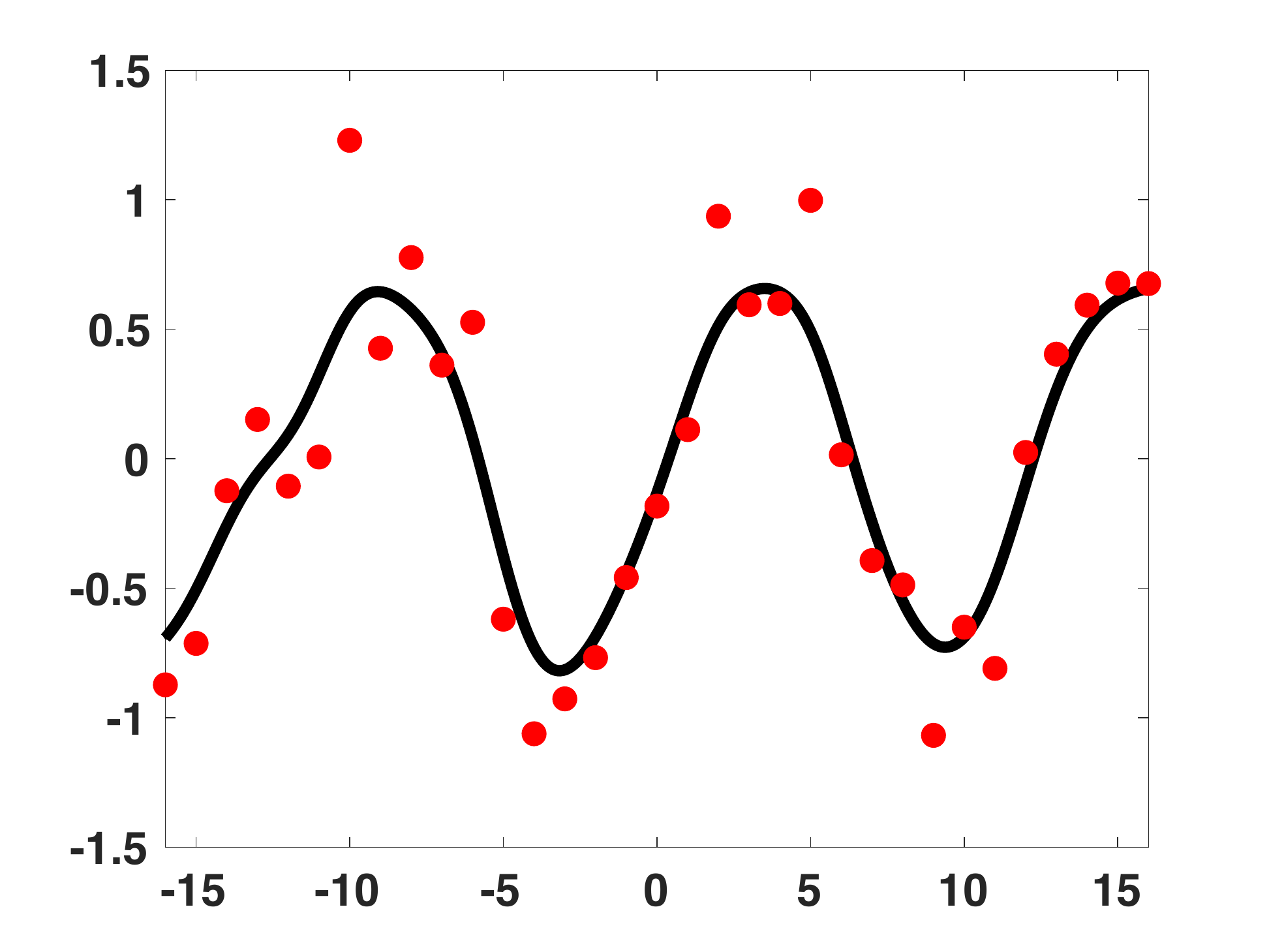}}	
}
\centerline{	
	\subfigure[$\lambda=0.1$]{\includegraphics[width=7cm]{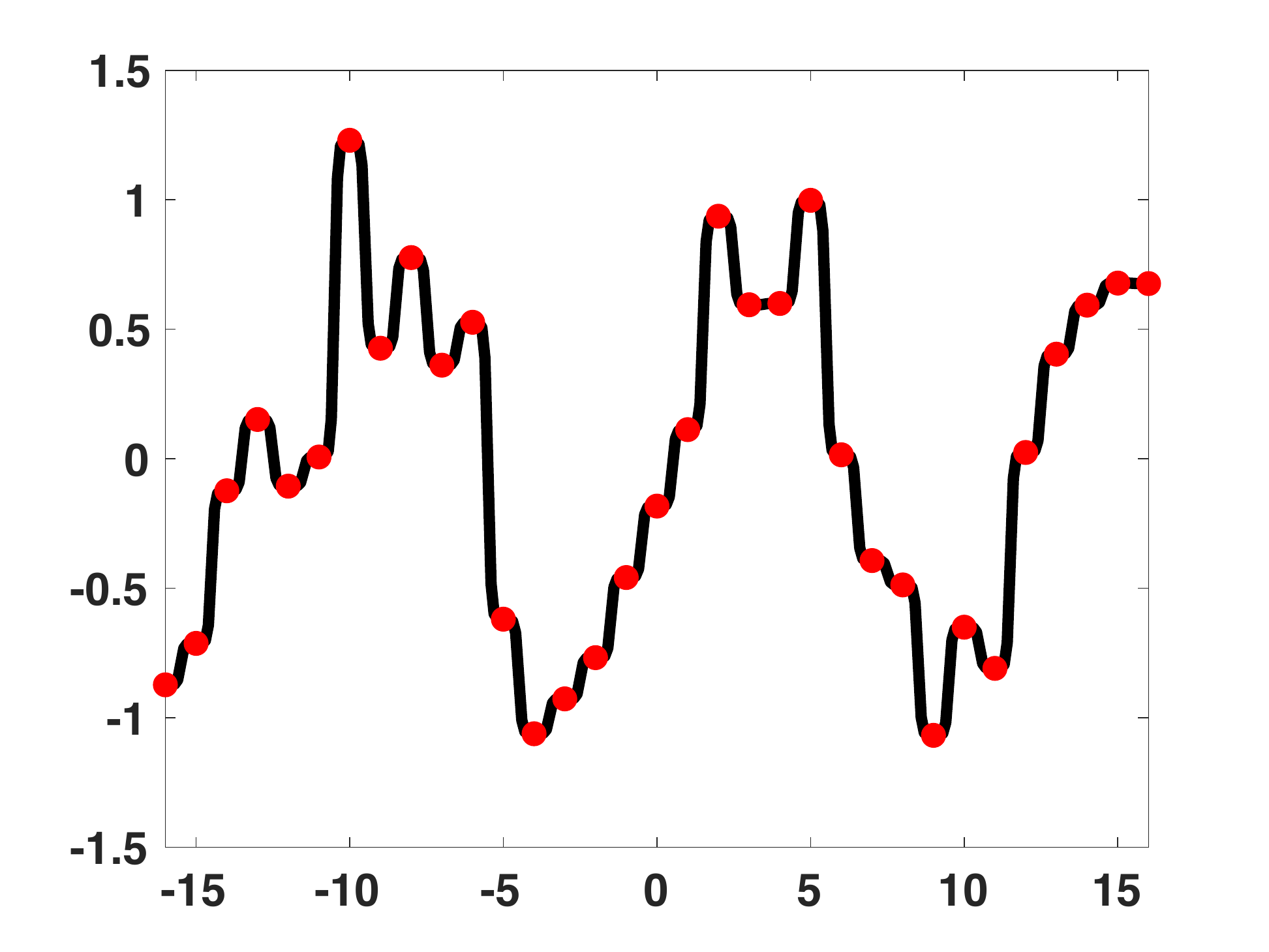}}
	\subfigure[$\lambda=100$]{\includegraphics[width=7cm]{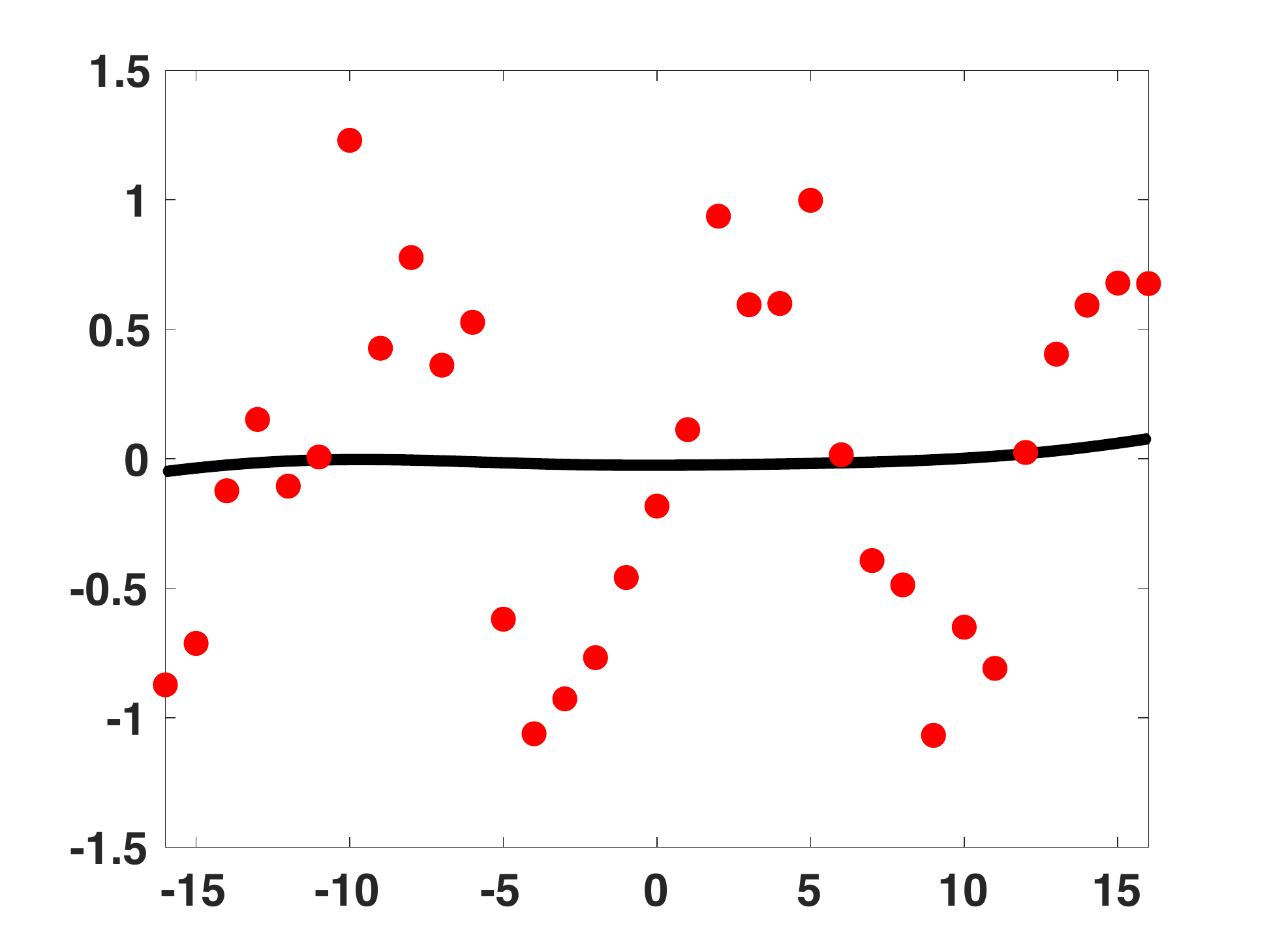}}	
}
	\caption{Examples of  $\widehat{f}(x)$ (with $x\in \mathbb{R}$) when $h_\lambda(x,z)=\exp\left(-(x-z)^2/\lambda\right)$ and $\lambda\in\{0.1,1,3,100\}$. The data points are shown with red dots.
 }
	\label{figConvexSmoother}
\end{figure}

Generally, the linear combination above in Eq.~\eqref{LinearSmootherEq_aquiSection8} is {\it not} a convex combination.
Often, people considers linear smoothers defined as a convex combination where the weight function are positive and the sum is 1. For instance, consider when auxiliary weighting function  $h_\lambda({\bf x},{\bf x}_n)\geq 0$ is used to assign weights to $x_n$
based on its distance from $\x$. The parameter $\lambda \in \mathbb{R}$ indicates the bandwidth (the width of the neighborhood), determined
from the training data. One example is {\it the  Nadaraya-Watson estimator} where
  \begin{equation}
 \label{NWestimator}
\widehat{f}({\bf x})=\sum_{n=1}^N  \frac{h_\lambda({\bf x},{\bf x}_n)}{\sum_{j=1}^N h_\lambda({\bf x},{\bf x}_j)} y_n=\sum_{n=1}^N \varphi_n({\bf x},{\bf x}_n)  y_n,
\end{equation}
where $\varphi_n({\bf x},{\bf x}_n)=\frac{h_\lambda({\bf x},{\bf x}_n)}{\sum_{j=1}^N h_\lambda({\bf x},{\bf x}_j)}$. Note that, with this definition, 
$$
\sum_{n=1}^N \varphi_n({\bf x},{\bf x}_n)=1.
$$
Figure \ref{figConvexSmoother} provides some examples of  $\widehat{f}(x)$ (with $x\in \mathbb{R}$) when $h_\lambda(x,z)=\exp\left(-(x-z)^2/\lambda\right)$ and different values of $\lambda$.
 The form of this estimator above is quite general. For instance, it contains the k-nearest neighbors algorithm (kNN) for regression as a specific case (with a specific choice of $h_\lambda({\bf x},{\bf x}_n)$).  See the next sections for further details.
 
\subsubsection{Derivation of Nadaraya-Watson estimator}
So far we have considered the outputs $y_n$ as random variables (affected by random perturbations), whereas the inputs $\x_n$ are considered as auxiliary deterministic information.
Now, let us consider both $\x_n$ and $y_n$ as random variables with joint density $p(\x,y)$. Namely, we assume
$$
[\x_n,y_n]\sim p(\x,y) \qquad n=1,\ldots,N. 
$$
We can try to estimate $p(\x,y)$ via kernel density estimation,
\begin{eqnarray}
\widehat{p}(\x,y)= \frac{1}{N} \sum_{n=1}^N h_{\lambda_x}(\x-\x_n) h_{\lambda_y}(y-y_n).
\end{eqnarray}
where $\int_{\mathcal{X}}h_{\lambda_x}(\x) d\x=1$ and $\int_{\mathbb{R}}h_{\lambda_y}(y) dy=1$. Moreover,  $\int_{\mathcal{X}} \x h_{\lambda_x}(\x) d\x={\bf 0}$ and $\int_{\mathbb{R}}y h_{\lambda_y}(y) dy=0$. The regression function is defined as 
\begin{equation}\label{NWequationTEO}
\widehat{f}(\x)=E[y|\x]=\int_{\mathbb{R}} y p(y|\x) dy= \frac{\int_{\mathbb{R}} y p(\x,y) dy}{\int_{\mathbb{R}}  p(\x,y) dy}.
\end{equation}
Replacing $p(\x,y)$ with $\widehat{p}(\x,y)$, then we have  
\begin{align*}
\int_{\mathbb{R}} y \widehat{p}(\x,y) dy&=\frac{1}{N}\int_{\mathbb{R}} y \sum_{n=1}^N h_{\lambda_x}(\x-\x_n) h_{\lambda_y}(y-y_n) dy, \\
&= \frac{1}{N} \sum_{n=1}^N h_{\lambda_x}(\x-\x_n) y_n,
\end{align*}
where we have used $\int_{\mathbb{R}}y h_{\lambda_y}(y-y_n) dy=y_n$, since $\int_{\mathbb{R}}y h_{\lambda_y}(y) dy=0$, i.e., the mean is zero by assumption and, hence the mean of  $h_{\lambda_y}(y-y_n)$ is $y_n$ (it is just a translation of the pdf). Moreover,
\begin{align*}
\int_{\mathbb{R}}  \widehat{p}(\x,y) dy&=\int_{\mathbb{R}}  \sum_{n=1}^N h_{\lambda_x}(\x-\x_n) h_{\lambda_y}(y-y_n) dy, \\
&= \frac{1}{N} \sum_{n=1}^N h_{\lambda_x}(\x-\x_n).
\end{align*}
Thus, replacing the numerator and denominator of Eq.~\eqref{NWequationTEO} with the two approximations above, finally we can write 
\begin{equation}
\widehat{f}(\x)\approx \sum_{n=1}^N  \frac{h_{\lambda_x}({\bf x},{\bf x}_n)}{\sum_{j=1}^N h_{\lambda_x}({\bf x},{\bf x}_j)} y_n.
\end{equation}
This is the  Nadaraya-Watson estimator with $\varphi_n({\bf x},{\bf x}_n)=\frac{ h_{\lambda_x}({\bf x},{\bf x}_n)}{\sum_{j=1}^N  h_{\lambda_x}({\bf x},{\bf x}_j)}$ \cite{bishop2006pattern}.

\subsection{Other examples of linear smoothers}

In section, we describe some well-known linear smoothers that are encompassed in Eq.~\eqref{LinearSmootherEq_aquiSection8}.
 
 \subsubsection{k-Nearest Neighbors (kNN)} 
 In this section, we replace the real parameter $\lambda$ with an integer value $k\in \mathbb{N}^+$.
 In the kNN technique for regression, given an integer value $1\leq k \leq N$, we have 
 $$
 h_k({\bf x},{\bf x}_n)= 1,
 $$
 if $\x_n$ is one of the $k$ nearest inputs of $\x$ (within the $N$ possible inputs $\x_n$), otherwise
 $$
 h_k({\bf x},{\bf x}_n)= 0,
 $$ 
 if $\x_n$ does not belong to the set $k$ of nearest inputs of $\x$. Let us consider now the two extreme cases. If $k=1$, only one function $h_k({\bf x},{\bf x}_{j^*})$ will be equal to $1$ (where ${\bf x}_{j^*}$ represents to the closest input to $\x$). As a consequence, for all $\x$ such that ${\bf x}_{j^*}$ is the closest input then $\widehat{f}({\bf x})=y_{j^*}$, i.e., we obtain an interpolator. If $k=N$, all functions $h_k(\x,\x_{j^*})=1$ and, as  a consequence,
 $$
 \widehat{f}({\bf x})=\frac{1}{N}\sum_{n=1}^N y_n, \quad \quad \forall \x\in \mathcal{X},
 $$
 i.e., we obtain a constant approximation, equal to the arithmetic mean of the outputs $y_n$ \cite{murphy12}.

 \subsubsection{Inverse distance weighting}
 Another example is the so-called {\it inverse distance weighting} method for  multivariate interpolation with the choice
 \begin{align*}
h_\lambda({\bf x},{\bf x}_n)&=\frac{1}{d_\lambda({\bf x},{\bf x}_n)^p}, \quad \mbox{ if } \quad  {\bf x}\neq {\bf x}_n, 
\end{align*}
where $d_\lambda({\bf x},{\bf x}_n)$  is a distance (metric operator) and $p>0$ is a positive real value  \cite{Shepard68}. When ${\bf x}= {\bf x}_n$, we directly set $\widehat{f}({\bf x}_n)=y_n$ (i.e., we have an interpolator). Note that the weight $h_\lambda({\bf x},{\bf x}_n)$ decreases as $d_\lambda({\bf x},{\bf x}_n)$ grows. 

 \subsubsection{Polynomial interpolation with Lagrange bases}
\begin{figure}[!h]
	\centering
\centerline{	
	\subfigure[]{\includegraphics[width=8cm]{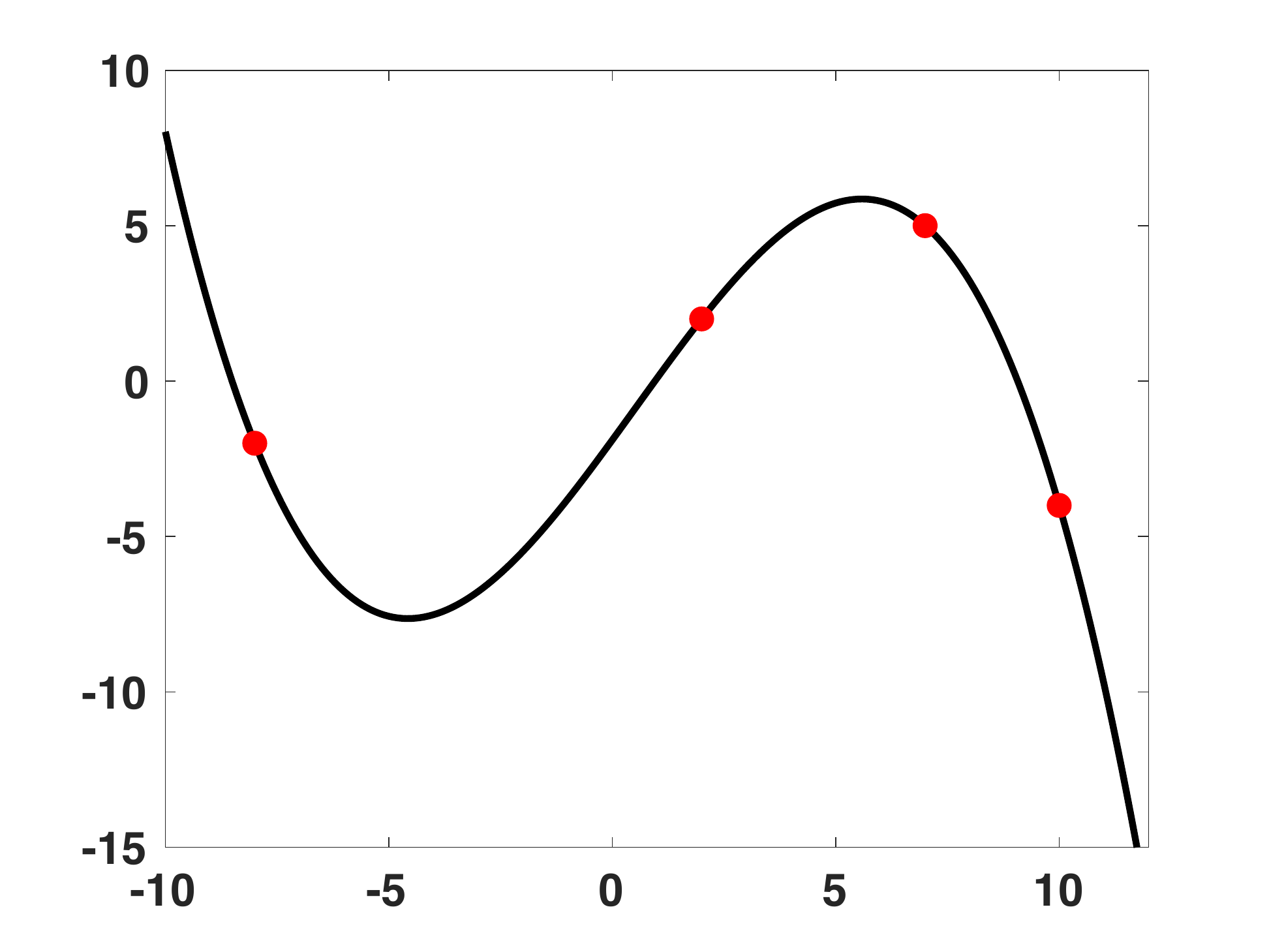}}
	\subfigure[]{\includegraphics[width=8cm]{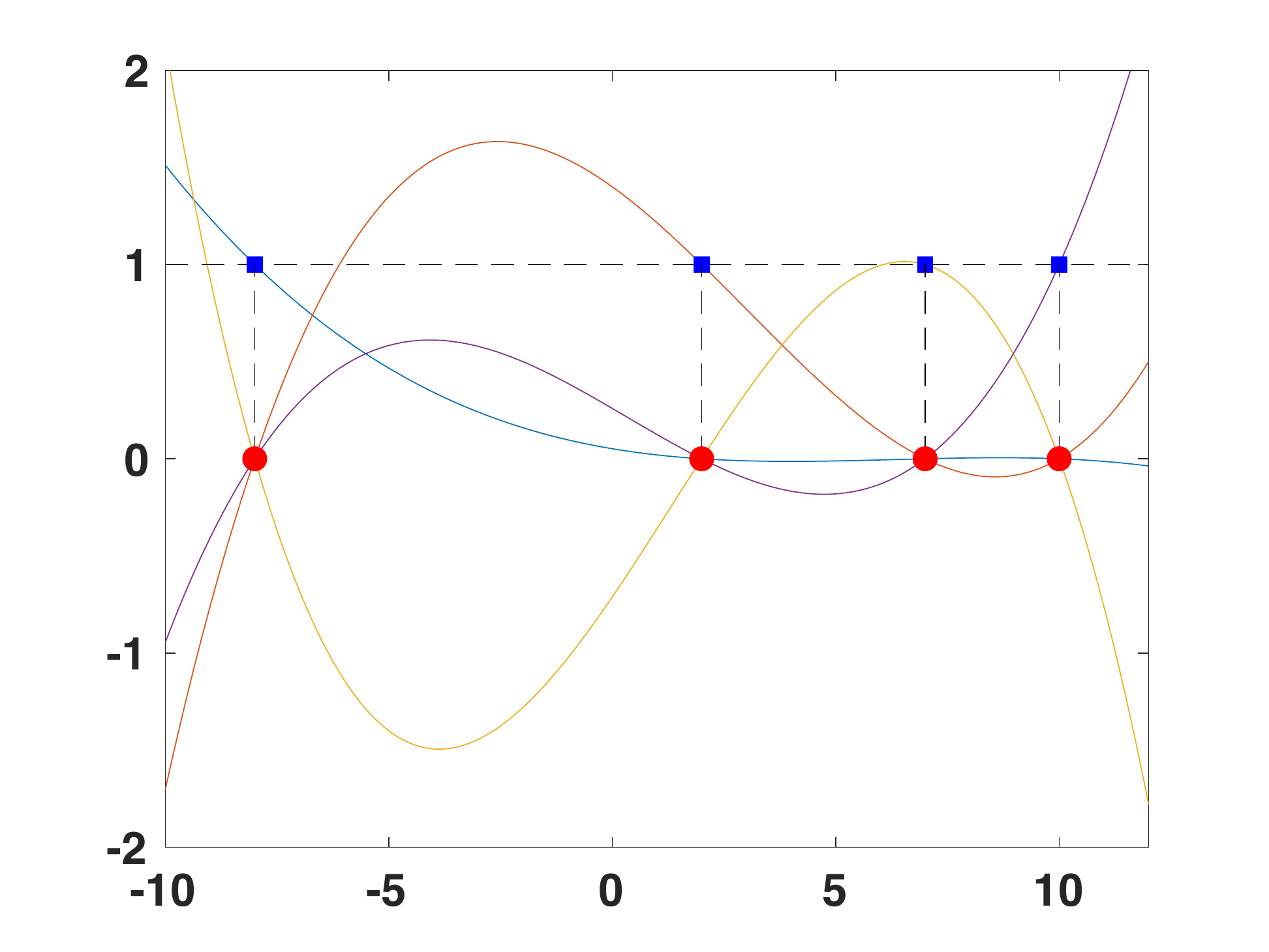}}
}
	\caption{{\bf (a)} Example of polynomial interpolation of $N=4$ data points. {\bf (b)} The corresponding Lagrange functions $\varphi_n(x,x_n)=L_n(x)$.  Note that, at each data input $x_n$, all  $ \varphi_n(x, x_n)$ are zero except the $n$-th function where $ \varphi_n(x_n, x_n)=1$, i.e., $\varphi_n(x_j, x_n)=\delta_{jn}$, as in Figure \ref{figRVMGPLinearSmoother}(c).  
 }
	\label{figLagrangeINT}
\end{figure}

In this section, we focus on the interpolation problem as typycally addressed in the initial courses of numerical analysis \cite{Burden00}.
Let us consider for simplicity the scalar input case, $x_i\in \mathbb{R}$. Consider the problem of obtaining the polynomial interpolation of  order $N-1$ of $N$ data $\{x_i,y_i\}_{i=1}^N$. Let us also define the Lagrange polynomial weighting functions \cite{Burden00,Plybon92},  
\begin{align}
	\varphi_n(x,x_n)&=L_{n}(x) \notag \\ &=\frac{(x-x_1)\cdots(x-x_{k-1})(x-x_{k+1})\cdots (x-x_{N})}{(x_n-x_1)\cdots(x_n-x_{k-1})(x_n-x_{n+1})\cdots (x_n-x_{N})}, \\
&=\prod_{i=1,i\neq n}^{N}\frac{x-x_i}{x_n-x_i}, \qquad n=1,\ldots,N.
\end{align}
Note that $L_n(x_n)=1$ and $L_n(x_i)=0$ for all $i\neq n$, i.e., we have $L_n(x_i)=\delta_{in}$ (as in Figure \ref{figRVMGPLinearSmoother}(c) for RVM and GP), which is exactly the condition for obtaining an interpolator, i.e., $\widehat{f}(x_n)=y_n$. Note that  Lagrange functions $L_n(x)$ are also polynomials of order $N-1$. 
 The polynomial interpolator $\widehat{f}(x)$ can be written as
\begin{align}
\widehat{f}(x)=\sum_{n=1}^NL_{n}(x) y_n,
\end{align}
i.e., a linear combination of the outputs $y_n$ with $\varphi_n(x,x_n)=L_{n}(x)$ (or, equivalently, a linear combination of the Lagrange functions $L_{n}(x)$). Figure \ref{figLagrangeINT} gives an example of polynomial interpolation of $N=4$ data points and the corresponding Lagrange functions $L_n(x)$.
 
\subsubsection{Ideal Fourier interpolation} \label{FourierRecFilter}

\begin{figure}[!h]
	\centering
\centerline{	
	\subfigure[]{\includegraphics[width=8cm]{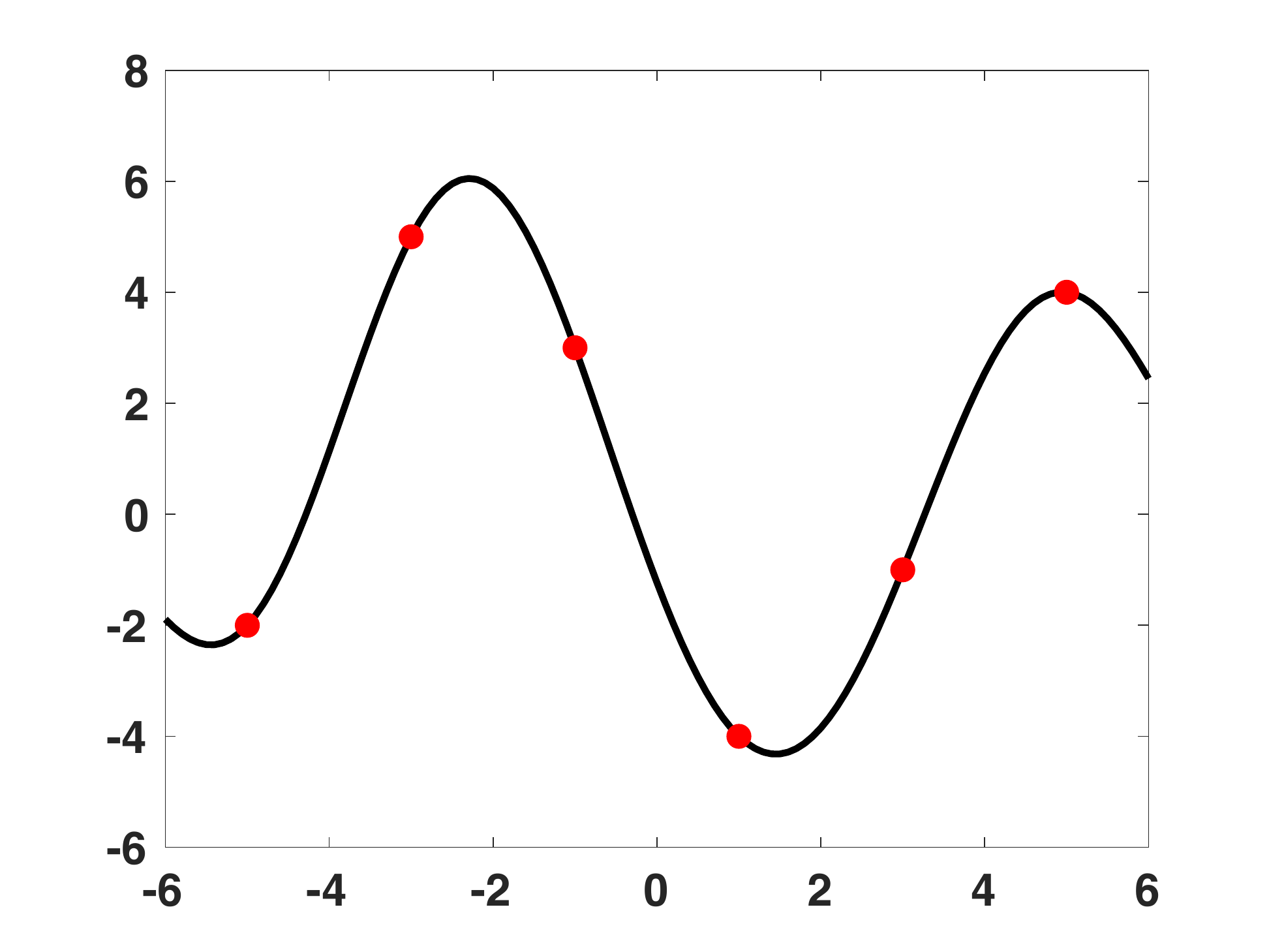}}
	\subfigure[]{\includegraphics[width=8cm]{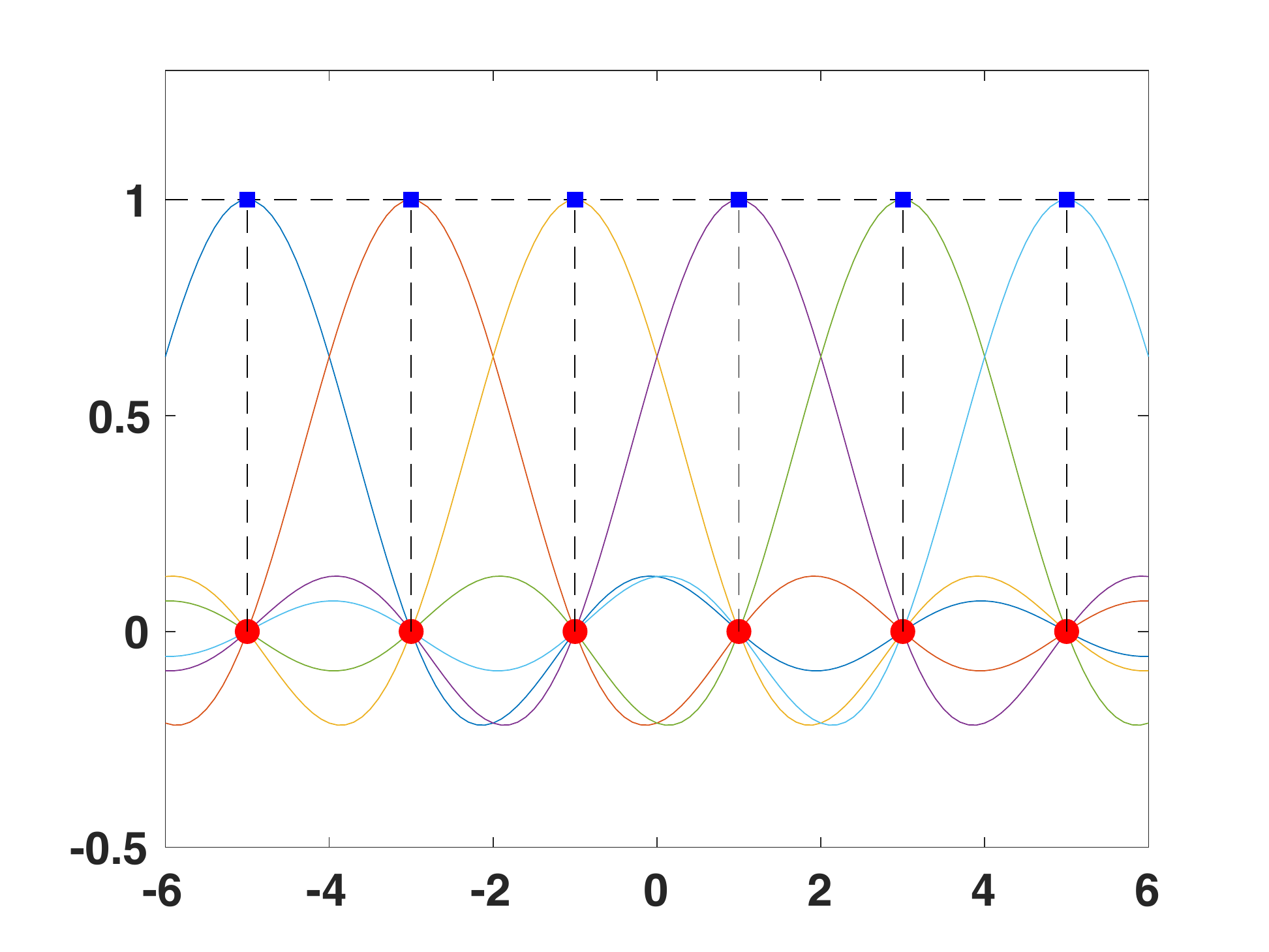}}
}
	\caption{{\bf (a)} Example of the ideal Fourier interpolation of $N=6$ data points. {\bf (b)} The corresponding weighting sinc functions $\varphi_n(x,x_n)=\mbox{sinc}(x,x_n)$.  Note that, at each data input $x_n$, all  $ \varphi_n(x, x_n)$ are zero except the $n$-th function where $ \varphi_n(x_n, x_n)=1$, i.e., $\varphi_n(x_j, x_n)=\delta_{jn}$, as in Figure \ref{figRVMGPLinearSmoother}(c) and Figure \ref{figLagrangeINT}(b).  
 }
	\label{figFourierINT}
\end{figure}

The interpolation idea is employed as upsampling in signal processing in a context of  equidistant inputs \cite{Kamen04,Proakis00}. 
For simplicity, let us consider a scalar input $x\in \mathbb{R}$. Moreover, consider an infinite number of equidistant inputs, i.e., 
$$
x_n=nT_0, \quad T_0\in \mathbb{R}.
$$  
This is an important difference and limitation compared  with the previous cases: here we consider equidistant inputs with a sampling period $T_0$. 
In this scenario, a well-known interpolator $\widehat{f}(x)=\sum_{n=1}^N  \varphi_n( x,x_n) y_n$  is the ideal Fourier interpolator, where 
\begin{eqnarray}
 \varphi_n( x,x_n) = \mbox{sinc}(x,x_n)=T_0\frac{\sin\left(\frac{\pi}{T_0} (x-nT_0)\right)}{\pi(x-nT_0)} .
\end{eqnarray}
Note that $\varphi_i( x_n, x_n)=1$, indeed
$$
\mbox{sinc}(x_n,x_n)=T_0\frac{\sin\left(\frac{\pi}{T_0} (nT_0-nT_0)\right)}{\pi (nT_0-nT_0)}=1 \quad \mbox{(solving the indeterminate form)}
$$
 and $\varphi_n(x_i,x_n)=0$, if $i\neq n$, indeed
$$
\mbox{sinc}(x_i,x_n)=\frac{\sin\left(\pi (i-n)\right)}{\pi (i-n)}=\frac{\sin\left(k\pi\right)}{k\pi }=0, \qquad k=i-n\in \mathbb{Z} \backslash\{0\},
$$ 
since $\sin(k\pi)=0$ with $k\in \mathbb{Z}$. Then,  we have again $\varphi_n( x_i,x_n) = \delta_{in}$ which is the condition to obtain  $\widehat{f}(x_n)=y_n$ for all $n$. The choice of $ \varphi_n( x,x_n) = \mbox{sinc}(x,x_n)$ is due to the Fourier transform of a sin function is an ideal rectangle filter  (for more details see \cite{Kamen04}). An example is shown in Figure \ref{figFourierINT}.

 \subsubsection{Finite and infinite impulse response filters (FIR, IIR)}\label{DigitalFilterSect}

Here, we describe two classes of discrete filters connected to the previous techniques.
Consider again a {\it scalar and discrete} input $x=t\inÊ\mathbb{Z}$,
representing a discrete time index, i.e., $t=\ldots,-2,-1,0,1,2,3,\ldots$.
Moreover,  we consider {\it consecutive} time instants, $t=1,2,\ldots,N$, we can use the simpler notation
$$
\{t_n,y_n\}_{n=1}^N=\{t,y_t\}_{t=1}^N,
$$
removing the sub-index $n$.  In this context, the Finite Impulse Response (FIR) filters are defined  
\begin{align}
\widehat{f}(t)=\widehat{f}_t&=a_0 y_t+a_1 y_{t-1}+a_1 y_{t-2}\ldots+a_R y_{t-R}, \nonumber \\
&=\sum_{r=0}^R a_r y_{t-r},\label{FIReq}
\end{align}
where $a_r$ and $R$ are constant values decided by the user \cite{Kamen04,Proakis00}. The value $R$ is the {\it order} of the filter. Comparing  Eq.~\eqref{LinearSmootherEq_aquiSection8} and the expression above, we can see that a FIR filter is also a linear smoother with coefficients $a_r$ for $r=0,\ldots,R$, and the linear combination only considers the previous $R$ samples $y_{t-1},\ldots,y_{t-R}$ and the current sample $y_t$. If $a_r=\frac{1}{R}$ for all $r$, we have a low-pass filter which compute the arithmetic mean of $R+1$ outputs $y_t, y_{t-1},\ldots,y_{t-R}$. Therefore, we have a ``sliding'' memory window of length $R$.  The FIR filters are similar to the kNN approach but considering $k=R$ nearest neighbors only in {\it the past}, i.e., $t'\leq t$  (not in the future, i.e., $t'>t$).  In a FIR filter,  the output sequence is a weighted sum of the most recent values. In terms of neural network architectures, this can be seen as a linear type of time delay neural network with fixed weights (a standard multi-layer perceptron taking a time window as input)  \cite{DuSwamy}. 

A memory window of infinite length can be obtained considering a FIR filter of order $R=\infty$, i.e., with an infinite amount of coefficients. Thus, the sum in Eq.~\eqref{FIReq} would be a series. In that case, the FIR filter with $R=\infty$ is converted into an Infinite Impulse Response (IIR) filter \cite{Kamen04,Proakis00}. An equivalent way of expressing a IIR filter is incorporating an autoregressive part in Eq.~\eqref{FIReq}, i.e., 
\begin{align}
\widehat{f}_t&=\sum_{\ell=1}^L  b_\ell \widehat{f}_{t-\ell} + \sum_{r=0}^R a_r y_{t-r},\label{IIReq}
\end{align}
where  $b_\ell$ and $L$ are anther constant values, decided by the user. In the context of stochastic filtering, these filters are also called Autoregressive Moving Average (ARMA) models. However, in the ARMA filters, the inputs of the system are unknown noise realizations.
 In terms of neural network architectures, an IIR filter can be interpreted as a linear version of recurrent neural network.







\section{GPs  and Kalman filtering}\label{GPKalmanSect}
In this section, we describe the connection, differences and similarities between the GP approach and the Kalman filtering (KF) approach, from the simplest case to the most general case, progressively \cite{Hartikainen10,pmlr-v22-sarkka12,Sarkka_2013}. 
The Kalman filter is a recursive  MMSE estimator of the state of in a linear state-space model with Gaussian noise perturbations (see Eq. \eqref{aquiModelStateSpace}, below). 
Note that  the MMSE and MAP estimators coincide in this context, i.e., under the assumptions of linearity and Gaussian noises. For more clarifications, see below. 

\subsection{Kalman filter in discrete time}

 For the sake of simplicity, we change the notation considering {\it scalar and discrete} inputs
\begin{equation}
x=t\inÊ\mathbb{Z},
\end{equation}
representing a discrete time index. Later on, we will consider also a continuous time index. The dataset is then $\{t_n,y_n\}_{n=1}^N$. Moreover, if we consider {\it consecutive} time instants, $t=1,2,\ldots,N$, we can use the notation
$$
\{t_n,y_n\}_{n=1}^N=\{t,y_t\}_{t=1}^N,
$$
removing the sub-index $n$. The observation vector  and  the corresponding values of the hidden function  is 
$$
{\bf y}=[y_1,\ldots,y_N]^{\top}, \mbox{ and } {\bf f}=[f_1,\ldots,f_N]^{\top},
$$
and the observation model is 
\begin{equation}
y_t=f_t+e_t, \quad \mbox{ and } \quad {\bf y}={\bf f}+{\bf e},
\end{equation}
where  ${\bf e}=[e_1,\ldots,e_N]^{\top} \sim \mathcal{N}({\bf e}|{\bf 0},\sigma_e^2 {\bf I}_N)$ with ${\bf I}_N$ is an $N\times N$ unit matrix. The likelihood function is again $p({\bf y}|{\bf f})=\mathcal{N}({\bf y}|{\bf f},\sigma_e^2 {\bf I}_N)$. Therefore, the observation model (likelihood) is exactly the same. 

 {\Remark The index $t$ is playing the role of the input, the variable $f_t=f(t)$ is the hidden function at the instant $t$ and $y_t$ is the corresponding observation at the input $t$. Note that we are also considering a normalized uniform sampling case, i.e., $t_i-t_j=1$ for all $i,j$.   }
\newline
\newline
However, instead of assuming directly a covariance function $k(t,t')$  as prior information, we  consider an autoregressive (AR) model over $f_t$, i.e.,
\begin{equation}\label{PropEq}
f_t=\gamma f_{t-1}+v_t, \quad \mbox{ with }\quad |\gamma|<1,
\end{equation}
where $v_t \sim \mathcal{N}(v|0,\sigma_v^2)$, inducing a transition probability $p(f_t|f_{t-1})$. The complete {\it state-space model} is formed by the transition (prior) and observation (likelihood) equations (densities), i.e.,
\begin{gather} \label{aquiModelStateSpace}
\left\{
    \begin{split}
        f_t&=\gamma  f_{t-1}+v_{t}, \\
        y_t& = f_t + e_t,
    \end{split}
    \right.   \Longrightarrow 
    \left\{
    \begin{split}
        &p(f_t|f_{t-1}), \\
      & p(y_t|f_t),
    \end{split}
    \right.
    \qquad t=1,2\ldots,N.
\end{gather}
 Assuming also $p(f_1|f_0)=p(f_1)\sim \mathcal{N}(f_1|0,\sigma_v^2)$, the complete prior density and likelihood function are
 \begin{align}
 p({\bf f}) &=\prod_{t=1}^N p(f_t|f_{t-1})=\mathcal{N}({\bf f}|{\bf 0},{\bm \Psi}), \\
 p({\bf y}|{\bf f})&=\prod_{t=1}^N p(y_t|f_t)=\mathcal{N}({\bf y}|{\bf f},\sigma_e^2 {\bf I}_N).
  \end{align}
 where the $N\times N$ covariance matrix ${\bm \Psi}$ is generated by the autoregressive process in Eq.~\eqref{PropEq} is given below. 
  
 {\Remark The recursion $f_t=\gamma  f_{t-1}+v_{t}$  (equivalent to transition probability $p(f_t|f_{t-1})$) induces a prior over $\f$. Namely,  the recursive equation plays the same role of a  kernel function $k(t,t')$ in the GP derivation. Indeed, we see below that we can obtain an equivalent a  kernel/covariance function $k(t,t')$. }

\subsubsection{Equivalent kernel function of an AR model}
For simplicity, let start the recursion in Eq.~\eqref{PropEq} with $f_0=0$. Then, we can write 
\begin{eqnarray}
E[f_t]=\gamma E[f_{t-1}]+E[v_t]=0, \qquad \forall t.
\end{eqnarray}
Hence, $E[f_1,\ldots,f_N]=0$. Moreover, regarding the variance,  we have 
\begin{eqnarray}
&&\mbox{var}[f_t]=\gamma^2\mbox{var}[f_{t-1}]+\sigma_v^2  \nonumber \\
&&\mbox{var}[f_t]=\gamma^4\mbox{var}[f_{t-2}]+(\gamma^2+1)\sigma_v^2  \nonumber \\
&& \qquad\quad\quad \vdots \nonumber \\
&&\mbox{var}[f_t]=\gamma^{2t}\mbox{var}[f_{0}]+\left(\sum_{i=0}^{t-1} \gamma^{2i}\right)\sigma_v^2.
\end{eqnarray}
Since we start with $f_0=0$ then $\mbox{var}[f_{0}]=0$. However, in any case, the variance of  initial condition $\mbox{var}[f_{0}]$ goes to zero as $t$ approaches infinity since $|\gamma|<1$. Therefore, we can write
\begin{eqnarray}
\mbox{var}[f_t]=\frac{1-\gamma^{2t}}{1-\gamma^2}\sigma_v^2,
\end{eqnarray}
and as $t \rightarrow \infty$, the stationary variance is
\begin{eqnarray}
\sigma_f^2=\mbox{var}[f_\infty]=\frac{\sigma_v^2}{1-\gamma^2}.
\end{eqnarray}
Recall that $|\gamma|<1$, so that the expression above is finite and positive.
The diagonal of ${\bm \Psi}_t$ is given by the values $\mbox{var}[f_t]$.
Similarly, the stationary auto-covariance function 
\begin{align*}
 r(\tau)=\psi(f_t,f_{t+\tau})=\psi(t,t+\tau)&=  \\
&= E[(f_t-\mu_t)(f_{t+\tau}-\mu_{t-\tau})], \\
&=E[f_t f_{t+\tau}],
\end{align*}
depends only to the instants $t$ and $t-\tau$ (actually, only on the different $\tau$). It is possible to show that 
\begin{equation}
r(\tau)=\gamma \cdot r(\tau-1) \quad \mbox{ where }  \quad r(0)=\sigma_f^2.
\end{equation}
Then, we have $r(1)=\gamma \sigma_f^2$,  $r(2)=\gamma^2 \sigma_f^2$, so that   $r(\tau)=\gamma^{|\tau|} \sigma_f^2$. Replacing  $\sigma_f^2=\frac{\sigma_v^2}{1-\gamma^2}$ in the previous expression, we obtain the formula of the stationary covariance function
\begin{eqnarray}
r(\tau)=\psi(t,t\pm\tau)=\frac{\gamma^{|\tau|}\sigma_v^2}{1-\gamma^2}, \quad \tau\in \mathbb{Z}.
\end{eqnarray}
 So that, after a transient, we can write 
\begin{equation}
[{\bm \Psi}]_{i,j}= \psi(i,j)=\frac{\gamma^{|i-j|}\sigma_v^2}{1-\gamma^2},
\end{equation}
for all $i,j=1,\ldots,N$.

\subsubsection{Covariance and precision matrices in the stationary regime}

Let us now assume that the marginal distribution
of $f_0$ is Gaussian with mean zero and variance $\frac{\sigma_v^2}{1-\gamma^2}$ (recall that $|\gamma| <1$), which is simply
the stationary distribution of this process. Therefore, considering as an example $N=5$, the covariance ${\bm \Psi}$ and the precision ${\bf P}={\bm \Psi}^{-1}$ matrices in the stationary regime is
 $$
 {\bm \Psi}=\sigma_v^2
\begin{bmatrix}
1 & \gamma &  \gamma^2  &  \gamma^3 &  \gamma^4\\
\gamma & 1 & \gamma & \gamma^2 & \gamma^3 \\
\gamma^2 & \gamma & 1 & \gamma & \gamma ^2\\
\gamma^3 & \gamma^2 & \gamma & 1 & \gamma \\
\gamma^4 & \gamma^3 & \gamma^2 & \gamma & 1 \\
\end{bmatrix},
\quad {\bf P}=\frac{1}{\sigma_v^2}
\begin{bmatrix}
1 & -\gamma &  0 &  0 &  0 \\
-\gamma & 1+\gamma^2 & -\gamma & 0 & 0 \\
0 & -\gamma & 1+\gamma^2 & -\gamma & 0 \\
0 & 0 & -\gamma & 1+\gamma^2 & -\gamma \\
0 & 0 & 0 & -\gamma & 1 \\
\end{bmatrix}.
 $$
Note that the precision matrix ${\bf P}$ is the tridiagonal matrix, i.e., with zero entries outside the diagonal and first off-diagonals. The tridiagonal form is due to the fact that $f_i$ and $f_j$ are {\it conditionally independent} for $|i-j|>1$ given the rest of variables. It is interesting to remark the entries in the covariance matrix ${\bm \Psi}$ only give direct information
about the {\it marginal} dependence structure, not about the {\it conditional} dependence.


\subsubsection{Filtering, smoothing, prediction}

Given the state-space model in Eq.~\eqref{aquiModelStateSpace}, at each time instant, we have an additional variable $f_t$ and an additional observation $y_t$. 
Several algorithms in the literature tackle different inference problems, corresponding to different posterior and/or predictive densities.
\newline
\newline
{\bf Complete and partial smoothing.} As we already have seen, the complete smoothing problem consider the joint posterior density 
$$
p(f_1,\ldots,f_N|y_1,\ldots,y_N)=p({\bf f}|{\bf y}).
$$
Other partial smoothing densities can be considered in this scenario, for instance,
$$
p(f_t|y_1,\ldots,y_N)=p(f_t|{\bf y}), \quad t<N.
$$ 
or considering different time instants, for instance, $p(f_{t_1}, f_{t_2}|y_1,\ldots,y_T)$ with $ t_1,t_2<N$. More generally, people are often interested in studying the posterior
$$
p(f_1,\ldots,f_t|y_1,\ldots,y_t), \quad t\leq N,
$$
where we analyze the vector $[f_1,\ldots,f_t]^{\top}$ considering only the observations $[y_1,\ldots,y_t]^{\top}$ (assuming unknown the data  $y_{t+1},\ldots,y_N$). 
\newline
\newline
{\bf Filtering.}  The filtering problem corresponds to the study of the following posterior densities  
\begin{equation}
p(f_t|y_1,\ldots,y_t),  \quad t\leq N,
\end{equation}
where we have only the variable $f_t$ given all the measurements $[y_1,\ldots,y_t]^{\top}$ obtained so far, i.e., assuming unknown the future observations  $y_{t+1},\ldots,y_N$. Generally, the people consider the sequential problem considering the sequence of filtering posteriors
    \begin{align*}
     &p(f_{1}|y_1),   \\
     &p(f_{2}|y_1,y_2),   \\
       &\vdots \\
     &p(f_t|y_1,\ldots,y_{t-1},y_t),
    \end{align*}
providing recursive solutions. 
\newline
\newline
{\bf Prediction at lag-$\tau$.} in time series analysis, one often consider the predictive density
    $$
    p(f_{t+\tau}|y_1,\ldots,y_{t}),  \quad \tau\geq 1,
    $$
where we are interested in inferring the variable $f_{t+\tau}$ in the future instant $t'=t+\tau$, observing only the measurements until time $t$.

\subsubsection{Discrete Kalman solution for filtering}

{\Remark The standard discrete Kalman filter provides the recursive equations for computing the mean $\widehat{\mu}_{t|t}$ and variance $\widehat{\sigma}_{t|t}^2$ of the filtering posterior density, i.e.,
\begin{eqnarray}
 p(f_t|y_1,\ldots,y_t)=\mathcal{N}(f_t|\widehat{\mu}_{t|t},\widehat{\sigma}_{t|t}^2 ).
\end{eqnarray} }
\newline
Let us also denote $\widehat{\mu}_{t|t-1}$ and  $\widehat{\sigma}_{t|t-1}^2$ the mean and variance of the predictive density
\begin{eqnarray}
 p(f_{t}|y_1,\ldots,y_{t-1})=\mathcal{N}(f_t|\widehat{\mu}_{t|t-1},\widehat{\sigma}_{t|t-1}^2).
\end{eqnarray}
The Kalman equations provide recursively the means and variances of these two densities as new observations are obtained. From the instant $t-1$ to $t$, the sequential Kalman solution, for computing mean and variance of the predictive density $p(f_{t}|y_1,\ldots,y_{t-1})$, is then given by 
\begin{gather}
\left\{
\begin{split}
\widehat{\mu}_{t|t-1}&=\gamma \widehat{\mu}_{t-1|t-1}  \\
\widehat{\sigma}_{t|t-1}^2&=\gamma^2 \widehat{\sigma}_{t-1|t-1}^2+\sigma_v^2 
\end{split}
\right.
\end{gather}
 and, for the filtering pdf $p(f_t|y_1,\ldots,y_t)$, we have
\begin{gather}
\left\{
\begin{split}
\widehat{\mu}_{t|t}&=\frac{\sigma_e^2}{\widehat{\sigma}_{t|t-1}^2+\sigma_e^2} \widehat{\mu}_{t|t-1}+\frac{\widehat{\sigma}_{t|t-1}^2}{\widehat{\sigma}_{t|t-1}^2+\sigma_e^2}y_t \\
\widehat{\sigma}_{t|t}^2 &=\frac{\widehat{\sigma}_{t|t-1}^2\sigma_e^2}{\widehat{\sigma}_{t|t-1}^2+\sigma_e^2}.
\end{split}
\right.
\end{gather}
 Defining the precision values as $\widehat{p}_{t|t-1}=\frac{1}{\widehat{\sigma}_{t|t-1}^2}$ and  $p_e=\frac{1}{\sigma_{e}^2}$, we can rewrite the last two equations as
\begin{gather}
\left\{
\begin{split}
\widehat{\mu}_{t|t}&=\frac{\widehat{p}_{t|t-1}}{\widehat{p}_{t|t-1}+\bar{p}_e} \bar{\mu}_{t}+\frac{\bar{p}_e}{\widehat{p}_{t|t-1}+\bar{p}_e}y_t, \\
\widehat{p}_{t|t}&=\widehat{p}_{t|t-1}+p_e.
\end{split}
\right.
\end{gather}

{\Remark  The standard Kalman filter focuses on the sequence of filtering densities, $p(f_t|y_1,\ldots,y_t)=\mathcal{N}(f_t|\widehat{\mu}_t,\widehat{\sigma}_t^2)$ for $t=1,\ldots,N$.
We can have a complete equivalence with the GP solution for smoothing, if we consider a Kalman approach for smoothing, i.e., considering the density 
$p(f_{1:N}|y_{1:N})=p({\bf f}|{\bf y})$.
}
\newline
\newline
Note that we have considered scalar values $f_1,\ldots,f_N$ and $y_1,\ldots,y_N$ to be coherent to the rest of the paper, and facilitate the comparison with the other techniques. However, the Kalman filter can be directly generalized for multivariate/multioutput case, i.e.,  at each iteration we can have vectors ${\bf f}_t$ and ${\bf y}_t$ of the observations.

\subsubsection{Backward filter for partial Kalman smoothing}
Let us consider that we have already run the {\it forward} Kalman filter described above, obtaining $\widehat{\mu}_{t|t-1}$, $\widehat{\sigma}_{t|t-1}^2$, $\widehat{\mu}_{t|t}$ and $\widehat{\sigma}_{t|t}^2$ for all $t=1,\ldots,N$. Now, we focus on the partial smoothing densities, i.e.,
\begin{eqnarray}\label{Kalmanpartialsmooth}
 p(f_{t}|y_1,\ldots,y_{N})=p(f_{t}|\y)=\mathcal{N}(f_t|\widehat{\mu}_{t|N},\widehat{\sigma}_{t|N}^2), \qquad \forall t<N.
\end{eqnarray}
Then, we can consider the following backward recursion (from $t=N-1$ to $t=1$):
\begin{gather}
\left\{
\begin{split}
\widehat{\mu}_{t|N}&=\widehat{\mu}_{t|t}+\gamma\frac{\widehat{\sigma}_{t|t}^2}{\widehat{\sigma}_{t+1|t}^2}(\widehat{\mu}_{t+1|N}-  \widehat{\mu}_{t+1|t}), \\
\widehat{\sigma}_{t|N}^2&=\widehat{\sigma}_{t|t}^2+\left(\gamma\frac{\widehat{\sigma}_{t|t}^2}{\widehat{\sigma}_{t+1|t}^2}\right)^2(\widehat{\sigma}_{t+1|N}-  \widehat{\sigma}_{t+1|t}).
\end{split}
\right.
\end{gather}
  For the solution of the complete smoothing problem, see \cite{Rauch65,Einicke07,Einicke08,Sarkka_2013}.
 Let us consider again the GP solution in Eqs.~\eqref{GPmu1}--\eqref{GPvar1} computed in a training input $t=1,2,\ldots,N$,  i.e.,
 \begin{eqnarray*}
\mu_{f|y}(t)&=& \widehat{f}(t)={\bm \psi}(t)^{\top}({\bm \Psi} +\sigma_e^2 {\bf I}_N)^{-1}\y, \label{GPmu1_2} \\
\sigma_{f|y}^2(t)&=&\psi(t,t)-{\bm \psi}(t)^{\top}({\bm \Psi} +\sigma_e^2 {\bf I}_N)^{-1}{\bm \psi}(t), \label{GPvar1_2}
\end{eqnarray*}
where $\psi(t,t')=\frac{\gamma^{|t-t'|}\sigma_v^2}{1-\gamma^2}$ with $t,t'\in \mathbb{N}^+$,  ${\bf y}=[y_1,\ldots,y_N]^{\top}$, 
$$
{\bm \psi}(t)=[\psi(t,1),\ldots,\psi(t,N)]^{\top},\quad \mbox{ and }\quad  [{\bm \Psi}]_{i,j}= \psi(i,j)=\frac{\gamma^{|i-j|}\sigma_v^2}{1-\gamma^2}.
 $$
 These mean and variance completely define the partial smoothing density
\begin{equation}\label{aquiGPpartialsmooth}
p(f_{t}|\y)=\mathcal{N}(f_t|\mu_{f|y}(t),\sigma_{f|y}^2(t)), \qquad \forall t<N.
\end{equation}
 {\Remark  It is possible to show that $\mu_{f|y}(t)=\widehat{\mu}_{t|N}$ and $\sigma_{f|y}^2(t)=\widehat{\sigma}_{t|N}^2$, clearly considering the equivalent kernel function, induced by the propagation equation in the state-space model.  
     }

\subsection{Continous-time state-space models}
 In order to obtain a complete equivalence to GP models and a sequential Kalman solutions we have to consider a continuous input variable,
 $t\in \mathbb{R}$. 
 In this scenario, the space-state model is  formed by a linear differential equation with constant coefficients  and an observation equation. The prior information is included by the  linear differential equation which plays the same role of the kernel/covariance function in the GP models.  A  linear differential equation with constant coefficients  of order $R$ with a Gaussian white noise input $v(t)$,
 \begin{equation}
\frac{d^R f(t)}{dt^R}+a_{R-1}\frac{d^{R-1} f(t)}{dt^{R-1}}+\cdots+a_{1}\frac{d f(t)}{dt}+a_0 f(t)=v(t),
\end{equation}
can be rewritten as a first order vector Markov process, i.e.,
 \begin{equation}\label{aquiMP}
\frac{d {\bf f}(t)}{dt}={\bf A} {\bf f}(t)+ {\bf b} v(t)
\end{equation} 
 where ${\bf f}(t)=\left[\frac{d^{R-1} f(t)}{dt^{R-1}},\ldots, \frac{d f(t)}{dt}, f(t) \right]^{\top}$ is an $R\times 1$ vector, 
 $$
 {\bf A}=
\begin{bmatrix}
 1  & 0 & \cdots & 0 & 0  \\
 0 &  1  &\cdots &0 & 0  \\
\vdots & \vdots&\vdots &\vdots & \vdots    \\
0 & 0&\cdots &1 & 0    \\
-a_{R-1} &-a_{R-2} & \cdots&-a_{1} & -a_0 \\
\end{bmatrix}, 
\qquad 
{\bf b}=
\begin{bmatrix}
0 \\
0 \\
0 \\
\vdots    \\
1 \\
\end{bmatrix},
 $$
 are a $R \times R$ matrix and an $R\times 1$ vector, respectively. Note also that 
 $$
 f(t)={\bf b}^{\top} {\bf f}(t),
 $$
 i.e., we can extract $f(t)$ from the vector ${\bf f}(t)$ by the multiplication above.
It possible to compute the {\it power spectral density} of ${\bf f}(t)$  (a) replacing $f(t)={\bf b}^{\top} {\bf f}(t)$ in Eq.~\eqref{aquiMP}, (b) taking the Fourier transform to both sides of Eq.~\eqref{aquiMP}, after replacing  $f(t)={\bf b}^{\top} {\bf f}(t)$. Moreover,  since the noise $v(t)$ is white, we have the its power spectral density is $S_V(\omega)=c$ where $c>0$. After some  algebra and rearrangement, this procedure yields  \cite{Hartikainen10,pmlr-v22-sarkka12}  
\begin{align}
S_F(\omega)&={\bf b}^{\top}( {\bf A}+j \omega {\bf I})^{-1}{\bf b} S_V(\omega) {\bf b}^{\top}\left[( {\bf A}+j \omega {\bf I})^{-1}\right]^{\top}{\bf b}, \\
&= c {\bf b}^{\top}( {\bf A}+j \omega {\bf I})^{-1}{\bf b}{\bf b}^{\top}\left[( {\bf A}+j \omega {\bf I})^{-1}\right]^{\top}{\bf b},
\end{align}
In the stationary state (i.e., when the process has run an infinite amount of time), the stationary covariance function $\psi(t,t')=\psi(\tau)$ of $f(t)$ (with $\tau=|t-t'|$) can be expressed as
inverse Fourier transform of its spectral density $S_F(\omega)$, hence
\begin{align}
\psi(\tau)=\frac{1}{2\pi} \int_{-\infty}^{+\infty} S_F(\omega)e^{j\omega \tau} d\omega.
\end{align}
We have shown that a linear differential equation determines a covariance function over $f(t)$  \cite{Hartikainen10,pmlr-v22-sarkka12,Sarkka_2013}. 
 
\section{Summary}\label{ConclSect}
In this work, we have provided a joint introduction to RVMs and GPs for regression, including within this framework the tasks of filtering, smoothing, and interpolation. The probabilistic derivation of both methods is given, along  with several observations and recommendations for the use of these methods in practice. We have highlighted the connections between them and to related techniques such as kernel ridge regression, kernel smoothers, Fourier interpolators and Kalman filtering. We have also remarked the benefits and drawbacks of each schemes. 
RVMs allow the choice of more general basis functions whereas the behavior of the predictive variance is generally counterintuitive. GPs present a good behavior of the predictive variance but the choice of kernel functions is more restrictive. The Kalman smoothing method provides the same solution as that of a GP with a specific kernel function, which is implicitly induced by the considered propagation equation in the state-space model. 

{ \footnotesize
\section*{ \footnotesize Acknowledgements}
Luca Martino acknowledges  support by  the Agencia Estatal de Investigaci{\'o}n AEI (project PID2019-105032GB-I00).}

\bibliographystyle{IEEEtranN}
\bibliography{heretical} 

\appendix

\section{Alternative formulation of the RVM variance}
\label{RVMvarianceAPP}
In this section, we analyze the expression of the variance of RVM. 
\newline
Let us consider the following generic matrices ${\bf Z}$ of size $M\times M$,  ${\bf U}$ of size $N\times M$,  ${\bf L}$ of size $N\times N$ and  ${\bf V}$ of size $M\times N$, the following {\it matrix inversion lemma} \citep{Hager89}, \citep[Appendix A]{rasmussen2003gaussian} is satisfied,
\begin{equation}
\left({\bf Z}+{\bf U}{\bf L}{\bf V}^{\top} \right)^{-1}={\bf Z}^{-1}-{\bf Z}^{-1}{\bf U}\left({\bf L}^{-1}+{\bf V}^{\top}{\bf Z}^{-1}{\bf U}\right)^{-1}{\bf V}^{\top}{\bf Z}^{-1}.
\end{equation}
Using this matrix inversion lemma with $Z^{-1}={\bm \Sigma}_\rho$, ${\bf L}^{-1}=\sigma_e^2{\bf I}_N$ and $U=V={\bm \Psi}^{\top}$ and considering the variance of RVM, we obtain
\begin{equation}
\left({\bm \Sigma}_\rho^{-1}+\frac{1}{\sigma_e^2}{\bm \Psi}^{\top}{\bm \Psi} \right)^{-1}={\bm \Sigma}_\rho-{\bm \Sigma}_\rho{\bm \Psi}^{\top}\left(\sigma_e^2{\bf I}_N+{\bm \Psi}{\bm \Sigma}_\rho{\bm \Psi}^{\top}\right)^{-1}{\bm \Psi}{\bm \Sigma}_\rho.
\end{equation}
Replacing in Eq.~\eqref{Eq41var}, where we consider the matrix  ${\bm \Psi}$ instead of ${\bm \Psi}$, we have
\begin{eqnarray*}
\widehat{\sigma}(\bx)&=& {\bm \psi}(\bx)^\top \left({\bm \Sigma}_\rho-{\bm \Sigma}_\rho{\bm \Psi}^{\top}\left(\sigma_e^2{\bf I}_N+{\bm \Psi}{\bm \Sigma}_\rho{\bm \Psi}^{\top}\right)^{-1}{\bm \Psi}{\bm \Sigma}_\rho\right){\bm \psi}(\bx),   \\
&=&  {\bm \psi}(\bx)^\top{\bm \Sigma}_\rho{\bm \psi}(\bx)- {\bm \psi}(\bx)^\top{\bm \Sigma}_\rho{\bm \Psi}^{\top}\left(\sigma_e^2{\bf I}_N+{\bm \Psi}{\bm \Sigma}_\rho{\bm \Psi}^{\top}\right)^{-1}{\bm \Psi}{\bm \Sigma}_\rho{\bm \psi}(\bx).
\end{eqnarray*}

\end{document}